\documentclass[10pt]{article} 
\usepackage[preprint]{tmlr}

\usepackage{amsmath,amsfonts,bm}

\def\eqref#1{equation~\ref{#1}}

\def\1{\bm{1}}

\DeclareMathAlphabet{\mathsfit}{\encodingdefault}{\sfdefault}{m}{sl}
\SetMathAlphabet{\mathsfit}{bold}{\encodingdefault}{\sfdefault}{bx}{n}

\usepackage[utf8]{inputenc}
\usepackage[T1]{fontenc}
\usepackage{url}
\usepackage{booktabs}
\usepackage{amsfonts}
\usepackage{nicefrac}
\usepackage{microtype}
\usepackage{xcolor}
\usepackage{colortbl}
\definecolor{revcolor}{RGB}{0,90,180}
\newif\ifshowrev
\showrevfalse

\newenvironment{revised}{\ifshowrev\color{revcolor}\fi}{}
\usepackage{graphicx}
\usepackage{multirow}
\usepackage{amsmath}
\usepackage{enumitem}
\usepackage{hyperref}
\usepackage{caption}
\usepackage{makecell}
\usepackage{needspace}

\title{Beyond Row-Level Prediction: Evaluating and Learning\\Composition for Reusable Table Representations}
\author{
\centering
Ali Younes$^{1,2}$\thanks{Work done during an internship at SAP} \qquad
Saeed Ghoorchian$^{1}$ \qquad
Maximilian Schambach$^{1}$ \\[0.4em]
Georgia Chalvatzaki$^{2,3}$ \qquad
Johannes Höhne$^{1}$ \\[0.9em]
{\normalfont\small $^{1}$SAP SE, Berlin, Germany} \\
{\normalfont\small $^{2}$TU Darmstadt, Darmstadt, Germany} \\
{\normalfont\small $^{3}$Hessian AI, Darmstadt, Germany}
\\[0.4em]
{\normalfont\small Correspondence: \texttt{ali@robot-learning.de}}
}

\begin{document}
\maketitle

\begin{abstract}
\begin{revised}
Tables are compositional, with content spread across a schema, cell values, rows, and columns. Many large-scale operations such as retrieval, overlap estimation, deduplication, and data curation need these local signals composed into one reusable embedding vector per table. Table representation methods, however, are currently mostly evaluated through row-level prediction or supervised downstream tasks, which do not measure the quality of table-level embeddings directly. To address these limitations, we introduce a benchmark that evaluates table-level embeddings directly through four desiderata, namely consistency under partial views, discriminability across label granularities, robustness to benign perturbations, and efficiency. Across twelve methods on synthetic families and real open-source corpora, lightweight schema- and text-based fingerprints are consistently among the strongest. The sharpest contrast comes from two tabular foundation models, TabPFN and \textsc{ConTextTab}. Although both are strong row-level predictors, their states, composed by mean, max, or three further parameter-free pooling rules and compared by cosine similarity, trail the strongest fingerprints. We therefore ask whether the gap lies in the encoder or in the compositional rule. Keeping \textsc{ConTextTab} frozen, we learn only the compositional rule over its cell states, using support-conditioned distribution matching (SCDM), a self-supervised loss we propose that matches the distribution of embedding similarities to the distribution of partition overlaps within and across tables. Across all fifteen head configurations we trained, the learned rule raises the Spearman correlation between embedding similarity and partial-view overlap from $0.04$ to $0.43$--$0.50$ on synthetic data and improves triplet ranking, clustering, and retrieval, with smaller but systematic gains on real data. A downstream parent-table retrieval task confirms these findings, with fingerprints remaining strongest and learned composition improving markedly over fixed pooling. How local states are composed, and with what objective, substantially affects the quality of the table-level embeddings extracted from a predictive model.
\end{revised}
\end{abstract}

\section{Introduction}

Foundation models transformed language and vision not only by improving task accuracy, but also by producing reusable representations that support retrieval, transfer, and lightweight adaptation~\citep{devlin2019bert, radford2021clip, caron2021dino}. Tabular data has not yet reached this point. Many practical operations on table corpora, such as indexing, deduplication, overlap estimation, retrieval, and curation, act on whole tables, and they need one embedding vector per table that stays meaningful when only part of the table is observed. Table representation methods, however, are mostly evaluated through row-level prediction or supervised downstream tasks~\citep{hollmann2025tabpfn, spinaci2025ConTextTab, hoppe2025tabembedbench}. Naturally, these evaluations show how well a model predicts, not how good its table-level embedding is.

Tables are compositional. Their content is spread across column names, cell values, and row and column structure, and every table-level embedding is built by composing these local signals. Some methods compose them implicitly, for example by hashing a serialized table or by reading a table-level node of a graph encoder~\citep{chen2023hytrel, pugnaloni2025armadilo}. Others compose them explicitly, with an encoder that produces row- or cell-level states and a compositional rule, usually a fixed pooling such as averaging, that maps them to one embedding. This view separates two questions that row-level evaluation conflates, namely what the local states contain and how well a given compositional rule turns them into a reusable table-level embedding. We use \emph{composition} in this representational sense and do not study compositional generalization.

\begin{figure}[t]
    \centering
    \vspace{-1.0cm}
    \includegraphics[width=\linewidth]{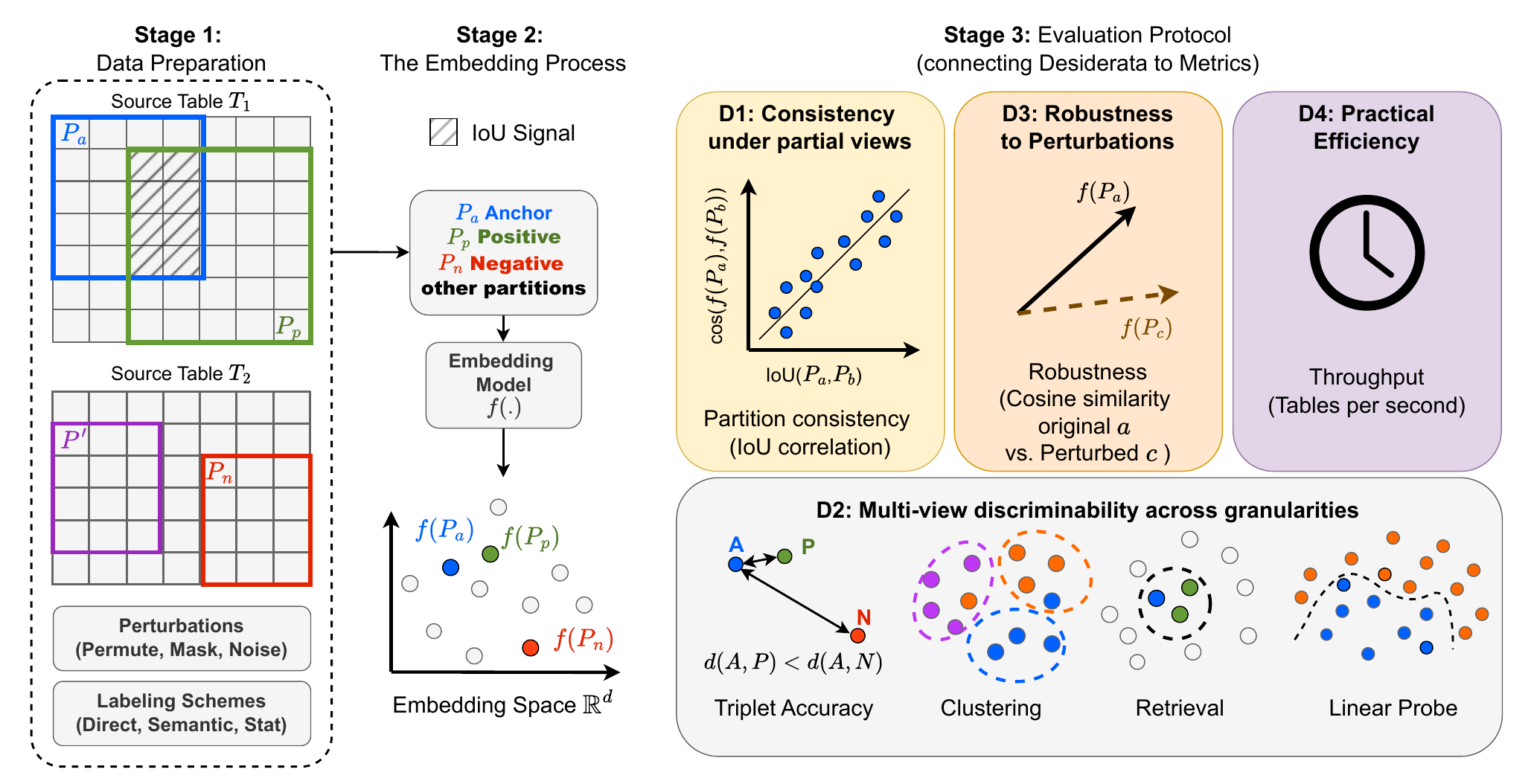}
    \caption{The evaluation pipeline. Stage 1 samples partitions from source tables and attaches labeling schemes and perturbations. Stage 2 maps each partition to a single table-level embedding, either directly or by composing local encoder states with a fixed or learned compositional rule. Stage 3 measures the four desiderata through concrete metrics, so lightweight baselines and foundation-model-derived representations are compared under the same observation interface.}
    \label{fig:intro}
    \vspace{-0.8cm}
\end{figure}

We introduce an evaluation pipeline that evaluates table-level embeddings directly through four desiderata, namely consistency under partial views (D1), discriminability across label granularities (D2), robustness to benign perturbations (D3), and efficiency (D4). Partitions sampled from source tables give controlled partial views, and embeddings are compared by cosine similarity. This lets the evaluation pipeline measure, for example, whether the similarity of two views tracks their overlap, a relation that recent work treats as a signal in its own right~\citep{pugnaloni2025armadilo, zecchini2024sloth}. We compare twelve methods covering schema and statistical fingerprints, serialization-based text encoders, specialized table encoders, and representations derived from tabular foundation models. The comparison covers synthetic families with known generative factors~\citep{qu2025tabicl} and real corpora from CARTE, OpenML-CC18, and OpenML-CTR23~\citep{kim2024carte, bischl2021openml, fischer2023openmlctr23}.

Lightweight schema- and text-based fingerprints are consistently among the strongest methods (Section~\ref{sec:results}). The sharpest contrast comes from tabular foundation models, which are strong row-level predictors~\citep{hollmann2023tabpfn, spinaci2025ConTextTab}. We study two TFMs that represent the two common output interfaces, TabPFN, which produces row-level states conditioned on a target, and \textsc{ConTextTab}, which produces semantics-aware cell-level states. Composed by mean, max, or three further parameter-free pooling rules, the states of both models trail the strongest fingerprints on partial-view consistency and discriminability.

The fixed composition results raise the central question of the paper, namely whether the states of these models lack the relevant information or fixed pooling fails to expose it. To separate the two, we keep a pretrained \textsc{ConTextTab} encoder frozen and change only the compositional rule, replacing fixed pooling with a lightweight learned head (Section~\ref{sec:frozen_ctt}). The head needs a training signal that reflects the desiderata and requires no external labels. We obtain it from the benchmark's own sampling, which records for every pair of partitions whether they come from the same table and how much they overlap. We propose support-conditioned distribution matching (SCDM), a loss that matches the distribution of embedding similarities to the distribution of partition overlaps, both within a source table and across a batch of tables. Because the head sees only the frozen states, its natural baseline is fixed composition of the same states, not a different table encoder. The learned compositional rule substantially improves partial-view consistency, triplet ranking, clustering, and retrieval.

Finally, a downstream parent-table retrieval task validates these findings (Section~\ref{sec:downstream}). Across the twelve benchmark methods, the benchmark ranking closely tracks retrieval quality, and the learned compositional rule improves retrieval markedly over fixed pooling, although the best fingerprints still lead. For the models we study, weak table-level embeddings under fixed pooling therefore do not imply weak encoder states. The compositional rule, and the objective that shapes it, deserve the same attention as the encoder.

In summary, our contributions are as follows:
\begin{itemize}[leftmargin=*]
\item \textbf{An evaluation pipeline for table-level embeddings.} Four desiderata are measured on partial views across twelve methods and two corpora, and the findings are validated by a downstream parent-table retrieval task. The evaluation pipeline separates a model's local states from its compositional rule, so any model that exposes local states can be evaluated with fixed or learned rules.
\item \textbf{An empirical finding.} Lightweight fingerprints are among the strongest table-level embeddings, and the states of two tabular foundation models trail them under five parameter-free pooling rules.
\item \textbf{SCDM, a self-supervised loss for learning compositional rules from partial views.} On frozen \textsc{ConTextTab} states, it substantially improves partial-view consistency, discriminability, and parent-table retrieval, showing that learning a compositional rule can improve the quality of the table-level embedding.
\end{itemize}

\section{Related Work}
\label{sec:related_work}

Tabular foundation models (TFMs) have achieved strong predictive performance through in-context learning. Their architectures broadly follow two approaches. TabICL-like models, including TabICL~\citep{qu2025tabicl}, TabICLv2~\citep{qu2026tabiclv2}, TabPFN-3~\citep{grinsztajn2026tabpfn3}, TabFM~\citep{kong2026tabfm}, and Kumo Tabular~\citep{nvidia2026kumo}, perform staged attention to aggregate row- and column-wise information, effectively compressing each row into a compact representation before a separate in-context learning stage for prediction. In contrast, TabPFN-v2-like models, including TabPFN-2.5~\citep{grinsztajn2025tabpfn25}, Mitra~\citep{zhang2025mitra}, Mitra-v2~\citep{tao2026mitrav2}, and \textsc{ConTextTab}~\citep{spinaci2025ConTextTab}, interleave row- and column-wise attention while retaining cell-level representations throughout. FlexTab~\citep{polewczyk2026flextab} learns target-agnostic row embeddings for use with task-specific decoders. We study two predictive TFMs with different representation interfaces: TabPFN, from which we extract target-conditioned row embeddings with the embedding approach of~\citet{ye2025a}, the interface that TabICL-like models also provide, and \textsc{ConTextTab}, which exposes contextual cell embeddings. These models allow us to evaluate fixed composition at different representation granularities, while the cell-level states of \textsc{ConTextTab} support a controlled learned-composition experiment. Our goal is to study the resulting table-level representation interface, rather than to compare the predictive performance of the latest TFMs.

A second line of work studies table representations more directly. Earlier pretrained encoders such as TaBERT~\citep{yin2020tabert}, TABBIE~\citep{iida2021tabbie}, and TURL~\citep{deng2020turl} learn representations for table understanding tasks. Observatory~\citep{cong2023observatory} characterizes relational-table embeddings, while subsequent work~\citep{hoppe2025tabembedbench} compares task-agnostic embeddings on downstream learning and anomaly-detection tasks. Recent benchmarks have broadened this evaluation. TEmBed~\citep{vogel2026tembed} evaluates embeddings at the cell, row, column, and table levels across data tasks, and demonstrates that the best model depends on the task and the granularity, while TEmBed-T~\citep{poostforoushan2026tembedt} extends it to table-level embeddings along several complementary properties and finds that retrieval performance alone does not capture embedding quality. TRL-Bench~\citep{pang2026trlbench} evaluates row, column, and table embeddings using shared lightweight heads, while TabBench~\citep{qiang2026tabembed} evaluates embedding models on tabular classification and retrieval tasks. Our benchmark is complementary in two ways. First, we evaluate embeddings from controlled partial views of the same source table, measuring whether their similarity tracks the overlap between views. Second, we compare fixed and learned composition over the same frozen encoder states and under the same evaluation interface, allowing us to study the role of composition independently of encoder training.

A third line of work proposes specialized representation methods for structured tables or targets related table-level similarity tasks. Our benchmark includes HyTREL~\citep{chen2023hytrel}, which encodes tables through hypergraph structure, and Armadillo~\citep{pugnaloni2025armadilo}, which learns graph-based table embeddings for overlap estimation; relatedly, recent work~\citep{zecchini2024sloth} formalizes largest-table-overlap detection as a problem in its own right. Whole-table representations are equally central to table discovery, from Table Union Search~\citep{nargesian2018tus} over open data to contrastive and sketch-based encoders such as Starmie~\citep{fan2023starmie} and TabSketchFM~\citep{khatiwada2024tabsketchfm}, and critical re-evaluations of table-discovery benchmarks~\citep{boutaleb2025fishy} and metric-learning evaluations~\citep{musgrave2020metric} caution that lexical-overlap artifacts and evaluation design can inflate reported gains.

Composing a set of local vectors into one representation is a long-standing problem outside tabular learning. Unweighted averaging of word vectors is a weak sentence representation, and simple reweighting already improves it substantially~\citep{arora2017simple}. Permutation-invariant set functions~\citep{zaheer2017deepsets}, attention-based pooling over sets of instances~\citep{ilse2018attention}, and learned pooling with seed vectors in the Set Transformer~\citep{lee2019settransformer} replace fixed averages with trainable aggregators. Our learned heads follow this line and pool the cell states of a frozen TFM into rows and rows into a table with learned weights. Their training signal comes from SCDM, the loss we propose, which builds on listwise distribution matching as in ListNet~\citep{cao2007listnet}, using partition overlap to define target distributions over two supports: partitions of the same source table and the full batch. Contrastive objectives used to train table encoders, such as those of Starmie~\citep{fan2023starmie} and TabEmbed~\citep{qiang2026tabembed}, update the whole encoder, whereas SCDM trains only the compositional rule over frozen states, so the experiment isolates the table-level interface rather than comparing encoders.

\section{Evaluating Whole-Table Representations}
\label{sec:desiderata}

\subsection{Table-level embeddings from local states}
\label{sec:composition}

\begin{revised}
A table representation method maps a table $T$ to a vector $z(T)\in\mathbb{R}^d$, and we compare two tables by cosine similarity, $\mathrm{sim}(T_a,T_b)=\cos\bigl(z(T_a),z(T_b)\bigr)$. Methods differ in how they build $z(T)$. Fingerprints, serialization-based methods, and HyTREL emit a table-level embedding directly. TFMs produce local states, one per row (TabPFN) or per cell (\textsc{ConTextTab}). A table-level embedding requires a compositional rule $C$,
\begin{equation}
    z(T) = C\bigl(H(T)\bigr),
\end{equation}
where $H(T)$ denotes the set of local encoder states. In the main evaluation pipeline (Section~\ref{sec:results}), $C$ is a fixed, parameter-free rule, namely mean, max, sum, min, or standard-deviation pooling over rows or cells. In Section~\ref{sec:frozen_ctt}, we keep the encoder frozen and replace $C$ by a learned head $C_\theta$.

All four desiderata below are properties of the resulting table-level embeddings. D1--D3 are computed from the embeddings themselves, and D4 concerns the cost of computing them. Wherever two embeddings are compared, we use cosine similarity for every method because it is the standard similarity for embedding retrieval and is invariant to the scale of an embedding, and the learned heads are trained for it. The benchmark therefore evaluates the table-level interface formed by an encoder, a compositional rule, and cosine similarity, rather than an intrinsic property of the encoder. A poor score under fixed pooling does not by itself show that the encoder states lack the relevant information. Because any model that exposes local states can be paired with fixed or learned rules, the evaluation applies unchanged to future TFMs.
\end{revised}

\subsection{Desiderata}
\begin{table*}[t]
\centering
\caption{Benchmark summary. Part (A) lists the families of table representation methods compared under the shared observation interface, and part (B) lists the desiderata, main metrics, and corresponding evaluation targets. Full details are provided in Appendix~\ref{app:embedder_details} and Appendix~\ref{app:evaluation}.}
\label{tab:benchmark_summary_main}
\footnotesize
\setlength{\tabcolsep}{5pt}

\textbf{(A) Table representation methods}

\vspace{0.2em}
\begin{tabular}{p{2.5cm}p{4.5cm}p{8.0cm}}
\toprule
\textbf{Family} & \textbf{Methods} & \textbf{Main signal / representation} \\
\midrule
Structural \& statistical &
HashingSchema, SchemaContent, TableStatistics, StatisticalSummary, MatrixFactorization &
Schema tokens, metadata, summary statistics, and numeric spectrum; direct table-level embeddings. \\
\midrule
Serialization-based text &
HashingText, LLMText &
CSV-style serialization encoded by feature hashing (HashingText) or by the all-MiniLM-L6-v2 sentence-transformer (LLMText). \\
\midrule
Specialized representation &
TableVectorizer, HyTREL, Armadillo &
Mixed-type row vectorization, hypergraph encoding, or graph neural encoding with pooled table-level embeddings. \\
\midrule
Foundation-model-derived &
TabPFN, \textsc{ConTextTab} &
Row- or cell-level hidden states of pretrained TFMs, composed into one table-level embedding by a fixed pooling rule (mean or max in the main tables). \\
\bottomrule
\end{tabular}

\vspace{0.6em}
\textbf{(B) Desiderata and metrics}

\vspace{0.2em}
\begin{tabular}{p{0.9cm}p{3.0cm}p{5.0cm}p{6.1cm}}
\toprule
\textbf{Des.} & \textbf{Main metric(s)} & \textbf{What it measures} & \textbf{Evaluation target} \\
\midrule
D1 &
Spearman correlation between IoU and cosine similarity &
Whether representation similarity tracks overlap across partial views of the same source table &
Partial-view consistency \\
\midrule
D2 &
TR-Avg, CL-Avg, Recall@5, linear probe accuracy &
Whether representations support local ranking, global grouping, retrieval, and label decoding under multiple labeling schemes &
Discriminability across \emph{Direct}, \emph{Semantic}, \emph{Semantic+Difficulty} (synthetic only), and \emph{Stat} labels \\
\midrule
D3 &
Cosine similarity after permutation, masking, and typed noise &
Whether representations remain stable under benign perturbations &
Robustness \\
\midrule
D4 &
Tables per second &
Embedding throughput under the shared observation interface &
Efficiency \\
\bottomrule
\end{tabular}
\end{table*}

\begin{revised}
Corpus-scale operations such as retrieval, overlap estimation, deduplication, source recovery, data curation, and the table-selection step of tabular RAG use table-level embeddings directly, without downstream fine-tuning. We derive four desiderata from these operations and measure each through proxy tasks, since labeled ground truth for table similarity is scarce. Each desideratum tests a property of the embedding, so a method is better when its embeddings satisfy D1-D4. The desiderata operationalize these uses rather than give an exhaustive definition of table representation quality, and Table~\ref{tab:benchmark_summary_main} and Figure~\ref{fig:intro} map them to their metrics.
\paragraph{Consistency under partial views (D1).}
Tables are often observed only in part, through cached extracts, sampled partitions, or truncated context windows, and overlap estimation and approximate deduplication must compare such partial views. D1 therefore asks whether the cosine similarity of two partial views of the same table increases with their overlap, measured by the Spearman correlation with their intersection over union. No method receives partition coordinates or overlap values, which are used only for scoring, so D1 measures how well the observed content preserves overlap-sensitive similarity.
\paragraph{Discriminability across label granularities (D2).}
Curation operations rely on different equivalence relations, namely source identity for deduplication, semantic groups for stratified sampling across domains and tasks, and coarse statistical types for filtering and balancing. D2 therefore asks whether table-level embeddings group and separate partitions under each of these label granularities, measured locally by triplet ranking, globally by clustering, and operationally by retrieval and linear probing. A method that relies on a single cue, such as schema overlap, does not score well under all of them (Appendix~\ref{sec:app_results}).
\paragraph{Robustness to benign perturbations (D3).}
Indexing, nearest-neighbor retrieval, and deduplication over automatically extracted tables should not depend on row or column order or on small errors in cell values. D3 therefore measures the cosine similarity between a partition and its row- and column-permuted, masked, and noise-perturbed versions, which should remain close to one.
\paragraph{Practical efficiency (D4).}
Corpus-scale use requires embeddings that are cheap to compute. D4 measures throughput in tables per second, which together with D1--D3 gives the quality-cost trade-off of each method and is measured in a controlled encode-only setting (Appendix~\ref{sec:app_efficiency}).
\end{revised}

\subsection{Benchmark setup}

\label{sec:benchmark_setup}

Although representation methods conceptually operate on full tables, our benchmark applies them to \emph{table partitions} sampled from the same source table. These partitions are valid subtables and allow us to compare diverse representation methods under controlled partial views, perturbations, and shared input budgets.

We evaluate a broad set of table representation methods (Table~\ref{tab:benchmark_summary_main}) spanning lightweight fingerprints, serialization-based text encoders, specialized relational encoders, and representations derived from tabular foundation models. Methods that emit row- or cell-level outputs are reduced to a single table-level embedding by a fixed pooling rule. We include two tabular foundation models that represent the two common output interfaces, TabPFN with row-level states conditioned on a target and \textsc{ConTextTab} with semantics-aware cell-level states. Serialization-based methods operate on a deterministic CSV-style representation of the same table view. To assess discriminability under different notions of similarity, we attach labels at multiple granularities. \emph{Direct} labels correspond to source-table identity, \emph{Semantic} labels capture higher-level grouping, \emph{Semantic+Difficulty} further refines synthetic semantic labels by generator difficulty, and \emph{Stat} labels encode coarse statistical signatures. Details are given in Appendix~\ref{app:embedder_details},~\ref{app:evaluation}, and~\ref{sec:exp_params}.

We evaluate on both synthetic and real open-source corpora in order to balance controlled structure with real-world variability. In both settings, we sample $100$ source tables and compute the benchmark metrics on $K_p=10$ partitions per table under a shared observation budget. We repeat the full evaluation over $10$ random seeds, yielding $10{,}000$ sampled partition instances per embedder before metric-specific pair, triplet, and retrieval constructions. The synthetic benchmark includes families with controlled generative factors spanning formulaic, geometric, temporal, and SCM-based structure, while the real-data benchmark is built from CARTE, OpenML-CC18, and OpenML-CTR23. Full dataset construction details, sampling configurations, and label definitions are deferred to Appendix~\ref{app:data_details} and Appendix~\ref{sec:exp_params}.

\section{How Good Are Table-Level Embeddings under Fixed Composition?}
\label{sec:results}

We first measure the quality of table-level embeddings obtained from TFM states by standard parameter-free composition, relative to lightweight and specialized representations evaluated under the same interface. All twelve methods are evaluated on the synthetic and real corpora, with TabPFN and \textsc{ConTextTab} each composed by mean and by coordinatewise max pooling; three further pooling rules follow in Section~\ref{sec:pooling_main}. Tables~\ref{tab:results_synthetic} and~\ref{tab:results_real} report D1--D3 and the resulting ranks; D4 is measured separately (Appendix~\ref{sec:app_efficiency}).

\subsection{Synthetic benchmark results}
\begin{figure*}[t]
\centering
\small
\captionof{table}{Evaluation Results for Synthetic data (mean $\pm$ 95\% CI over 10 seeds).}
\label{tab:results_synthetic}
{\renewcommand{\arraystretch}{0.88}%
\resizebox{\textwidth}{!}{
\begin{tabular}{lcc|ccccc|cccc|c}
\toprule
\multirow{2}{*}{Embedder} & \multicolumn{2}{c}{D1 Consistency} & \multicolumn{5}{c}{D2 Label-Based - Avg} & \multicolumn{4}{c}{D3 Robustness} & \multicolumn{1}{c}{Overall} \\
\cmidrule(lr){2-3} \cmidrule(lr){4-8} \cmidrule(lr){9-12} \cmidrule(lr){13-13}
 & Spearman & Rank & TR-Avg & CL-Avg & R@5 & LP & Rank & Perm & Noise & Mask & Rank & Avg Rank \\
\midrule
HashingSchema & \textbf{0.59$\pm$0.01} & \cellcolor{green!60}1.00 & \textbf{0.53$\pm$0.01} & \textbf{0.54$\pm$0.02} & 0.71$\pm$0.01 & 0.61$\pm$0.02 & 4.75 & \textbf{1.00$\pm$0.00} & 0.98$\pm$0.00 & 0.99$\pm$0.00 & 5.50 & \cellcolor{green!60}3.75 \\
SchemaContent & 0.29$\pm$0.02 & 7.00 & 0.01$\pm$0.00 & 0.26$\pm$0.02 & 0.78$\pm$0.01 & 0.60$\pm$0.02 & 9.25 & \textbf{1.00$\pm$0.00} & \textbf{1.00$\pm$0.00} & \textbf{1.00$\pm$0.00} & \cellcolor{green!40}3.33 & 6.53 \\
TableStatistics & 0.34$\pm$0.03 & 6.00 & 0.13$\pm$0.01 & 0.34$\pm$0.02 & 0.71$\pm$0.01 & 0.50$\pm$0.02 & 9.25 & \textbf{1.00$\pm$0.00} & \textbf{1.00$\pm$0.00} & \textbf{1.00$\pm$0.00} & \cellcolor{green!20}3.50 & 6.25 \\
StatisticalSummary & 0.38$\pm$0.05 & 5.00 & 0.36$\pm$0.01 & 0.44$\pm$0.01 & 0.91$\pm$0.01 & 0.56$\pm$0.02 & \cellcolor{green!20}4.25 & \textbf{1.00$\pm$0.00} & \textbf{0.99$\pm$0.00} & \textbf{1.00$\pm$0.00} & 4.17 & 4.47 \\
MatrixFactorization & 0.42$\pm$0.04 & \cellcolor{green!20}3.00 & 0.23$\pm$0.01 & 0.32$\pm$0.02 & 0.79$\pm$0.02 & 0.49$\pm$0.02 & 8.00 & \textbf{1.00$\pm$0.00} & \textbf{1.00$\pm$0.00} & \textbf{1.00$\pm$0.00} & \cellcolor{green!60}2.17 & \cellcolor{green!20}4.39 \\
HashingText & 0.54$\pm$0.04 & \cellcolor{green!40}2.00 & 0.47$\pm$0.01 & 0.46$\pm$0.03 & \textbf{1.00$\pm$0.00} & \textbf{0.90$\pm$0.03} & \cellcolor{green!60}2.00 & \textbf{1.00$\pm$0.00} & 0.65$\pm$0.02 & 0.97$\pm$0.00 & 8.17 & \cellcolor{green!40}4.06 \\
LLMText & 0.28$\pm$0.02 & 9.00 & 0.51$\pm$0.01 & \textbf{0.54$\pm$0.03} & 0.88$\pm$0.01 & 0.77$\pm$0.02 & \cellcolor{green!40}2.50 & 0.97$\pm$0.00 & 0.98$\pm$0.00 & \textbf{0.99$\pm$0.00} & 7.00 & 6.17 \\
TableVectorizer & 0.29$\pm$0.01 & 8.00 & 0.17$\pm$0.01 & 0.22$\pm$0.01 & 0.74$\pm$0.01 & 0.38$\pm$0.02 & \cellcolor{red!40}11.50 & 0.29$\pm$0.01 & \textbf{1.00$\pm$0.00} & 0.80$\pm$0.02 & \cellcolor{red!20}10.33 & 9.94 \\
HyTREL & 0.16$\pm$0.03 & 10.00 & 0.24$\pm$0.01 & 0.33$\pm$0.02 & 0.84$\pm$0.01 & 0.72$\pm$0.02 & 5.00 & \textbf{0.99$\pm$0.00} & 0.98$\pm$0.00 & 0.89$\pm$0.00 & 9.33 & 8.11 \\
Armadillo & 0.39$\pm$0.03 & 4.00 & 0.31$\pm$0.01 & 0.31$\pm$0.02 & 0.89$\pm$0.01 & 0.67$\pm$0.04 & 5.00 & \textbf{1.00$\pm$0.00} & 0.48$\pm$0.01 & 0.87$\pm$0.00 & 10.17 & 6.39 \\
TabPFN (mean) & -0.08$\pm$0.05 & \cellcolor{red!40}13.00 & 0.17$\pm$0.00 & 0.25$\pm$0.02 & 0.73$\pm$0.01 & 0.50$\pm$0.02 & 10.75 & 0.75$\pm$0.02 & 0.93$\pm$0.01 & 0.99$\pm$0.00 & 10.00 & \cellcolor{red!40}11.25 \\
TabPFN (max) & -0.17$\pm$0.04 & \cellcolor{red!60}14.00 & 0.21$\pm$0.01 & 0.25$\pm$0.01 & 0.71$\pm$0.01 & 0.48$\pm$0.02 & \cellcolor{red!20}11.25 & 0.74$\pm$0.02 & 0.93$\pm$0.01 & 0.95$\pm$0.00 & \cellcolor{red!60}12.00 & \cellcolor{red!60}12.42 \\
\textsc{ConTextTab} (mean) & 0.04$\pm$0.03 & 11.00 & 0.18$\pm$0.00 & 0.27$\pm$0.02 & 0.70$\pm$0.01 & 0.55$\pm$0.02 & 9.75 & 0.97$\pm$0.00 & 0.97$\pm$0.00 & 0.97$\pm$0.00 & \cellcolor{red!40}10.33 & 10.36 \\
\textsc{ConTextTab} (max) & 0.03$\pm$0.04 & \cellcolor{red!20}12.00 & 0.08$\pm$0.01 & 0.26$\pm$0.02 & 0.68$\pm$0.01 & 0.52$\pm$0.01 & \cellcolor{red!60}11.75 & 0.97$\pm$0.00 & 0.97$\pm$0.00 & 0.97$\pm$0.00 & 9.00 & \cellcolor{red!20}10.92 \\
\bottomrule
\end{tabular}
}}

\vspace{0.35em}

\centerline{\includegraphics[width=0.88\textwidth]{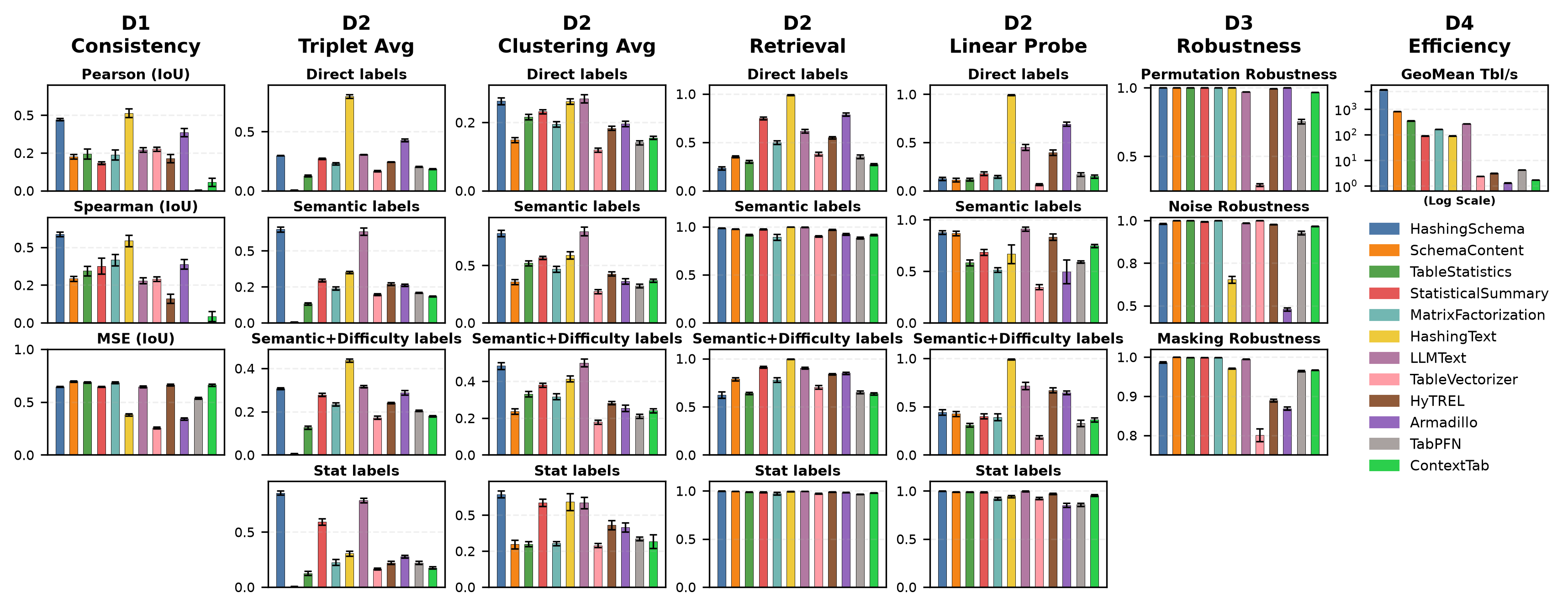}}
\captionof{figure}{Metric summary for Synthetic data across the benchmark dimensions.}
\label{fig:metrics_bar_grid_synthetic}
\end{figure*}

On synthetic data (Table~\ref{tab:results_synthetic}, Figure~\ref{fig:metrics_bar_grid_synthetic}), lightweight hashing and statistical-summary embedders are among the strongest methods, with HashingSchema and HashingText leading the quality-only ranking.
HashingSchema achieves the best D1 consistency and the best overall average rank, while HashingText provides the strongest D2 retrieval and linear probe signals (both near ceiling on Recall@5 and LP), and LLMText yields the strongest D2 clustering alignment.
The four pooled TFM variants reach D1 correlations between $-0.17$ and $0.04$, against $0.54$--$0.59$ for HashingText and HashingSchema, and trail the strongest lightweight methods on every D2 component; in the descriptive overall ranking they occupy the bottom four positions.

D2 metrics are label-dependent (Appendix~\ref{sec:app_results}). For \emph{Direct} identity labels, HashingText achieves the strongest triplet ranking under hard and cluster-hard negatives (Table~\ref{tab:results_synthetic_direct} in Appendix).
For the \emph{Stat} labeling scheme, HashingSchema is near-ceiling on both triplet and clustering metrics (Table~\ref{tab:results_synthetic_stat} in Appendix).
Increasing label granularity from \emph{Semantic} to \emph{Semantic+Difficulty} reduces hard-negative triplet accuracy across most schema-centric and statistical summary methods (Table~\ref{tab:results_synthetic_meta_3} in Appendix), while text/content-based methods remain comparatively stronger on strict triplet ranking.

For D3 robustness, most methods saturate on permutation robustness, with TableVectorizer as the exception due to its ordering sensitivity (Table~\ref{tab:results_synthetic}).
Noise robustness is most challenging for value-/content-sensitive methods (Figure~\ref{fig:metrics_bar_grid_synthetic}).
The weakness of pooled TFM states is therefore specific. Their robustness is high ($0.97$ for \textsc{ConTextTab} under every perturbation), but they rank lowest on D1 and among the lowest on D2.

\subsection{Real-data benchmark results}
\begin{figure*}[t]
\vspace{-0.5cm}
\centering
\small
\captionof{table}{Evaluation Results for Real data (mean $\pm$ 95\% CI over 10 seeds).}
\label{tab:results_real}
{\renewcommand{\arraystretch}{0.88}%
\resizebox{\textwidth}{!}{
\begin{tabular}{lcc|ccccc|cccc|c}
\toprule
\multirow{2}{*}{Embedder} & \multicolumn{2}{c}{D1 Consistency} & \multicolumn{5}{c}{D2 Label-Based - Avg} & \multicolumn{4}{c}{D3 Robustness} & \multicolumn{1}{c}{Overall} \\
\cmidrule(lr){2-3} \cmidrule(lr){4-8} \cmidrule(lr){9-12} \cmidrule(lr){13-13}
 & Spearman & Rank & TR-Avg & CL-Avg & R@5 & LP & Rank & Perm & Noise & Mask & Rank & Avg Rank \\
\midrule
HashingSchema & 0.31$\pm$0.03 & \cellcolor{green!40}2.00 & \textbf{0.59$\pm$0.03} & \textbf{0.40$\pm$0.03} & 0.96$\pm$0.01 & 0.84$\pm$0.02 & \cellcolor{green!60}2.25 & \textbf{1.00$\pm$0.00} & 0.94$\pm$0.01 & 0.95$\pm$0.01 & 8.83 & \cellcolor{green!40}4.36 \\
SchemaContent & 0.26$\pm$0.01 & \cellcolor{green!20}3.00 & 0.11$\pm$0.01 & 0.17$\pm$0.01 & 0.80$\pm$0.01 & \textbf{0.87$\pm$0.02} & 8.75 & \textbf{1.00$\pm$0.00} & \textbf{1.00$\pm$0.00} & \textbf{1.00$\pm$0.00} & \cellcolor{green!40}3.00 & \cellcolor{green!20}4.92 \\
TableStatistics & 0.21$\pm$0.02 & 5.00 & 0.08$\pm$0.01 & 0.24$\pm$0.01 & 0.76$\pm$0.01 & 0.59$\pm$0.02 & 9.25 & \textbf{1.00$\pm$0.00} & \textbf{1.00$\pm$0.00} & \textbf{1.00$\pm$0.00} & \cellcolor{green!60}2.83 & 5.69 \\
StatisticalSummary & 0.19$\pm$0.01 & 6.00 & 0.31$\pm$0.01 & 0.31$\pm$0.02 & 0.85$\pm$0.01 & 0.54$\pm$0.02 & 6.00 & \textbf{1.00$\pm$0.00} & \textbf{1.00$\pm$0.00} & \textbf{1.00$\pm$0.00} & 3.50 & 5.17 \\
MatrixFactorization & 0.16$\pm$0.02 & 8.00 & 0.21$\pm$0.00 & 0.20$\pm$0.01 & 0.65$\pm$0.01 & 0.28$\pm$0.02 & 10.75 & \textbf{1.00$\pm$0.00} & \textbf{1.00$\pm$0.00} & \textbf{1.00$\pm$0.00} & \cellcolor{green!20}3.17 & 7.31 \\
HashingText & \textbf{0.41$\pm$0.03} & \cellcolor{green!60}1.00 & 0.51$\pm$0.02 & 0.28$\pm$0.03 & \textbf{0.98$\pm$0.00} & \textbf{0.87$\pm$0.02} & \cellcolor{green!20}2.50 & \textbf{1.00$\pm$0.00} & 0.73$\pm$0.02 & 0.98$\pm$0.00 & 7.50 & \cellcolor{green!60}3.67 \\
LLMText & 0.13$\pm$0.03 & 10.00 & 0.55$\pm$0.02 & 0.40$\pm$0.04 & \textbf{0.97$\pm$0.00} & 0.86$\pm$0.02 & \cellcolor{green!40}2.25 & 0.91$\pm$0.01 & 0.95$\pm$0.00 & 0.99$\pm$0.00 & 8.67 & 6.97 \\
TableVectorizer & 0.03$\pm$0.03 & \cellcolor{red!40}13.00 & 0.18$\pm$0.01 & 0.17$\pm$0.01 & 0.58$\pm$0.01 & 0.29$\pm$0.03 & \cellcolor{red!60}12.25 & 0.33$\pm$0.01 & 0.98$\pm$0.00 & 0.85$\pm$0.01 & \cellcolor{red!60}10.67 & \cellcolor{red!60}11.97 \\
HyTREL & 0.15$\pm$0.03 & 9.00 & 0.25$\pm$0.01 & 0.21$\pm$0.01 & 0.85$\pm$0.01 & 0.76$\pm$0.02 & 6.50 & \textbf{0.99$\pm$0.00} & 0.89$\pm$0.01 & 0.96$\pm$0.00 & \cellcolor{red!20}10.33 & 8.61 \\
Armadillo & 0.25$\pm$0.03 & 4.00 & 0.35$\pm$0.01 & 0.21$\pm$0.01 & 0.86$\pm$0.02 & 0.65$\pm$0.03 & 5.75 & \textbf{1.00$\pm$0.00} & 0.60$\pm$0.02 & 0.78$\pm$0.01 & \cellcolor{red!40}10.50 & 6.75 \\
TabPFN (mean) & 0.06$\pm$0.04 & \cellcolor{red!20}12.00 & 0.13$\pm$0.01 & 0.20$\pm$0.01 & 0.60$\pm$0.02 & 0.42$\pm$0.02 & \cellcolor{red!20}11.50 & 0.79$\pm$0.01 & 0.98$\pm$0.00 & 0.97$\pm$0.00 & 9.00 & \cellcolor{red!20}10.83 \\
TabPFN (max) & 0.07$\pm$0.04 & 11.00 & 0.16$\pm$0.01 & 0.18$\pm$0.01 & 0.60$\pm$0.01 & 0.42$\pm$0.02 & \cellcolor{red!40}11.75 & 0.78$\pm$0.01 & 0.97$\pm$0.01 & 0.97$\pm$0.00 & 10.33 & \cellcolor{red!40}11.03 \\
\textsc{ConTextTab} (mean) & 0.19$\pm$0.03 & 7.00 & 0.25$\pm$0.01 & 0.22$\pm$0.01 & 0.76$\pm$0.01 & 0.70$\pm$0.02 & 6.75 & 0.97$\pm$0.00 & 0.97$\pm$0.00 & 0.97$\pm$0.00 & 9.33 & 7.69 \\
\textsc{ConTextTab} (max) & -0.06$\pm$0.03 & \cellcolor{red!60}14.00 & 0.14$\pm$0.01 & 0.23$\pm$0.02 & 0.72$\pm$0.02 & 0.64$\pm$0.02 & 8.75 & 0.98$\pm$0.00 & 0.98$\pm$0.00 & 0.98$\pm$0.00 & 7.33 & 10.03 \\
\bottomrule
\end{tabular}
}}

\vspace{0.35em}

\centerline{\includegraphics[width=0.88\textwidth]{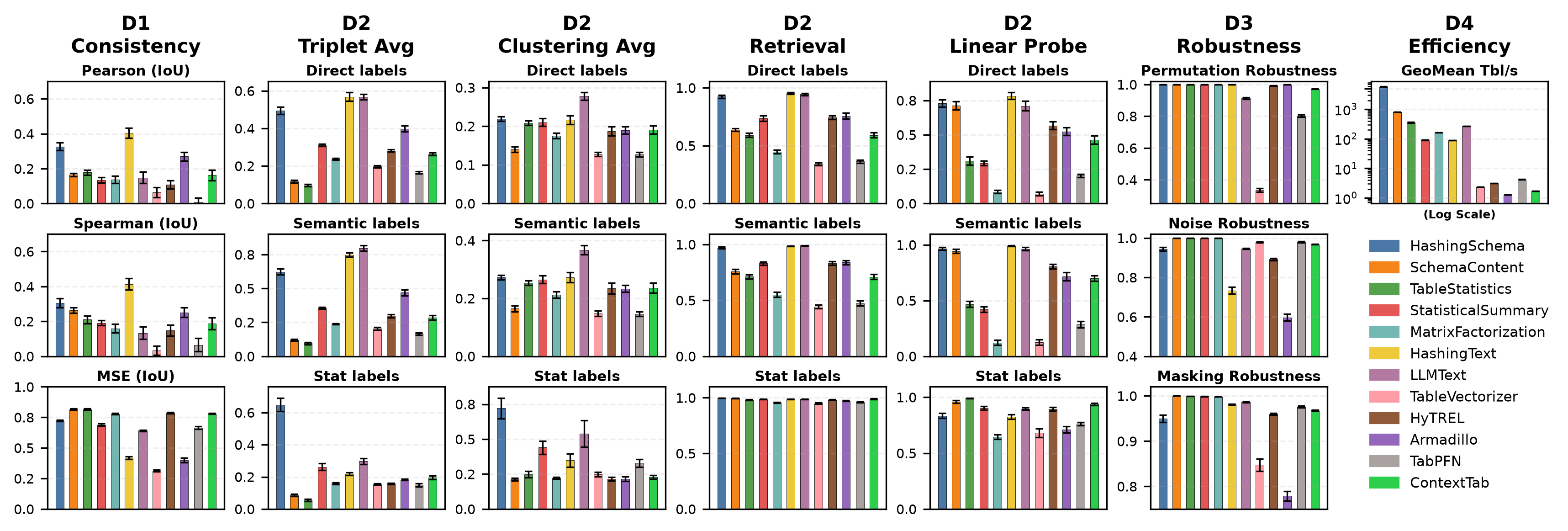}}
\captionof{figure}{Metric summary for Real data across the benchmark dimensions.}
\vspace{-0.5cm}
\label{fig:metrics_bar_grid_real}
\end{figure*}

On real open-source data (Table~\ref{tab:results_real}, Figure~\ref{fig:metrics_bar_grid_real}), 
lightweight hashing and statistical-summary embedding models remain consistently among the top-performing methods.
HashingText ranks best overall, combining the strongest D1 consistency with strong D2 retrieval performance, while HashingSchema follows closely, reflecting the continued importance of schema/type cues in heterogeneous benchmark suites.
LLMText achieves strong D2 clustering and retrieval but ranks poorly on D1 (Table~\ref{tab:results_real}). The synthetic pattern carries over, with schema- and text-based methods leading and pooled TFM states lagging on D1 and D2.

The per-label D2 breakdown (Appendix~\ref{sec:app_results}) shows that the \emph{Direct} labeling scheme is substantially harder on real tables than on synthetic data, with low hard-negative triplet scores across most methods (Table~\ref{tab:results_real_direct} in Appendix) even when retrieval remains high.
\emph{Stat} labels are the easiest and align strongly with schema cues (Table~\ref{tab:results_real_stat} in Appendix), while \emph{Semantic} labels score in between identity and coarse type groupings (Table~\ref{tab:results_real_meta_2} in Appendix).

As in synthetic data, D3 permutation robustness is near-saturated for most methods except TableVectorizer (Table~\ref{tab:results_real}).
Noise robustness again differentiates value-/content-sensitive methods (Figure~\ref{fig:metrics_bar_grid_real}).
Pooled TFM states again trail on D1 and D2, with \textsc{ConTextTab} above TabPFN. Under mean pooling, \textsc{ConTextTab} reaches a D1 of $0.19$ and TabPFN $0.06$, against $0.41$ for HashingText.

\subsection{Is the result specific to mean pooling?}

\label{sec:pooling_main}

\begin{table}[!t]
\centering
\small
\caption{Five parameter-free pooling rules applied to the same TabPFN and \textsc{ConTextTab} states (Appendix~\ref{app:pooling}), with the two strongest fingerprints for reference (mean $\pm$ 95\% CI over 10 seeds). Mean and max rows match Tables~\ref{tab:results_synthetic} and~\ref{tab:results_real}. Bold marks the best value per column within each group.}
\label{tab:pooling_main}
{\renewcommand{\arraystretch}{0.9}%
\resizebox{\textwidth}{!}{
\begin{tabular}{l|cccc|cccc}
\toprule
\multirow{2}{*}{Embedder} & \multicolumn{4}{c|}{Synthetic} & \multicolumn{4}{c}{Real} \\
\cmidrule(lr){2-5} \cmidrule(lr){6-9}
 & D1 Spearman & TR-Avg & CL-Avg & R@5 & D1 Spearman & TR-Avg & CL-Avg & R@5 \\
\midrule
HashingSchema & \textbf{0.59$\pm$0.01} & \textbf{0.53$\pm$0.01} & \textbf{0.54$\pm$0.02} & 0.71$\pm$0.01 & 0.31$\pm$0.03 & \textbf{0.59$\pm$0.03} & \textbf{0.40$\pm$0.03} & 0.96$\pm$0.01 \\
HashingText & 0.54$\pm$0.04 & 0.47$\pm$0.01 & 0.46$\pm$0.03 & \textbf{1.00$\pm$0.00} & \textbf{0.41$\pm$0.03} & 0.51$\pm$0.02 & 0.28$\pm$0.03 & \textbf{0.98$\pm$0.00} \\
\midrule
TabPFN (mean) & -0.08$\pm$0.05 & 0.17$\pm$0.00 & \textbf{0.25$\pm$0.02} & \textbf{0.73$\pm$0.01} & 0.06$\pm$0.04 & 0.13$\pm$0.01 & \textbf{0.20$\pm$0.01} & \textbf{0.60$\pm$0.02} \\
TabPFN (max) & -0.17$\pm$0.04 & \textbf{0.21$\pm$0.01} & \textbf{0.25$\pm$0.01} & 0.71$\pm$0.01 & \textbf{0.07$\pm$0.04} & 0.16$\pm$0.01 & 0.18$\pm$0.01 & \textbf{0.60$\pm$0.01} \\
TabPFN (sum) & -0.08$\pm$0.05 & 0.17$\pm$0.00 & \textbf{0.25$\pm$0.02} & \textbf{0.73$\pm$0.01} & 0.06$\pm$0.04 & 0.13$\pm$0.01 & \textbf{0.20$\pm$0.01} & \textbf{0.60$\pm$0.02} \\
TabPFN (min) & -0.15$\pm$0.04 & \textbf{0.21$\pm$0.00} & \textbf{0.25$\pm$0.01} & 0.71$\pm$0.01 & 0.06$\pm$0.03 & 0.16$\pm$0.01 & 0.19$\pm$0.01 & 0.57$\pm$0.01 \\
TabPFN (std) & \textbf{0.15$\pm$0.04} & \textbf{0.21$\pm$0.00} & \textbf{0.25$\pm$0.01} & 0.70$\pm$0.01 & 0.03$\pm$0.03 & \textbf{0.20$\pm$0.00} & 0.18$\pm$0.01 & 0.54$\pm$0.01 \\
\midrule
\textsc{ConTextTab} (mean) & 0.04$\pm$0.03 & \textbf{0.18$\pm$0.00} & 0.27$\pm$0.02 & 0.70$\pm$0.01 & \textbf{0.19$\pm$0.03} & \textbf{0.25$\pm$0.01} & 0.22$\pm$0.01 & 0.76$\pm$0.01 \\
\textsc{ConTextTab} (max) & 0.03$\pm$0.04 & 0.08$\pm$0.01 & 0.26$\pm$0.02 & 0.68$\pm$0.01 & -0.06$\pm$0.03 & 0.14$\pm$0.01 & 0.23$\pm$0.02 & 0.72$\pm$0.02 \\
\textsc{ConTextTab} (sum) & 0.04$\pm$0.03 & \textbf{0.18$\pm$0.00} & 0.27$\pm$0.02 & 0.70$\pm$0.01 & \textbf{0.19$\pm$0.03} & \textbf{0.25$\pm$0.01} & 0.22$\pm$0.01 & 0.76$\pm$0.01 \\
\textsc{ConTextTab} (min) & 0.02$\pm$0.03 & 0.09$\pm$0.00 & 0.24$\pm$0.01 & 0.66$\pm$0.01 & -0.08$\pm$0.03 & 0.15$\pm$0.01 & 0.22$\pm$0.01 & 0.69$\pm$0.01 \\
\textsc{ConTextTab} (std) & \textbf{0.07$\pm$0.04} & 0.01$\pm$0.00 & \textbf{0.28$\pm$0.03} & \textbf{0.71$\pm$0.01} & -0.05$\pm$0.03 & 0.03$\pm$0.01 & \textbf{0.27$\pm$0.03} & \textbf{0.77$\pm$0.01} \\
\bottomrule
\end{tabular}
}}
\end{table}

\begin{revised}
Mean pooling is a natural default, but a weak result under averaging could be specific to that rule. We therefore compute five parameter-free rules (mean, max, sum, min, and std) from a single forward pass of each TFM, so that the encoder, inputs, partitions, perturbations, and seeds are identical and only the final reduction differs (Appendix~\ref{app:pooling}). Changing the rule moves individual metrics but not the conclusion (Table~\ref{tab:pooling_main}). Here and in Section~\ref{sec:frozen_ctt}, which compare compositional rules over the same states, we omit the linear probe, because it trains its own classifier on top of each embedding and would add a second learned stage to the comparison. No rule brings either TFM close to the strongest fingerprints, and under mean and max, the two rules ranked in the main tables, both TFMs stay below every schema-, text-, and statistics-based fingerprint in the overall average rank. Sum coincides with mean on all cosine-based metrics, as expected from the scale invariance of cosine similarity. The largest metric-level change is TabPFN's synthetic D1, which rises from $-0.08$ under mean to $0.15$ under std, still far below the strongest fingerprints.

A second concern is that the benchmark could reward exact column-name overlap. Renaming all columns positionally before extraction (Appendix~\ref{app:column_anonymization}) shows that schema-only methods depend on headers, as the synthetic D1 of HashingSchema drops by $0.29$. HashingText, which also sees cell values, changes by at most $0.05$ in D1 and $0.01$ in D2 average, so the fingerprints' lead does not rest on header matching alone.

The evidence in this section concerns TFM states composed by fixed rules and compared by cosine similarity. It shows that this interface yields weak table-level embeddings; it does not show whether the encoder states contain the missing information. Section~\ref{sec:frozen_ctt} tests this directly.
\end{revised}

\section{Can a Learned Compositional Rule over Frozen States Do Better?}
\label{sec:frozen_ctt}

\begin{figure*}[t]
    \centering
    \vspace{-1.0cm}
    \includegraphics[width=0.72\linewidth]{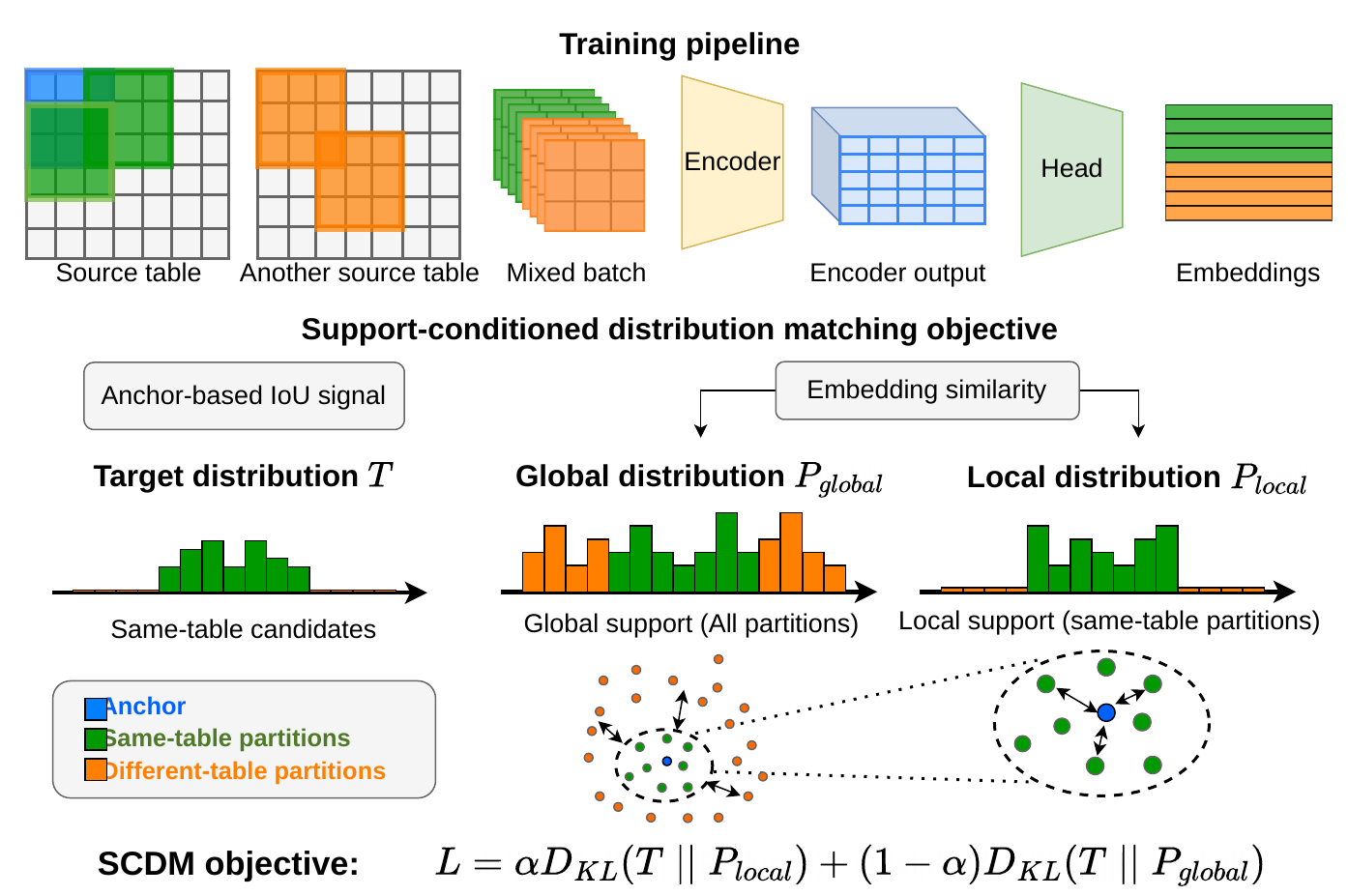}
    \caption{Learning table-level composition on top of frozen \textsc{ConTextTab}. At the top, partitions sampled from one source table and from other source tables are collated into mixed batches, encoded by the frozen backbone, and aggregated by a trainable representation head into table embeddings. At the bottom, support-conditioned distribution matching defines a target distribution $T$ from anchor-based overlap and matches it with local and global similarity distributions induced by the learned embeddings.}
    \label{fig:frozen_ctt_case_study}
    \vspace{-10pt}
\end{figure*}

Section~\ref{sec:results} leaves two explanations for the weak pooled embeddings, namely that the encoder states lack the information needed for D1 and D2 or that they contain it in a form that fixed pooling does not expose. Freezing the encoder and learning only the compositional rule distinguishes these cases. If a learned head $C_\theta$ improves the table-level embeddings while $H(T)$ is unchanged, the frozen states contain information that the learned interface can exploit for these desiderata.

We choose \textsc{ConTextTab} because it is a semantics-aware TFM that explicitly processes column titles, which matters given the strength of schema-sensitive baselines in Section~\ref{sec:results}, and because it is the stronger of the two TFMs under mean pooling on both corpora. Freezing it ensures that any gain comes from the head rather than from additional encoder training. The heads are not intended as a competing table encoder. They never see the raw table, only the frozen states, so their natural baseline is fixed composition of the same states (Section~\ref{sec:pooling_main}). The head, however, changes two things at once, the aggregation architecture and the training objective, which is aligned with D1 and D2. The experiment therefore shows what a learned compositional rule with aligned supervision can extract from frozen states; it does not separate the two.

For each sampled partition $P$, the frozen encoder produces contextual cell states $H(P)\in\mathbb{R}^{n\times m\times d}$, and a trainable head maps them to a table-level embedding $z(P)=C_\theta\bigl(H(P)\bigr)\in\mathbb{R}^{d}$. We study three head families, \textbf{Attention}, \textbf{GatedAttention}, and \textbf{GatedMLP}, each of which pools cells into rows and rows into a table with learned weights. We write \textsc{ConTextTab}-R (for \emph{Representation}) for the frozen \textsc{ConTextTab} encoder combined with a learned head, e.g.\ \textsc{ConTextTab}-R Attention.

Figure~\ref{fig:frozen_ctt_case_study} summarizes the setup, including the loss we propose. 
For an anchor partition $A$, let $\mathcal{B}$ be the other partitions in the mixed batch and $\mathcal{S}\subseteq\mathcal{B}$ those from the same source table. The loss compares an overlap-based target $T$, defined on $\mathcal{S}$ and zero elsewhere, with two distributions of the similarities $s(Q)=\cos\bigl(z(A),z(Q)\bigr)/\tau$, $\tau=0.07$, normalized over the same-table candidates (local support) or over the full batch (global support),
\begin{equation}
T(Q)=\frac{\mathrm{IoU}(A,Q)}{\sum_{R\in\mathcal{S}}\mathrm{IoU}(A,R)},
\qquad
P_{\mathrm{local}}(Q)=\frac{\exp s(Q)}{\sum_{R\in\mathcal{S}}\exp s(R)},
\qquad
P_{\mathrm{global}}(Q)=\frac{\exp s(Q)}{\sum_{R\in\mathcal{B}}\exp s(R)}.
\end{equation}
We call the resulting objective support-conditioned distribution matching (SCDM),
\begin{equation}
\mathcal{L} = \alpha\, D_{\mathrm{KL}}\left(T \,\|\, P_{\mathrm{local}}\right)
+ (1-\alpha)\, D_{\mathrm{KL}}\left(T \,\|\, P_{\mathrm{global}}\right),
\end{equation}
where $\alpha$ controls the trade-off between fine within-table overlap structure and broader cross-table discrimination.
The loss differs from the D1 Spearman metric and is contrastive only implicitly, through the global normalization. The heads are trained on real tables only and with a different partition sampler than the benchmark, so the synthetic results measure transfer to unseen table families (Appendix~\ref{sec:app_head_training_details}).

\begin{revised}
Each learned head enters the benchmark as a single fixed model, in the same way as every other method. For each head family and value of $\alpha$, we train three models with different seeds, score each on a separate validation split of $20$ real tables (the mean of D1 Spearman and the averaged D2 metrics), and keep the checkpoint with the highest validation score (Appendix~\ref{sec:app_head_training_details}). This checkpoint is evaluated on the benchmark with the same $10$ sampler seeds as all other methods, so its confidence intervals, like theirs, reflect evaluation sampling rather than variation across training runs. We evaluate each head family at five values of $\alpha$ and state the conclusions over this full sweep (Appendix~\ref{sec:app_head_ablation}); Table~\ref{tab:head_ablation_compact_combined} summarizes it with one configuration per head family and corpus, the one with the lowest D1--D3 average rank. On synthetic data, all $15$ configurations (three heads, five values of $\alpha$) improve over pooled \textsc{ConTextTab} on D1 ($0.43$--$0.50$ vs.\ $0.04$) and on triplet ranking, clustering, and retrieval. On real data, all $15$ improve clustering and Recall@5, and $14$ of $15$ improve D1 ($0.21$--$0.27$ vs.\ $0.19$) and triplet ranking; the exception, Attention with $\alpha=1$, does not improve D1 ($0.17$) or triplet ranking ($0.24$ vs.\ $0.25$).
\end{revised}

The learned heads improve the table-level embeddings substantially although the encoder is unchanged (Table~\ref{tab:head_ablation_compact_combined}). On synthetic data, D1 rises from $0.04$ to $0.48$--$0.50$ and retrieval from $0.70$ to $0.88$--$0.89$; on real data, the margin is smaller but systematic, with D1 rising from $0.19$ to $0.24$--$0.27$ and retrieval from $0.76$ to $0.95$--$0.96$. Triplet ranking and clustering also improve on both corpora. The heads close most of the D1 gap to the fingerprints on synthetic data but less on real data, and they keep D3 at $0.98$--$1.00$, on par with the most robust fingerprints, whereas HashingText loses robustness to noise ($0.65$ and $0.73$). The leaderboard recomputed with the heads (Appendix~\ref{sec:app_augmented_leaderboard}) ranks \textsc{ConTextTab}-R Attention first on synthetic and second on real data, behind HashingText; pooled \textsc{ConTextTab} ranks 14th and 12th.

\begin{table*}[t]
\centering
\small
\caption{Learned compositional rules on frozen \textsc{ConTextTab} states, compared with fixed pooling of the same states and with the two strongest fingerprints (mean $\pm$ 95\% CI over 10 seeds). Each head is the validation-selected checkpoint at the $\alpha$ with the lowest D1--D3 average rank (all $\alpha$ in Appendix~\ref{sec:app_head_ablation}). Bold marks the best value per column within each group; italic rows give the best head's gain over pooling.}
\label{tab:head_ablation_compact_combined}
{\renewcommand{\arraystretch}{0.88}%
\resizebox{\textwidth}{!}{
\begin{tabular}{llc|c|ccc|ccc}
\toprule
\multirow{2}{*}{\textbf{Data}} & \multirow{2}{*}{\textbf{Embedder}} & \multirow{2}{*}{$\alpha$} & \multicolumn{1}{c|}{D1 Consistency} & \multicolumn{3}{c|}{D2 Label-Based - Avg} & \multicolumn{3}{c}{D3 Robustness} \\
\cmidrule(lr){4-4} \cmidrule(lr){5-7} \cmidrule(lr){8-10}
 & & & Spearman & TR-Avg & CL-Avg & R@5 & Perm & Noise & Mask \\
\midrule
\multirow{8}{*}{Synthetic} & HashingSchema & -- & \textbf{0.59$\pm$0.01} & \textbf{0.53$\pm$0.01} & \textbf{0.54$\pm$0.02} & 0.71$\pm$0.01 & \textbf{1.00$\pm$0.00} & \textbf{0.98$\pm$0.00} & \textbf{0.99$\pm$0.00} \\
& HashingText & -- & 0.54$\pm$0.04 & 0.47$\pm$0.01 & 0.46$\pm$0.03 & \textbf{1.00$\pm$0.00} & \textbf{1.00$\pm$0.00} & 0.65$\pm$0.02 & 0.97$\pm$0.00 \\
\cmidrule(lr){2-10}
& \textsc{ConTextTab} (pooled) & -- & 0.04$\pm$0.03 & 0.18$\pm$0.00 & 0.27$\pm$0.02 & 0.70$\pm$0.01 & 0.97$\pm$0.00 & 0.97$\pm$0.00 & 0.97$\pm$0.00 \\
\cmidrule(lr){2-10}
& \textsc{ConTextTab}-R Attention & 0.75 & \textbf{0.50$\pm$0.02} & \textbf{0.34$\pm$0.01} & \textbf{0.46$\pm$0.02} & 0.88$\pm$0.01 & \textbf{1.00$\pm$0.00} & \textbf{1.00$\pm$0.00} & \textbf{0.99$\pm$0.00} \\
& \textsc{ConTextTab}-R GatedAttention & 1.00 & 0.48$\pm$0.02 & 0.31$\pm$0.01 & \textbf{0.46$\pm$0.03} & \textbf{0.89$\pm$0.01} & \textbf{1.00$\pm$0.00} & \textbf{1.00$\pm$0.00} & \textbf{0.99$\pm$0.00} \\
& \textsc{ConTextTab}-R GatedMLP & 0.75 & 0.48$\pm$0.02 & 0.30$\pm$0.01 & 0.43$\pm$0.04 & 0.88$\pm$0.01 & \textbf{1.00$\pm$0.00} & \textbf{1.00$\pm$0.00} & \textbf{0.99$\pm$0.00} \\
& \emph{Best head $-$ pooled} & & \emph{+0.46} & \emph{+0.16} & \emph{+0.19} & \emph{+0.19} & \emph{+0.03} & \emph{+0.03} & \emph{+0.02} \\
\midrule
\multirow{8}{*}{Real} & HashingSchema & -- & 0.31$\pm$0.03 & \textbf{0.59$\pm$0.03} & \textbf{0.40$\pm$0.03} & 0.96$\pm$0.01 & \textbf{1.00$\pm$0.00} & \textbf{0.94$\pm$0.01} & 0.95$\pm$0.01 \\
& HashingText & -- & \textbf{0.41$\pm$0.03} & 0.51$\pm$0.02 & 0.28$\pm$0.03 & \textbf{0.98$\pm$0.00} & \textbf{1.00$\pm$0.00} & 0.73$\pm$0.02 & \textbf{0.98$\pm$0.00} \\
\cmidrule(lr){2-10}
& \textsc{ConTextTab} (pooled) & -- & 0.19$\pm$0.03 & 0.25$\pm$0.01 & 0.22$\pm$0.01 & 0.76$\pm$0.01 & 0.97$\pm$0.00 & 0.97$\pm$0.00 & 0.97$\pm$0.00 \\
\cmidrule(lr){2-10}
& \textsc{ConTextTab}-R Attention & 0.25 & \textbf{0.27$\pm$0.03} & \textbf{0.41$\pm$0.02} & \textbf{0.30$\pm$0.03} & \textbf{0.96$\pm$0.01} & \textbf{1.00$\pm$0.00} & 0.99$\pm$0.00 & 0.98$\pm$0.00 \\
& \textsc{ConTextTab}-R GatedAttention & 0.50 & 0.26$\pm$0.03 & 0.38$\pm$0.02 & 0.27$\pm$0.02 & 0.95$\pm$0.01 & \textbf{1.00$\pm$0.00} & 0.99$\pm$0.00 & \textbf{0.99$\pm$0.00} \\
& \textsc{ConTextTab}-R GatedMLP & 1.00 & 0.24$\pm$0.03 & 0.31$\pm$0.02 & 0.27$\pm$0.03 & 0.95$\pm$0.01 & \textbf{1.00$\pm$0.00} & \textbf{1.00$\pm$0.00} & \textbf{0.99$\pm$0.00} \\
& \emph{Best head $-$ pooled} & & \emph{+0.08} & \emph{+0.16} & \emph{+0.08} & \emph{+0.20} & \emph{+0.03} & \emph{+0.03} & \emph{+0.02} \\
\bottomrule
\end{tabular}
}}
\vspace{-0.5cm}
\end{table*}

\begin{revised}
A learned compositional rule therefore recovers from the frozen states much of the partial-view consistency and discriminability that fixed pooling leaves unused. The weak pooled \textsc{ConTextTab} embeddings of Section~\ref{sec:results} should not be read as evidence that the encoder states lack this information, because the same frozen states, composed by a learned head, move from near the bottom of the benchmark into its top group. Because the encoder is frozen, the head can use only information exposed in those states. The remaining gap to the strongest fingerprints on real data may reflect the backbone, head capacity, the training objective, or the shift between training and evaluation partitions, and the experiment does not identify which. The heads add little cost, and their throughput stays within $5\%$ of pooled \textsc{ConTextTab} (Appendix~\ref{sec:app_efficiency}).
\end{revised}

\needspace{6\baselineskip}
\section{Do the Benchmark Findings Hold in Downstream Retrieval?}
\label{sec:downstream}

A downstream parent-table retrieval task tests whether the benchmark findings hold in a concrete operation, recovering the source table of a partial view. Each query is a sampled partition, the candidate corpus contains $100$ source tables, and each embedder is evaluated on $1000$ queries; we report Recall@1, Recall@5, Recall@10, and Mean Reciprocal Rank (MRR). Success requires partial views to stay close to their source table (D1) while being separated from other tables, which corresponds to the Direct-label component of D2. The task is a controlled closed-corpus task, narrower than table union search~\citep{nargesian2018tus}.

\begin{table*}[t]
\centering
\small
\caption{Parent-table retrieval on synthetic and real data (mean $\pm$ 95\% CI over 10 seeds).}
\label{tab:downstream_parent_table_retrieval}
{\renewcommand{\arraystretch}{0.9}%
\resizebox{\textwidth}{!}{
\begin{tabular}{lcccc|cccc}
\toprule
\multirow{2}{*}{Embedder} & \multicolumn{4}{c|}{Synthetic} & \multicolumn{4}{c}{Real} \\
\cmidrule(lr){2-5} \cmidrule(lr){6-9}
 & R@1 & R@5 & R@10 & MRR & R@1 & R@5 & R@10 & MRR \\
\midrule
HashingSchema & 0.10$\pm$0.00 & 0.47$\pm$0.01 & 0.78$\pm$0.02 & 0.28$\pm$0.00 & \textbf{0.67$\pm$0.02} & \textbf{0.97$\pm$0.01} & 0.98$\pm$0.00 & \textbf{0.80$\pm$0.01} \\
SchemaContent & 0.01$\pm$0.00 & 0.05$\pm$0.00 & 0.10$\pm$0.00 & 0.05$\pm$0.00 & 0.10$\pm$0.01 & 0.28$\pm$0.02 & 0.38$\pm$0.02 & 0.20$\pm$0.01 \\
TableStatistics & 0.04$\pm$0.00 & 0.19$\pm$0.01 & 0.33$\pm$0.01 & 0.13$\pm$0.00 & 0.04$\pm$0.01 & 0.16$\pm$0.01 & 0.25$\pm$0.02 & 0.11$\pm$0.01 \\
StatisticalSummary & 0.32$\pm$0.02 & 0.69$\pm$0.02 & 0.82$\pm$0.02 & 0.48$\pm$0.02 & 0.25$\pm$0.02 & 0.51$\pm$0.03 & 0.63$\pm$0.03 & 0.38$\pm$0.02 \\
MatrixFactorization & 0.19$\pm$0.01 & 0.49$\pm$0.03 & 0.61$\pm$0.04 & 0.33$\pm$0.02 & 0.10$\pm$0.01 & 0.27$\pm$0.02 & 0.38$\pm$0.03 & 0.19$\pm$0.01 \\
HashingText & \textbf{0.98$\pm$0.01} & \textbf{0.99$\pm$0.00} & \textbf{1.00$\pm$0.00} & \textbf{0.99$\pm$0.00} & 0.66$\pm$0.02 & \textbf{0.97$\pm$0.01} & 0.98$\pm$0.01 & \textbf{0.80$\pm$0.01} \\
LLMText & 0.24$\pm$0.01 & 0.64$\pm$0.02 & 0.85$\pm$0.02 & 0.42$\pm$0.01 & 0.61$\pm$0.03 & \textbf{0.97$\pm$0.01} & \textbf{0.99$\pm$0.00} & 0.76$\pm$0.02 \\
TableVectorizer & 0.06$\pm$0.01 & 0.16$\pm$0.01 & 0.22$\pm$0.02 & 0.13$\pm$0.01 & 0.03$\pm$0.01 & 0.12$\pm$0.01 & 0.19$\pm$0.01 & 0.09$\pm$0.01 \\
HyTREL & 0.15$\pm$0.01 & 0.40$\pm$0.01 & 0.57$\pm$0.02 & 0.28$\pm$0.01 & 0.29$\pm$0.01 & 0.62$\pm$0.02 & 0.73$\pm$0.02 & 0.44$\pm$0.01 \\
Armadillo & 0.63$\pm$0.03 & 0.81$\pm$0.02 & 0.87$\pm$0.02 & 0.71$\pm$0.02 & 0.45$\pm$0.01 & 0.77$\pm$0.01 & 0.84$\pm$0.01 & 0.60$\pm$0.01 \\
TabPFN & 0.04$\pm$0.00 & 0.13$\pm$0.00 & 0.21$\pm$0.01 & 0.10$\pm$0.00 & 0.03$\pm$0.01 & 0.11$\pm$0.01 & 0.19$\pm$0.01 & 0.09$\pm$0.01 \\
\textsc{ConTextTab} & 0.05$\pm$0.00 & 0.21$\pm$0.01 & 0.35$\pm$0.01 & 0.14$\pm$0.01 & 0.10$\pm$0.01 & 0.27$\pm$0.02 & 0.39$\pm$0.02 & 0.19$\pm$0.01 \\
\midrule
\multicolumn{9}{l}{\emph{Learned heads on frozen \textsc{ConTextTab}}} \\
\midrule
\textsc{ConTextTab}-R Attention & 0.20$\pm$0.01 & 0.50$\pm$0.02 & 0.65$\pm$0.02 & 0.34$\pm$0.01 & 0.35$\pm$0.01 & 0.70$\pm$0.03 & 0.81$\pm$0.02 & 0.51$\pm$0.01 \\
\textsc{ConTextTab}-R GatedAttention & 0.16$\pm$0.01 & 0.45$\pm$0.01 & 0.63$\pm$0.01 & 0.30$\pm$0.01 & 0.35$\pm$0.01 & 0.74$\pm$0.02 & 0.86$\pm$0.02 & 0.52$\pm$0.01 \\
\textsc{ConTextTab}-R GatedMLP & 0.13$\pm$0.01 & 0.39$\pm$0.01 & 0.55$\pm$0.02 & 0.26$\pm$0.01 & 0.29$\pm$0.01 & 0.63$\pm$0.02 & 0.75$\pm$0.03 & 0.45$\pm$0.01 \\
\bottomrule
\end{tabular}
}}
\end{table*}

Table~\ref{tab:downstream_parent_table_retrieval} shows that the retrieval results mirror the benchmark. Across the twelve benchmark methods, the D2 ranking of Tables~\ref{tab:results_synthetic} and~\ref{tab:results_real} agrees closely with the ranking by retrieval MRR (Spearman correlation $0.82$ on synthetic and $0.95$ on real data), and the overall average rank, which also includes D1 and D3, gives $0.62$ and $0.68$. These correlations are descriptive, computed over twelve methods, and partly expected, because D2 includes the Direct-label component that retrieval shares by design. On synthetic data, \textbf{HashingText} is dominant with near-perfect Recall@1 and MRR, while \textbf{Armadillo} is the strongest non-text baseline. On real data, \textbf{HashingSchema}, \textbf{HashingText}, and \textbf{LLMText} form the top group.

The learned heads improve markedly over pooled \textsc{ConTextTab}. On synthetic data, \textsc{ConTextTab}-R Attention reaches a Recall@1 of $0.20$ and an MRR of $0.34$, compared with $0.05$ and $0.14$ for pooled \textsc{ConTextTab}; on real data, \textsc{ConTextTab}-R GatedAttention reaches $0.35$ and $0.52$, compared with $0.10$ and $0.19$. SCDM is not trained on the retrieval metric, although it uses overlap and same-table identity signals that are closely related to source recovery; for the learned heads this part of the task is therefore a consistency check rather than an independent test, whereas the comparison among the twelve benchmark methods involves no training on the task. The heads do not close the gap to the strongest fingerprints, which reach a Recall@1 of $0.98$ (HashingText, synthetic) and $0.67$ (HashingSchema, real).

Parent-table retrieval thus isolates the table-selection step of tabular RAG, not the full pipeline.

\section{Discussion}
\label{sec:discussion}

\begin{revised}
\paragraph{What the benchmark diagnoses.} The benchmark evaluates the table-level interface formed by an encoder's local states, a compositional rule, and cosine similarity. For the two TFMs we study, fixed, parameter-free pooling over this interface yields table-level embeddings that trail lightweight fingerprints on D1 and D2, whereas a head trained with our proposed loss SCDM over the same frozen \textsc{ConTextTab} states substantially improves D1, D2, and source recovery. The weak pooled scores therefore cannot be attributed to the encoder alone, and the results do not point to a conflict between predictive training and table-level representation. They show instead that a useful table-level embedding does not follow automatically from predictive states and a default pooling rule, and whether learned composition closes the remaining gap for other encoders is open.
\end{revised}

\begin{revised}
\paragraph{Applying the benchmark to future TFMs.} Any model that exposes row- or cell-level states can be evaluated unchanged. The model supplies the states, and the harness scores them under each fixed pooling rule and, optionally, under a learned head trained with the protocol of Section~\ref{sec:frozen_ctt}. Reporting both settings separates what a model's states contain from what a default pooling rule exposes, and a large gain under the learned head indicates that the fixed rule limits the pooled embedding. The released code implements this through a shared \texttt{encode\_tables} interface (Appendix~\ref{app:code_implementation}). We apply it to two TFMs here, and the frozen-encoder protocol for TabPFN and newer TFMs is the natural next test.
\end{revised}

\paragraph{Why do simple fingerprints remain strong?} In many corpus-scale operations, schema and lexical cues are not spurious shortcuts but part of the operational definition of similarity. Column names, type patterns, and recognizable value tokens often provide exactly the information needed for indexing, candidate generation, source recovery, and approximate deduplication, which is why we retain schema metadata in the benchmark. The per-label D2 breakdown shows where these cues stop being enough, for example \emph{Direct} identity ranking under hard negatives or \emph{Semantic+Difficulty} labels, which require value-sensitive signals. The column-name anonymization ablation separates the two, as schema-only methods degrade when headers are removed while HashingText and name-light statistical controls remain stable. D1 is an overlap-oriented criterion rather than a universal definition of table similarity, so the lightweight-baseline conclusion also rests on the label-based D2 metrics, robustness, efficiency, and the anonymization ablation.

\paragraph{What the learned heads suggest for model design.} Because the frozen-encoder experiment changes the aggregation architecture and its objective together (Section~\ref{sec:frozen_ctt}), it identifies the compositional rule as a place where table-level embedding quality is gained, but not which ingredient matters most. Training the same heads with an objective unrelated to the desiderata, or training a simple fixed-architecture aggregator with the SCDM objective, would separate the two. A further direction is to include a table-level objective during TFM pretraining, so that the encoder is encouraged to produce states that compose well; the benchmark provides a way to measure whether such training helps beyond what a post-hoc head achieves.

\paragraph{Limitations.} The benchmark covers single-table representations compared by cosine similarity under a shared observation budget. Other similarity functions or learned metrics may change the ranking. Our fixed-composition findings concern two TFMs, and other TFMs may yield more or less useful table-level embeddings under fixed pooling. The learned-composition experiment uses a single backbone so that the compositional rule is the only trained component. Parent-table retrieval uses a closed corpus of $100$ tables to keep the target unambiguous, so open-world retrieval and multi-table reasoning are outside its scope. Pooled TFM embeddings also depend on the extraction interface, since TabPFN is conditioned on one column of each partition as its target and \textsc{ConTextTab} on a synthetic random target (Appendix~\ref{sec:embedders}), and other target choices may change the pooled embeddings. The confidence intervals of the learned heads reflect evaluation sampling for one validation-selected model per configuration, not variation across training runs. The learned-head results are also reported averaged over label types; whether their gains are uniform across Direct, Semantic, and Stat labels is not analyzed. Finally, several D3 robustness metrics saturate for permutation-invariant methods and separate methods only weakly.

\section{Conclusion}
\label{sec:conclusion}

\begin{revised}
Tables are compositional, and their table-level embeddings deserve to be evaluated directly rather than through row-level prediction. We introduced a benchmark that does so through partial-view consistency, discriminability across label granularities, robustness, and efficiency, and found lightweight schema- and text-based fingerprints consistently among the strongest of twelve methods. The states of TabPFN and \textsc{ConTextTab} composed by parameter-free pooling trail these fingerprints across five pooling rules. With the \textsc{ConTextTab} encoder frozen, compositional rules learned with SCDM substantially improve the same embeddings, and a downstream parent-table retrieval task confirms these findings. Weak table-level embeddings under fixed pooling therefore do not imply that the encoder states lack the relevant information, and the compositional rule and the objective that shapes it should be designed and evaluated as part of the model rather than left to a default pooling rule.
\end{revised}

\subsubsection*{Acknowledgments}
We would like to thank Sam Thelin, Marco Spinaci, Mojtaba Nayyeri, Myung Jun Kim, Markus Kohler, and Johannes Hoffart for their insightful comments and suggestions throughout the development of this work, which have greatly contributed to shaping its direction and quality.

\bibliography{references}
\bibliographystyle{tmlr}

\newpage

\appendix
\section{Appendix}

\subsection{Detailed descriptions of table embedding models}
\label{app:embedder_details}

A table-level embedding model is a function $f$ that maps a table $T$ to a fixed-dimensional vector 
$f(T) \in \mathbb{R}^d$.
To compare two tables under a fixed embedding method, we measure similarity between their corresponding embedding vectors. In this paper, we use cosine similarity as follows
$
\mathrm{sim}(T_a, T_b) = \cos\!\left(f(T_a), f(T_b)\right).
$
Although embedding models conceptually operate on full tables, our evaluation applies $f(\cdot)$ to 
\emph{table partitions}
generated from the same source table, which are themselves valid subtables. This yields a unified interface for benchmarking diverse embedding families under controlled partial-views and perturbation regimes (Section~\ref{app:evaluation}).

\subsubsection{Table embedding models}
\label{sec:embedders}

We evaluate table embedding models from multiple families with distinct inductive biases and input preprocessing.
Methods that emit row- or cell-level representations are mapped to a single table-level embedding via a fixed pooling operator.
For serialization-based methods,
we calculate a deterministic CSV-based string serialization $s(T)$ using a fixed number of rows.
Unless stated otherwise, we apply mean pooling. For row outputs $L \in \mathbb{R}^{n \times d}$, with $L_i \in \mathbb{R}^d$ the $i$-th row vector, we obtain the final embedding as $f(T)=\frac{1}{n}\sum_{i=1}^n L_i$.
For cell outputs $X \in \mathbb{R}^{n \times m \times d}$, with $X_{ij}\in \mathbb{R}^d$ a cell vector, we obtain the final embedding as $f(T)=\frac{1}{nm}\sum_{i=1}^{n}\sum_{j=1}^{m} X_{ij}$, where $n$ is the number of rows and $m$ is the number of columns.
All embedding models are evaluated under the same observation interface, meaning that if a method requires a fixed input budget (e.g., due to token/row limits), the limits are applied to the same table view used elsewhere in evaluation.
Hyperparameters are reported in Appendix~\ref{sec:exp_params} (Table~\ref{tab:embedder_hyperparams}).
Below, we list all table embedding models used in our experiments and group them into four distinct categories based on their underlying design principles.

\paragraph{(A) Structural \& statistical information.}
These methods are deterministic and emphasize schema and distributional signatures.
\begin{itemize}[leftmargin=*]
    \item \textbf{HashingSchema.} Schema tokens consist of column titles 
    paired with inferred data types, serialized as $\tau(T)=\{(\text{col}_j,\text{dtype}_j)\}_{j=1}^m$, and mapped by a hashing vectorizer $h$ with $d$ bins~\citep{scikit-learn}, yielding $f(T)=h(\tau(T))$.
    \item \textbf{SchemaContent.} Schema features are first serialized and hashed and subsequently concatenated with metadata features, $f(T) = [\phi_{\text{meta}}(T) \;\Vert\; h(\tau(T))]$, where $\phi_{\text{meta}}$ is a fixed-size vector that includes size, missingness, and type ratios.
    \item \textbf{TableStatistics.} A global feature vector $\phi(T)$ is formed from row/column counts, missingness ratios, average missing per row/column, and mean and standard deviation of per-column unique counts. Coarse type ratios (numeric/categorical/datetime) are appended, and the vector is padded or truncated to a fixed dimension.
    \item \textbf{StatisticalSummary.} For numeric columns $C_{\text{num}}$, per-column moments (min, max, mean, standard deviation, skew, median) are computed and aggregated by mean and standard deviation across columns. For categorical columns $C_{\text{cat}}$, cardinality, mode frequency, and entropy are computed and aggregated. For datetime columns $C_{\text{time}}$, min/max timestamps are aggregated. The concatenation yields $f(T)$.
    \item \textbf{MatrixFactorization.} Let $M_{\text{num}} \in \mathbb{R}^{n \times m'}$ be the numeric submatrix after mean imputation and column centering. Its singular values $\sigma_1 \ge \dots \ge \sigma_k$ define $f(T) = [\sigma_1, \dots, \sigma_k]$; if no numeric columns exist, a zero vector is returned.
\end{itemize}

\paragraph{(B) Serialization-based text embeddings.}
These methods treat tables as sequences and encode them with hashing or text models.
\begin{itemize}[leftmargin=*]
    \item \textbf{HashingText.} The table is serialized to a CSV-style string $s(T)$ and embedded as $f(T) = h(s(T))$ using a hashing vectorizer $h$~\citep{scikit-learn}.
    \item \textbf{LLMText.} A sentence-transformer encoder $e(\cdot)$~\citep{reimers2019sentencebert} is applied to the serialized table, $f(T) = e(s(T))$, mapping tabular content into a semantic text space.
\end{itemize}

\paragraph{(C) Specialized representation methods.}
These methods encode explicit relational structure beyond flat vectors or train a model to optimize a representation objective.
\begin{itemize}[leftmargin=*]
    \item \textbf{TableVectorizer~\citep{skrub}.} Each row is mapped by Skrub's mixed-type vectorizer $g(\cdot)$ to a feature vector; the table embedding is mean pooled, $f(T) = \frac{1}{n}\sum_i g(\text{row}_i)$, with padding or truncation to a fixed dimension.
    \item \textbf{HyTREL~\citep{chen2023hytrel}.} A hypergraph is constructed with cell nodes and hyperedges for the table, columns, and rows. The encoder returns hyperedge embeddings. We take the table-level hyperedge vector as $f(T)$. We employ the provided model checkpoint pretrained with contrastive and discriminative objectives.
    \item \textbf{Armadillo~\citep{pugnaloni2025armadilo}.} Tables are converted to a graph with row, column, and value nodes; each value node connects to its corresponding row and column. Initial node features are produced by a token/value encoder, and a GNN propagates them across the graph. The table embedding is the global mean pool of node representations, $f(T)=\frac{1}{|V|}\sum_{v\in V} z_v$. We use the provided model checkpoint pretrained to estimate the overlap ratio between tables.

\end{itemize}

\paragraph{(D) Foundation model-derived embeddings.}
We extract final-layer hidden states from pretrained state-of-the-art foundation models and pool them to obtain a table-level embedding.
\begin{itemize}[leftmargin=*]
    \item \textbf{TabPFN~\citep{hollmann2023tabpfn}, TabPFN-2.5~\citep{grinsztajn2025tabpfn25}.} TabPFN produces row embeddings only when conditioned on a target. We use the last column of each partition that has more than one distinct value as the target $y$ and the remaining columns as $X$, follow the extension proposed in~\citet{ye2025a} to extract row-level embeddings $L \in \mathbb{R}^{n \times d}$ conditioned on that target, and aggregate them into the table embedding $f(T) = \frac{1}{n}\sum_{i=1}^n L_i$. Taking the target from the partition itself keeps the interface task-independent and applicable to any table.
    \item \textbf{ConTextTab~\citep{spinaci2025ConTextTab}.} The in-context interface of \textsc{ConTextTab} also expects a target column. To avoid conditioning on a task-specific column, we append a synthetic regression target with standard-normal values drawn from the embedder's seeded random number generator. The encoder outputs a contextual grid over rows and columns; the target column and the query row are removed, giving $X \in \mathbb{R}^{n \times m \times d}$, and the grid is pooled, $f(T)=\frac{1}{nm}\sum_{i=1}^{n}\sum_{j=1}^{m} X_{ij}$.
\end{itemize}

\subsection{Detailed Evaluation Protocol}
\label{app:evaluation}

To evaluate the desiderata introduced in Section~\ref{sec:desiderata}, we propose a unified framework relying on proxy tasks that do not require downstream fine-tuning. We evaluate embeddings on \emph{table partitions} sampled as partial views of each source table, and attach multi-view supervision signals to probe different notions of similarity and grouping. All sampling and evaluation hyperparameters are fixed across methods and reported in the Appendix (Tables~\ref{tab:eval_config} and~\ref{tab:sampling_config}).

\subsubsection{Partition sampling and overlap}
For each source table $T$ with $n$ rows and $m$ columns, we sample $K_p$ partitions under a fixed observation budget (maximum number of rows/columns) applied consistently across all methods.\footnote{Exact budgets and sampling hyperparameters are reported in Appendix~\ref{sec:exp_params}.} We model a table as a typed relation $T = (R, C, X)$ with row index set $R = \{1, \dots, n\}$, column index set $C = \{1, \dots, m\}$, and cell values $X \in \mathcal{V}^{n \times m}$.
A \emph{partition} is a subtable induced by row and column subsets, $P = T[R', C']$. Let $\mathcal{P}(T)=\{P_1,\dots,P_{K_p}\}$ denote the sampled partitions from $T$.
We represent the extent of each partition by its coordinate set
\[
\mathcal{S}(P) = \{(i, j) \mid i \in R', j \in C'\},
\]
and define the overlap between two partitions $P_a$ and $P_b$ using the Intersection over Union (IoU),
\[
\mathrm{IoU}(P_a, P_b) = \frac{\lvert \mathcal{S}(P_a) \cap \mathcal{S}(P_b) \rvert}{\lvert \mathcal{S}(P_a) \cup \mathcal{S}(P_b) \rvert}.
\]
This overlap signal induces a natural notion of similarity between partial views of the same table and directly supports D1.
In the sampler, the overlap-ratio parameter does not fix a target IoU directly. Instead, it controls the forced-overlap portion of partition generation by reusing a fraction of the rows and columns from a previously sampled partition before filling the remaining entries at random.
We intentionally retain schema and column titles in this protocol. In corpus-scale tabular workflows such as indexing, retrieval, source recovery, curation, and approximate deduplication, column names and coarse types are often available metadata and therefore constitute a realistic signal rather than hidden supervision.

\subsubsection{Labeling schemes}
To assess different granularities of grouping, each partition $P$ is assigned labels under four schemes.
\begin{itemize}[leftmargin=*]
    \item \textbf{Direct.} Table identity, $y_{\text{dir}}(P)=\mathrm{id}(T)$, all partitions coming from the same table share the same label.
    \item \textbf{Semantic.} High-level semantic labels, $y_{\text{sem}}(P)=g(T)$, where $g$ encodes higher-level semantics from metadata or generative factors. For synthetic data, tuples of generative factors (e.g., \texttt{physics formula} + \texttt{density}) are mapped to distinct labels, while for real data tuples are formed from benchmark source and dataset identifier (e.g., \texttt{carte} + \texttt{nba draft}).
    \item \textbf{Semantic+Difficulty.} Semantic labels with an additional difficulty distinction for synthetic data (e.g., \texttt{physics formula} + \texttt{density} + \texttt{medium}), where the difficulty is a generative factor that controls the complexity of table values without altering other features.
    \item \textbf{Stat.} Coarse statistical signature, $y_{\text{stat}}(P)=l(P)$, where $l$ returns the majority coarse column type if its fraction exceeds a predefined threshold. A partition label defaults to a \textit{mixed label} if no coarse type fraction exceeds the threshold.
\end{itemize}

\subsubsection{Metrics}
To evaluate the performance of the different embedding models, we present a set of evaluation metrics that align with the desiderata (Figure~\ref{fig:intro}). Unless stated otherwise, retrieval, triplet accuracy, and the linear probe accuracy are computed under each labeling scheme (Direct/Semantic/Semantic+Difficulty/Stat).  We report scores averaged over label types for label-dependent metrics (D2 metrics) in the main Tables~\ref{tab:results_synthetic} and~\ref{tab:results_real}, with extended per-label breakdown in Tables~\ref{tab:results_synthetic_direct}--\ref{tab:results_synthetic_appendix_avg} and~\ref{tab:results_real_direct}--\ref{tab:results_real_appendix_avg} in Appendix~\ref{sec:app_results}, and per-metric summaries in Figures~\ref{fig:metrics_bar_grid_synthetic} and~\ref{fig:metrics_bar_grid_real}.

\paragraph{Partition consistency via IoU correlation (D1).}
We measure the Spearman correlation between the partition overlap $\mathrm{IoU}(P_a, P_b)$ and the embedding similarity $\cos\!\left(f(P_a), f(P_b)\right)$ over within-table pairs $\{(P_a,P_b)\mid P_a,P_b \in \mathcal{P}(T), a\neq b\}$. A higher correlation indicates that the embedding space preserves the similarity structure induced by overlapping partial views. We additionally report Pearson correlation and mean-squared-error (MSE) in Figures~\ref{fig:metrics_bar_grid_synthetic} and~\ref{fig:metrics_bar_grid_real}.

\paragraph{Triplet ranking, TR (D2).}
Given an anchor partition $A$, a positive partition $B$ with the same label ($y(B)=y(A), B\neq A$), and a negative partition $C$ with a different label ($y(C)\neq y(A)$), we evaluate whether the embedding places the positive closer to the anchor than the negative. We report the fraction of triplets satisfying
$\cos\!\left(f(A), f(B)\right) > \cos\!\left(f(A), f(C)\right)$ as a proxy for ranking capability. We report multiple versions of the triplet ranking depending on how the negative partition is sampled.
\textbf{TR-R} (random negative), samples $C$ uniformly from the negative pool.
\textbf{TR-H} (hardest negative), selects $C$ to be the most similar negative by cosine similarity.
\textbf{TR-CH} (cluster-hard negative), selects $C$ as the most similar negative restricted to the anchor's MiniBatch $k$-means cluster (with a different label). 
We aggregate $\textbf{TR-Avg}=\frac{1}{3}(\text{TR-R}+\text{TR-H}+\text{TR-CH})$, and report the different triplet metrics in Appendix~\ref{sec:app_results}.

\paragraph{Clustering alignment, CL (D2).}
We cluster embeddings using MiniBatch $k$-means and evaluate alignment with labels using
Purity, Normalized Mutual Information (NMI), and Adjusted Rand Index (ARI).
We aggregate $\textbf{CL-Avg}=\frac{1}{3}(\text{Purity}+\text{NMI}+\text{ARI})$, and report the different clustering metrics in Appendix~\ref{sec:app_results}.

\paragraph{Retrieval, Recall@$K_r$ (D2).}
For each anchor partition, we retrieve its $K_r$ nearest neighbors by cosine similarity, excluding the anchor. A retrieval query is successful if at least one of the top-$K_r$ neighbors shares the same label as the anchor. We report Recall@$K_r$ as the percentage of successful queries across all anchors.

\paragraph{Linear probe accuracy, LP (D2).}
We train a logistic regression classifier on top of the frozen partition embeddings to predict labels, assessing whether the embedding captures discriminative semantic or statistical information about the table.
Probe folds are source-group disjoint, meaning that partitions are split with stratified group $k$-folds grouped by Direct source-table identity, so no source table contributes partitions to both the training and test side of any retained fold. For the Direct label the group coincides with the class, so Direct LP uses label stratification and is read as closed-set source identification from unseen partial views of known tables. The full protocol, its old label-stratified counterpart, and their comparison are given in Appendix~\ref{app:lp_protocols}.

\paragraph{Robustness (D3).}
We perturb each partition by (i) permuting rows and columns, (ii) masking a subset of cells, or (iii) adding small noise to cell values, then measure the cosine similarity between the embeddings of the original and perturbed partitions.

\paragraph{Efficiency via throughput (D4).}
We report embedding throughput in tables per second (Tbl/s), measured as the wall-clock rate for computing embeddings over partitions measured on the same compute hardware. This enables comparison of the quality-cost trade-off across embedding methods.

\subsection{Data sources}
\label{app:data_details}
We evaluate on both synthetic and real open-source corpora to balance controlled semantics with real-world variability.
Across both settings, we sample $100$ source tables and compute the metrics on $K_p=10$ partitions per table under a shared observation budget (Table~\ref{tab:sampling_config}).

\paragraph{Synthetic data.}
Synthetic corpora are generated from predefined rules with controllable difficulty levels, enabling systematic variation in scale and nonlinearity while preserving table structure.
Each generator family provides generative factors that induce higher-level groupings in the embedding space, which are used for semantic (and semantic+difficulty) labels defined in Section~\ref{app:evaluation}.
We define four dataset families.
\begin{itemize}[leftmargin=*]
    \item \textbf{Physics formulas.} Fixed schema with numeric relations 
    (e.g., density, ideal gas, kinetic energy, Ohm's law)
    , emphasizing intra-column numerical reasoning.
    \item \textbf{Geometric shapes.} Mixed numeric/categorical tables with closed-form area/perimeter rules for different geometric shapes; additional random colors act as distractors.
    \item \textbf{Temporal reasoning.} Timestamped event sequences for process lifecycles (HR, order processing, invoicing). Each table is generated under a company-specific random profile with distinct evolution cycles.
    \item \textbf{SCM prior.} Data generated by structural causal models adapted from TabICL~\citep{qu2025tabicl} implementation, with controlled complexity and regression/classification targets.
\end{itemize}
Difficulty levels control numeric ranges and the number of SCM layers, yielding progressively harder variations without changing the labeling interface.

\paragraph{Real open-source data.}
Real tables are sampled from CARTE~\citep{kim2024carte}, OpenML-CC18~\citep{bischl2021openml}, and OpenML-CTR23~\citep{fischer2023openmlctr23}.
For each benchmark task, we form a table by concatenating the feature matrix with a single target column and evaluate on fixed-size partitions under the same budget as synthetic (Table~\ref{tab:sampling_config}).
We retain task metadata (benchmark source, dataset identifier; e.g., \texttt{carte+nba\_draft}) to derive semantic labels.

\newpage
\subsection{Further experimental details}
\label{sec:exp_params}

\begin{table*}[!ht]
\centering
\caption{Embedding model hyperparameters.}
\vspace{-0.3cm}
\label{tab:embedder_hyperparams}
\footnotesize
\begin{tabular}{llp{9cm}}
\toprule
\textbf{Embedder} & \textbf{Dim} & \textbf{Key Hyperparameters} \\
\midrule
HashingSchema & 1024 & \texttt{n\_features}=1024 (feature hashing on schema) \\
\midrule
SchemaContent & 256 & \texttt{n\_features}=256 \\
\midrule
TableStatistics & 14 & \texttt{output\_dim}=14 (fixed statistical features) \\
\midrule
StatisticalSummary & 22 & \makecell[l]{Numeric: min, max, mean, std, skew, median; \\ Categorical: cardinality, entropy} \\
\midrule
MatrixFactorization & 16 & \texttt{n\_components}=16, \texttt{center}=True \\
\midrule
HashingText & 1024 & \texttt{n\_features}=1024 (feature hashing on text) \\
\midrule
LLMText & 384 & \texttt{model}=all-MiniLM-L6-v2, \texttt{max\_rows}=32 \\
\midrule
TableVectorizer & 512 & \texttt{output\_dim}=512, \texttt{cardinality\_threshold}=10 \\
\midrule
HyTREL & 768 & \makecell[l]{\texttt{model}=bert-base-uncased, \texttt{batch\_size}=16 \\ \texttt{max\_token\_len}=64, \texttt{max\_col\_len}=20,  \texttt{max\_row\_len}=30} \\
\midrule
Armadillo & 300 & \makecell[l]{\texttt{gnn\_type}=GraphSAGE, \texttt{num\_layers}=3 \\ \texttt{hidden\_channels}=300, \texttt{initial\_embedding}=sha256} \\
\midrule
TabPFN & 192 & \makecell[l]{\texttt{n\_estimators}=1, \texttt{max\_samples}=50k, \texttt{max\_features}=2000 \\ \texttt{max\_classes}=10, \texttt{data source}=test}\\
\midrule
ConTextTab & 768 & \makecell[l]{\texttt{model\_size}=base, \texttt{classification}=cross-entropy \\ \texttt{regression}=l2, \texttt{num\_regression\_bins}=10} \\
\bottomrule
\end{tabular}
\end{table*}

\begin{table*}[!ht]
\centering
\caption{Evaluation metrics and their parameters.}
\vspace{-0.3cm}
\label{tab:eval_config}
\footnotesize
\begin{tabular}{lll}
\toprule
\textbf{Desideratum} & \textbf{Metric} & \textbf{Configuration} \\
\midrule
\multirow{3}{*}{D1: Consistency} & Pearson correlation & IoU vs. cosine similarity \\
 & Spearman correlation & IoU vs. cosine similarity \\
 & MSE & IoU vs. cosine similarity \\
\midrule
\multirow{4}{*}{D2: Triplet} & TR-R (Random) & Random negative, margin=0.01\\
 & TR-H (Hard) & Hardest negative mining \\
 & TR-CH (Cluster-Hard) & Hard negatives within clusters \\
 & TR-Avg & Mean of TR-R, TR-H, TR-CH \\
\midrule
\multirow{4}{*}{D2: Clustering} & Purity & KMeans, n\_cluster=10 \\
 & NMI & Normalized mutual information \\
 & ARI & Adjusted Rand index \\
 & CL-Avg & Mean of Purity, NMI, ARI \\
\midrule
D2: Retrieval & Recall@5 & Top-5 nearest neighbors \\
\midrule
D2: Linear Probe & Accuracy & Logistic regression (max\_iter=1000) \\
\midrule
\multirow{3}{*}{D3: Robustness} & Permutation & Row/column shuffle invariance \\
 & Noise & $\sigma$=[0.0, 0.01, 0.05, 0.1] Gaussian noise \\
 & Masking & [0,5,10,25]\% random cell masking \\
\midrule
D4: Efficiency & Tables/second & Embedding throughput (measured on Nvidia T4 GPU) \\
\bottomrule
\end{tabular}
\end{table*}

\begin{table}[!ht]
\centering
\caption{Data sampling and partitioning configuration.}
\vspace{-0.3cm}
\label{tab:sampling_config}
\footnotesize
\begin{tabular}{ll ll ll}
\toprule
\textbf{Parameter} & \textbf{Value} & \textbf{Parameter} & \textbf{Value} & \textbf{Parameter} & \textbf{Value} \\
\midrule
Tables & 100 & Partitions/table & 10 & Overlap ratio & 0.5 \\
Row fraction & [0.2, 0.5] & Column fraction & [0.2, 0.5] & Min partition & $10 \times 5$ \\
Seeds & 10 & Rows (synthetic) & 100 & Rows (real) & varies \\
\bottomrule
\end{tabular}
\end{table}

\newpage
\subsection{Linear-probe stratification}
\label{app:lp_protocols}

\paragraph{Label stratification and source-table identity.} The D2 linear probe fits a logistic-regression classifier on frozen partition embeddings. Stratifying the train/test split by label alone allows partitions drawn from the \emph{same} source table to fall on both sides of the boundary, since every source table contributes $K_p$ partitions. A probe can then score well by recognising the source table rather than by decoding the label, which rewards embeddings that preserve fine-grained source identity independently of the semantics being probed.

\paragraph{Direct labels.} For the \emph{Direct} label this is not a confound but the task definition, since the source table \emph{is} the class. Holding out whole source tables would remove each class from training and turn the probe into unseen-class classification. Direct LP is therefore probed under label stratification and should be read as a \emph{closed-set source-identification} test, which measures whether distinct partial views of a known set of tables can be attributed to their source table. Identical partitions never occur on both sides of the split, so the probe is not trivially solvable by memorising a repeated view; it is not, however, a test of generalisation to unseen source tables, and we do not interpret it as such. Because its split semantics differ from the group-stratified probes, we report Direct LP alongside the other labels but do not treat it as evidence of table-level generalisation.

\paragraph{Group stratification elsewhere.} For every other label type we stratify by group, taking Direct source-table identity as the grouping variable, so that no source table appears on both sides of a retained fold. The number of folds is bounded by the distinct source groups available per class (at most five), and a fold is scored only when every test class is represented in training. Two label types provide too few source groups per class to retain any fold---\emph{Semantic+Difficulty} on synthetic data and \emph{Semantic} on real data---and fall back to label stratification.

\paragraph{Effect of the two schemes.} Table~\ref{tab:lp_protocols} reports both schemes on every label type where group stratification is well posed. The differences concentrate exactly where source identity is most recoverable, namely content-fingerprint and identity-preserving embeddings on fine-grained labels. HashingText moves from $0.99$ to $0.63$ on synthetic Semantic labels, Armadillo from $0.72$ to $0.40$, and HyTREL from $0.91$ to $0.81$, while schema-level and statistical methods are essentially unchanged. Ranking methods by the probe alone is thus sensitive to the scheme (Spearman correlation between the two orderings $0.51$ on synthetic and $0.39$ on real data). The composite ranking is not, since in Tables~\ref{tab:results_synthetic} and~\ref{tab:results_real} lightweight schema- and text-based methods continue to outrank every pooled TFM on both datasets.

\begin{table*}[!ht]
\centering
\small
\caption{Linear-probe accuracy under label stratification and group stratification (mean $\pm$ 95\% CI over 10 seeds), for the label types where group stratification is well posed. \emph{Direct} labels are probed under label stratification by construction, since the source table is itself the class. TabPFN and \textsc{ConTextTab} are shown under mean pooling.}
\label{tab:lp_protocols}
\begin{tabular}{l|cc|cc|cc}
\toprule
 & \multicolumn{2}{c|}{Synthetic -- Semantic} & \multicolumn{2}{c|}{Synthetic -- Stat} & \multicolumn{2}{c}{Real -- Stat} \\
Embedder & Label & Group & Label & Group & Label & Group \\
\midrule
HashingSchema & 0.88$\pm$0.02 & 0.88$\pm$0.02 & 1.00$\pm$0.00 & 1.00$\pm$0.00 & 0.96$\pm$0.01 & 0.84$\pm$0.02 \\
SchemaContent & 0.88$\pm$0.02 & 0.87$\pm$0.02 & 1.00$\pm$0.00 & 0.99$\pm$0.00 & 0.97$\pm$0.01 & 0.96$\pm$0.01 \\
TableStatistics & 0.62$\pm$0.03 & 0.58$\pm$0.03 & 0.99$\pm$0.00 & 0.99$\pm$0.00 & 0.99$\pm$0.00 & 0.99$\pm$0.00 \\
StatisticalSummary & 0.71$\pm$0.03 & 0.68$\pm$0.03 & 0.99$\pm$0.01 & 0.99$\pm$0.00 & 0.92$\pm$0.02 & 0.90$\pm$0.01 \\
MatrixFactorization & 0.53$\pm$0.03 & 0.51$\pm$0.02 & 0.92$\pm$0.02 & 0.92$\pm$0.01 & 0.65$\pm$0.02 & 0.65$\pm$0.02 \\
HashingText & 0.99$\pm$0.00 & 0.63$\pm$0.03 & 0.98$\pm$0.00 & 0.94$\pm$0.01 & 0.93$\pm$0.01 & 0.83$\pm$0.02 \\
LLMText & 0.93$\pm$0.02 & 0.91$\pm$0.02 & 0.99$\pm$0.00 & 1.00$\pm$0.00 & 0.95$\pm$0.01 & 0.90$\pm$0.01 \\
TableVectorizer & 0.36$\pm$0.02 & 0.34$\pm$0.02 & 0.92$\pm$0.01 & 0.92$\pm$0.01 & 0.70$\pm$0.03 & 0.68$\pm$0.04 \\
HyTREL & 0.91$\pm$0.01 & 0.81$\pm$0.02 & 0.97$\pm$0.01 & 0.97$\pm$0.00 & 0.92$\pm$0.01 & 0.90$\pm$0.01 \\
Armadillo & 0.72$\pm$0.01 & 0.40$\pm$0.03 & 0.93$\pm$0.01 & 0.85$\pm$0.02 & 0.83$\pm$0.02 & 0.71$\pm$0.02 \\
TabPFN & 0.66$\pm$0.02 & 0.61$\pm$0.01 & 0.87$\pm$0.02 & 0.85$\pm$0.02 & 0.79$\pm$0.02 & 0.76$\pm$0.01 \\
\textsc{ConTextTab} & 0.80$\pm$0.02 & 0.74$\pm$0.01 & 0.95$\pm$0.01 & 0.96$\pm$0.01 & 0.96$\pm$0.01 & 0.94$\pm$0.01 \\
\bottomrule
\end{tabular}
\end{table*}

\paragraph{Direct LP is a distinct regime; the conclusion does not depend on it.} The two probes answer different questions. For the Semantic, Semantic+Difficulty, and Stat labels, several source tables share a label, so source-group-disjoint folds test generalisation to \emph{unseen} source tables within a known higher-level class. For \emph{Direct}, source-table identity is itself the target class, and the probe asks whether a newly sampled partial view of one of the \emph{known} source tables can be attributed to it. Source-group holdout is not an appropriate correction here, since removing a whole source group deletes that class from training and converts a closed-set probe into unseen-class classification. We therefore retain label stratification over independently sampled, non-identical partitions and read Direct LP as \emph{closed-set source-table identification from unseen partial views of known tables}. Shared source content across a table's partitions is not leakage in this regime but part of the task definition; we do not present Direct LP as generalisation to unseen source tables.

Because the headline LP column averages this closed-set probe together with the group-disjoint probes, we report a second aggregate that excludes Direct, averaging only the Semantic, Semantic+Difficulty, and Stat probes (Table~\ref{tab:lp_grouped}), to be compared with the LP column of Tables~\ref{tab:results_synthetic} and~\ref{tab:results_real}. The two aggregates agree on the substance, as lightweight schema- and text-based methods (HashingText, LLMText, SchemaContent, HashingSchema) hold the top linear-probe positions and both pooled TFMs remain below them on synthetic and real data. The composite ranking in Tables~\ref{tab:results_synthetic} and~\ref{tab:results_real} is therefore not an artefact of mixing the two evaluation regimes.

\begin{table*}[!ht]
\centering
\small
\caption{Linear-probe accuracy aggregated over the source-group-disjoint labels only (Semantic, Semantic+Difficulty, Stat; Direct excluded), mean $\pm$ 95\% CI over 10 seeds. The aggregate that includes the closed-set Direct probe is the LP column of Tables~\ref{tab:results_synthetic} and~\ref{tab:results_real}. TabPFN and \textsc{ConTextTab} use mean pooling. Best value per dataset in \textbf{bold}. Lightweight schema- and text-based methods lead under both aggregates, so the headline conclusion does not depend on the Direct regime.}
\label{tab:lp_grouped}
\begin{tabular}{l|c|c}
\toprule
Embedder & Synthetic & Real \\
\midrule
HashingSchema & 0.77$\pm$0.01 & 0.92$\pm$0.01 \\
SchemaContent & 0.77$\pm$0.02 & \textbf{0.95$\pm$0.01} \\
TableStatistics & 0.63$\pm$0.02 & 0.64$\pm$0.02 \\
StatisticalSummary & 0.69$\pm$0.01 & 0.58$\pm$0.01 \\
MatrixFactorization & 0.61$\pm$0.02 & 0.30$\pm$0.01 \\
HashingText & 0.87$\pm$0.03 & 0.94$\pm$0.01 \\
LLMText & \textbf{0.88$\pm$0.01} & 0.94$\pm$0.01 \\
TableVectorizer & 0.48$\pm$0.01 & 0.31$\pm$0.02 \\
HyTREL & 0.82$\pm$0.02 & 0.84$\pm$0.01 \\
Armadillo & 0.66$\pm$0.04 & 0.71$\pm$0.02 \\
TabPFN & 0.61$\pm$0.01 & 0.46$\pm$0.02 \\
\textsc{ConTextTab} & 0.69$\pm$0.01 & 0.78$\pm$0.02 \\
\bottomrule
\end{tabular}
\end{table*}

\subsection{Column-name anonymization ablation}
\label{app:column_anonymization}

The main benchmark intentionally preserves column names because schema metadata is normally available in table discovery, indexing, source-recovery, and RAG-style retrieval workflows. In these settings, column names are not merely shortcuts; they are part of the observed table and often encode the operational meaning of a column. Still, the same signal can make an evaluation overly sensitive to lexical schema overlap, especially for schema-only and serialization-based methods. We therefore add a controlled column-name anonymization ablation to separate exact schema-name matching from content- and structure-based representation quality.

The ablation keeps the task fixed. We sample and label partitions exactly as in the main benchmark, then copy each sampled partition and rename its columns positionally to \texttt{col\_1}, \ldots, \texttt{col\_m} immediately before representation extraction. Cell values, inferred dtypes, labels, coordinate sets, and partition coordinates are unchanged. The comparison therefore removes lexical column-name identity while preserving the same sampled views and evaluation targets. Tables~\ref{tab:anonymization_delta_synthetic} and~\ref{tab:anonymization_delta_real} report the anonymized values and the paired changes relative to the same partitions with original column names.  D2 Avg denotes the mean of TR-Avg, CL-Avg, Recall@5, and LP; table entries are rounded for display, while rank deltas are recomputed over the common baseline methods. Positive rank deltas indicate worse relative rank under anonymization.

The ablation has a useful property, since it changes one input channel while keeping the sampled values and targets fixed. The implementation-level effect differs by method. HashingSchema serializes column names and dtypes, so anonymization directly removes its central lexical feature. SchemaContent combines name/type schema tokens with metadata, so only part of its input is changed. HashingText and LLMText serialize tables with headers and values, so anonymization removes the header channel but leaves cell content intact. TableStatistics, StatisticalSummary, and MatrixFactorization serve as name-light controls because they operate on size, type, value, or numeric summaries. For imported neural encoders that also represent row/column structure, we report the same deltas but interpret them more cautiously.

The outcome is mixed in an informative way. Schema cues clearly matter, as HashingSchema drops sharply on synthetic data ($\Delta$D1 $=-0.29$, $\Delta$D2 Avg $=-0.12$) and also degrades on real data, especially in D2 ($\Delta$D2 Avg $=-0.18$). LLMText also loses D1 consistency under anonymization, suggesting that sentence-transformer similarity is sensitive to header lexical distribution for view-to-view matching, although its D2 Avg changes by only $-0.05$ in both corpora. At the same time, the strongest content-preserving lightweight method remains stable. HashingText changes by only $+0.01$ in D1 and $-0.01$ in D2 Avg on synthetic data, and by $-0.05$ in D1 and $-0.01$ in D2 Avg on real data. The name-light controls are also largely stable. Thus, schema information is a useful and realistic signal in table discovery settings, but the competitiveness of lightweight methods is not explained solely by exact column-name matching.

\begin{table*}[!ht]
\centering
\begingroup
\footnotesize
\setlength{\tabcolsep}{2.5pt}
\caption{Column-name anonymization deltas for synthetic data. Anon.\ reports values after renaming partition columns to positional names, and $\Delta$ the paired change relative to the same sampled partitions with their original column names. The input-channel column summarizes which implemented signal is most directly changed by anonymization. D2 Avg averages TR-Avg, CL-Avg, Recall@5, and LP. Positive $\Delta$Rank means worse rank after anonymization.}
\label{tab:anonymization_delta_synthetic}
\resizebox{\textwidth}{!}{
\begin{tabular}{llcc|cc|cc}
\toprule
\multirow{2}{*}{Embedder} & \multirow{2}{*}{Input channel changed} & \multicolumn{2}{c|}{D1 Spearman} & \multicolumn{2}{c|}{D2 Avg} & \multicolumn{2}{c}{Summary} \\
\cmidrule(lr){3-4} \cmidrule(lr){5-6} \cmidrule(lr){7-8}
 & & Anon. & $\Delta$ & Anon. & $\Delta$ & $\Delta$D3 Avg & $\Delta$Rank \\
\midrule
HashingSchema & Column names + dtypes & 0.29 & -0.29 & 0.49 & -0.12 & 0.00 & +2.68 \\
SchemaContent & Names/types + metadata & 0.26 & -0.02 & 0.36 & -0.05 & 0.00 & +0.25 \\
TableStatistics & No names; size/type stats & 0.32 & -0.02 & 0.42 & 0.00 & 0.00 & -0.38 \\
StatisticalSummary & No names; value summaries & 0.37 & -0.02 & 0.57 & 0.00 & 0.00 & -0.33 \\
MatrixFactorization & No names; numeric values & 0.39 & +0.02 & 0.45 & 0.00 & 0.00 & -0.92 \\
HashingText & Full CSV headers + values & 0.55 & +0.01 & 0.72 & -0.01 & -0.01 & -0.42 \\
LLMText & Head-row CSV headers + values & -0.02 & -0.32 & 0.62 & -0.05 & 0.00 & +1.18 \\
TableVectorizer & Row values after vectorization & 0.33 & +0.01 & 0.38 & 0.00 & 0.00 & -0.75 \\
HyTREL & Headers + cell values & 0.16 & -0.02 & 0.53 & -0.01 & -0.01 & -0.03 \\
Armadillo & Column/value graph structure & 0.39 & -0.02 & 0.56 & 0.00 & 0.00 & -0.33 \\
TabPFN & Values + target choice & -0.13 & 0.00 & 0.41 & 0.00 & 0.00 & -0.25 \\
\textsc{ConTextTab} & Column-name/content embeddings & -0.01 & -0.07 & 0.42 & 0.00 & 0.00 & -0.71 \\
\bottomrule
\end{tabular}
}
\endgroup
\end{table*}

\begin{table*}[!ht]
\centering
\begingroup
\footnotesize
\setlength{\tabcolsep}{2.5pt}
\caption{Column-name anonymization deltas for real data. Anon.\ reports values after renaming partition columns to positional names, and $\Delta$ the paired change relative to the same sampled partitions with their original column names. The input-channel column summarizes which implemented signal is most directly changed by anonymization. D2 Avg averages TR-Avg, CL-Avg, Recall@5, and LP. Positive $\Delta$Rank means worse rank after anonymization.}
\label{tab:anonymization_delta_real}
\resizebox{\textwidth}{!}{
\begin{tabular}{llcc|cc|cc}
\toprule
\multirow{2}{*}{Embedder} & \multirow{2}{*}{Input channel changed} & \multicolumn{2}{c|}{D1 Spearman} & \multicolumn{2}{c|}{D2 Avg} & \multicolumn{2}{c}{Summary} \\
\cmidrule(lr){3-4} \cmidrule(lr){5-6} \cmidrule(lr){7-8}
 & & Anon. & $\Delta$ & Anon. & $\Delta$ & $\Delta$D3 Avg & $\Delta$Rank \\
\midrule
HashingSchema & Column names + dtypes & 0.25 & -0.07 & 0.50 & -0.18 & -0.01 & +1.04 \\
SchemaContent & Names/types + metadata & 0.25 & -0.01 & 0.37 & -0.12 & 0.00 & +0.38 \\
TableStatistics & No names; size/type stats & 0.21 & -0.01 & 0.41 & -0.01 & 0.00 & +0.17 \\
StatisticalSummary & No names; value summaries & 0.21 & +0.01 & 0.50 & +0.01 & 0.00 & -0.33 \\
MatrixFactorization & No names; numeric values & 0.19 & +0.02 & 0.33 & -0.01 & 0.00 & -0.29 \\
HashingText & Full CSV headers + values & 0.35 & -0.05 & 0.65 & -0.01 & 0.00 & -0.35 \\
LLMText & Head-row CSV headers + values & -0.02 & -0.16 & 0.65 & -0.05 & -0.01 & +0.85 \\
TableVectorizer & Row values after vectorization & 0.01 & -0.01 & 0.30 & -0.01 & 0.00 & -0.33 \\
HyTREL & Headers + cell values & 0.12 & -0.04 & 0.50 & -0.02 & -0.01 & -0.08 \\
Armadillo & Column/value graph structure & 0.22 & +0.01 & 0.53 & 0.00 & 0.00 & -0.67 \\
TabPFN & Values + target choice & 0.06 & +0.01 & 0.34 & +0.01 & 0.00 & -0.42 \\
\textsc{ConTextTab} & Column-name/content embeddings & 0.18 & 0.00 & 0.46 & -0.02 & 0.00 & +0.04 \\
\bottomrule
\end{tabular}
}
\endgroup
\end{table*}

\newpage

\subsection{Non-learned pooling study for TFM embeddings}
\label{app:pooling}

Mean pooling alone cannot distinguish a broader limitation of non-learned aggregation from a limitation specific to averaging. We therefore evaluate \emph{all five} non-learned poolings (mean, max, sum, min, std) for both TabPFN and \textsc{ConTextTab}. All poolings are emitted from a \emph{single} forward pass per model, as TabPFN reduces the same row-level embedding tensor and \textsc{ConTextTab} the same contextual cell-state grid, so the backbone, preprocessing, inputs, partitions, seeds, and output dimensionality are held fixed, and the only difference is the final aggregation operator. The D3 perturbations (permutation, noise, masking) are likewise generated once and shared, so every pooling is scored on byte-identical perturbed tables. Mean and coordinatewise max are reported as matched non-learned controls in the main tables (Tables~\ref{tab:results_synthetic} and~\ref{tab:results_real}); the remaining rules (sum, min, std) appear here only as a sensitivity analysis. Tables~\ref{tab:pooling_contexttab} and~\ref{tab:pooling_tabpfn} report D1 and the D2 components other than the linear probe for all five, so no aggregation rule is chosen using the benchmark it is evaluated on.

Three observations follow. First, no non-learned pooling changes the conclusion, as the spread in D2 average across poolings is at most $0.06$ for \textsc{ConTextTab} and $0.01$ for TabPFN, far smaller than the gap to the lightweight baselines, and under mean and max, the two rules ranked in Tables~\ref{tab:results_synthetic} and~\ref{tab:results_real}, both TFMs rank below every lightweight method. Second, mean and sum coincide on every cosine-based metric because cosine similarity is scale-invariant and sum equals mean times the number of rows; their agreement is a sanity check rather than a duplicated configuration. Third, aggregate stability does not imply metric-level stability, since TabPFN's synthetic D1 consistency flips from $-0.08$ under mean to $+0.15$ under std, so no single non-learned rule dominates at the level of individual metrics.

\begin{table*}[!ht]
\centering
\small
\caption{\textsc{ConTextTab} under all non-learned poolings (mean $\pm$ 95\% CI over 10 seeds). D2 avg averages the triplet, clustering, and retrieval columns as displayed. Best value per column in \textbf{bold}.}
\label{tab:pooling_contexttab}
\begin{tabular}{lrrrrr}
\toprule
\multicolumn{6}{c}{\textbf{Real data}} \\
Pooling & D1 Spear. & Triplet avg & Cluster avg & R@5 & D2 avg \\
\midrule
mean & \textbf{0.19$\pm$0.03} & \textbf{0.25$\pm$0.01} & 0.22$\pm$0.01 & 0.76$\pm$0.01 & \textbf{0.41} \\
max & -0.06$\pm$0.03 & 0.14$\pm$0.01 & 0.23$\pm$0.02 & 0.72$\pm$0.02 & 0.36 \\
sum & \textbf{0.19$\pm$0.03} & \textbf{0.25$\pm$0.01} & 0.22$\pm$0.01 & 0.76$\pm$0.01 & \textbf{0.41} \\
min & -0.08$\pm$0.03 & 0.15$\pm$0.01 & 0.22$\pm$0.01 & 0.69$\pm$0.01 & 0.35 \\
std & -0.05$\pm$0.03 & 0.03$\pm$0.01 & \textbf{0.27$\pm$0.03} & \textbf{0.77$\pm$0.01} & 0.36 \\
\midrule
\multicolumn{6}{c}{\textbf{Synthetic data}} \\
Pooling & D1 Spear. & Triplet avg & Cluster avg & R@5 & D2 avg \\
\midrule
mean & 0.04$\pm$0.03 & \textbf{0.18$\pm$0.00} & 0.27$\pm$0.02 & 0.70$\pm$0.01 & \textbf{0.38} \\
max & 0.03$\pm$0.04 & 0.08$\pm$0.01 & 0.26$\pm$0.02 & 0.68$\pm$0.01 & 0.34 \\
sum & 0.04$\pm$0.03 & \textbf{0.18$\pm$0.00} & 0.27$\pm$0.02 & 0.70$\pm$0.01 & \textbf{0.38} \\
min & 0.02$\pm$0.03 & 0.09$\pm$0.00 & 0.24$\pm$0.01 & 0.66$\pm$0.01 & 0.33 \\
std & \textbf{0.07$\pm$0.04} & 0.01$\pm$0.00 & \textbf{0.28$\pm$0.03} & \textbf{0.71$\pm$0.01} & 0.33 \\
\bottomrule
\end{tabular}
\end{table*}

\begin{table*}[!ht]
\centering
\small
\caption{TabPFN under all non-learned poolings (mean $\pm$ 95\% CI over 10 seeds). Columns as in Table~\ref{tab:pooling_contexttab}. Best value per column in \textbf{bold}.}
\label{tab:pooling_tabpfn}
\begin{tabular}{lrrrrr}
\toprule
\multicolumn{6}{c}{\textbf{Real data}} \\
Pooling & D1 Spear. & Triplet avg & Cluster avg & R@5 & D2 avg \\
\midrule
mean & 0.06$\pm$0.04 & 0.13$\pm$0.01 & \textbf{0.20$\pm$0.01} & \textbf{0.60$\pm$0.02} & \textbf{0.31} \\
max & \textbf{0.07$\pm$0.04} & 0.16$\pm$0.01 & 0.18$\pm$0.01 & \textbf{0.60$\pm$0.01} & \textbf{0.31} \\
sum & 0.06$\pm$0.04 & 0.13$\pm$0.01 & \textbf{0.20$\pm$0.01} & \textbf{0.60$\pm$0.02} & \textbf{0.31} \\
min & 0.06$\pm$0.03 & 0.16$\pm$0.01 & 0.19$\pm$0.01 & 0.57$\pm$0.01 & \textbf{0.31} \\
std & 0.03$\pm$0.03 & \textbf{0.20$\pm$0.00} & 0.18$\pm$0.01 & 0.54$\pm$0.01 & \textbf{0.31} \\
\midrule
\multicolumn{6}{c}{\textbf{Synthetic data}} \\
Pooling & D1 Spear. & Triplet avg & Cluster avg & R@5 & D2 avg \\
\midrule
mean & -0.08$\pm$0.05 & 0.17$\pm$0.00 & \textbf{0.25$\pm$0.02} & \textbf{0.73$\pm$0.01} & 0.38 \\
max & -0.17$\pm$0.04 & \textbf{0.21$\pm$0.01} & \textbf{0.25$\pm$0.01} & 0.71$\pm$0.01 & \textbf{0.39} \\
sum & -0.08$\pm$0.05 & 0.17$\pm$0.00 & \textbf{0.25$\pm$0.02} & \textbf{0.73$\pm$0.01} & 0.38 \\
min & -0.15$\pm$0.04 & \textbf{0.21$\pm$0.00} & \textbf{0.25$\pm$0.01} & 0.71$\pm$0.01 & \textbf{0.39} \\
std & \textbf{0.15$\pm$0.04} & \textbf{0.21$\pm$0.00} & \textbf{0.25$\pm$0.01} & 0.70$\pm$0.01 & \textbf{0.39} \\
\bottomrule
\end{tabular}
\end{table*}

\newpage 

\subsection{Extended D2 per-label results}

This appendix provides a label-wise decomposition of D2, complementing the label-averaged D2 scores reported in the main paper.
For each label type, we report triplet ranking under three negative-sampling strategies (TR-R / TR-H / TR-CH), clustering alignment (Purity / NMI / ARI), retrieval (Recall@5), and linear-probe accuracy.
TR-H and TR-CH are typically harder than TR-R because they focus on high-similarity negatives (hard negatives) and confusable negatives from the anchor's embedding cluster (cluster-hard negatives), respectively.
The \textit{Avg Rank} column summarizes performance across the individual D2 components for that label type.
Linear-probe folds are stratified by source table wherever that split is well posed, and by label for \emph{Direct}, where the source table is itself the class (Appendix~\ref{app:lp_protocols}).

\label{sec:app_results}
\subsubsection{Synthetic data results}

\paragraph{Direct labeling exposes identity collisions under hard and cluster-hard negatives for most methods.}
Under Direct labeling (Table~\ref{tab:results_synthetic_direct}), several methods achieve high TR-R yet collapse to near-zero on TR-H and TR-CH, confirming that random negatives are often trivial, while the hard and cluster-hard negatives reveal embedding collisions.
HashingText is the clear outlier, remaining strong on both TR-H and TR-CH, consistent with preserving fine-grained identity features from serialized content.
Notably, Armadillo is the only other method with non-trivial TR-H and TR-CH (while still trailing HashingText), indicating partial identity preservation beyond schema/statistical fingerprints.
By contrast, schema-only and coarse statistical baselines (e.g., HashingSchema, TableStatistics, StatisticalSummary) retain reasonable TR-R but fail under hard negatives, consistent with many partitions sharing near-identical schema/type signatures.
Finally, pooled TFM embeddings (TabPFN, ConTextTab) show moderate TR-R performance but collapse under hard negatives, suggesting that pooling produces weak identity separation even when some coarse similarity is recoverable.

\begin{table*}[!ht]
\centering
\small
\caption{D2 Metrics for Synthetic data - Direct Labels (mean $\pm$ 95\% CI over 10 seeds)}
\label{tab:results_synthetic_direct}
\resizebox{\textwidth}{!}{
\begin{tabular}{lccccc|ccccc|ccc|c}
\toprule
\multirow{2}{*}{Embedder} & \multicolumn{5}{c}{Triplet (Direct)} & \multicolumn{5}{c}{Clustering (Direct)} & \multicolumn{3}{c}{Retrieval/Probe (Direct)} & \multicolumn{1}{c}{Overall} \\
\cmidrule(lr){2-6} \cmidrule(lr){7-11} \cmidrule(lr){12-14} \cmidrule(lr){15-15}
 & TR-R & TR-H & TR-CH & TR-Avg & Rank & Purity & NMI & ARI & CL-Avg & Rank & R@5 & LP & Rank & Avg Rank \\
\midrule
HashingSchema & 0.89$\pm$0.01 & 0.00$\pm$0.00 & 0.00$\pm$0.00 & 0.30$\pm$0.00 & 4.75 & 0.12$\pm$0.01 & 0.57$\pm$0.02 & 0.10$\pm$0.01 & 0.26$\pm$0.01 & \cellcolor{green!40}2.00 & 0.23$\pm$0.01 & 0.12$\pm$0.02 & \cellcolor{red!60}12.50 & 6.42 \\
SchemaContent & 0.02$\pm$0.00 & 0.00$\pm$0.00 & 0.00$\pm$0.00 & 0.01$\pm$0.00 & \cellcolor{red!60}13.50 & 0.07$\pm$0.01 & 0.34$\pm$0.01 & 0.03$\pm$0.00 & 0.15$\pm$0.01 & 10.50 & 0.35$\pm$0.01 & 0.11$\pm$0.02 & 11.00 & \cellcolor{red!40}11.67 \\
TableStatistics & 0.38$\pm$0.02 & 0.00$\pm$0.00 & 0.00$\pm$0.00 & 0.13$\pm$0.01 & \cellcolor{red!20}12.50 & 0.10$\pm$0.01 & 0.48$\pm$0.01 & 0.07$\pm$0.01 & 0.22$\pm$0.01 & 5.25 & 0.30$\pm$0.01 & 0.12$\pm$0.02 & \cellcolor{red!20}11.50 & 9.75 \\
StatisticalSummary & 0.80$\pm$0.01 & 0.00$\pm$0.00 & 0.01$\pm$0.00 & 0.27$\pm$0.00 & 4.25 & 0.11$\pm$0.01 & 0.51$\pm$0.01 & 0.07$\pm$0.00 & 0.23$\pm$0.01 & 4.25 & 0.75$\pm$0.01 & 0.18$\pm$0.02 & 4.50 & 4.33 \\
MatrixFactorization & 0.69$\pm$0.03 & 0.00$\pm$0.00 & 0.00$\pm$0.00 & 0.23$\pm$0.01 & 8.75 & 0.10$\pm$0.01 & 0.45$\pm$0.01 & 0.04$\pm$0.01 & 0.19$\pm$0.01 & 7.00 & 0.50$\pm$0.02 & 0.15$\pm$0.02 & 7.50 & 7.75 \\
HashingText & \textbf{0.98$\pm$0.00} & \textbf{0.70$\pm$0.02} & \textbf{0.72$\pm$0.02} & \textbf{0.80$\pm$0.01} & \cellcolor{green!60}1.00 & \textbf{0.13$\pm$0.01} & 0.57$\pm$0.01 & 0.09$\pm$0.01 & 0.26$\pm$0.01 & \cellcolor{green!20}2.50 & \textbf{0.99$\pm$0.00} & \textbf{0.99$\pm$0.00} & \cellcolor{green!60}1.00 & \cellcolor{green!60}1.50 \\
LLMText & 0.91$\pm$0.01 & 0.01$\pm$0.00 & 0.01$\pm$0.00 & 0.31$\pm$0.00 & \cellcolor{green!20}3.00 & 0.11$\pm$0.01 & \textbf{0.59$\pm$0.02} & \textbf{0.10$\pm$0.01} & \textbf{0.27$\pm$0.01} & \cellcolor{green!60}1.50 & 0.62$\pm$0.02 & 0.45$\pm$0.03 & \cellcolor{green!20}3.50 & \cellcolor{green!40}2.67 \\
TableVectorizer & 0.49$\pm$0.02 & 0.00$\pm$0.00 & 0.00$\pm$0.00 & 0.17$\pm$0.01 & 8.00 & 0.07$\pm$0.01 & 0.27$\pm$0.01 & 0.02$\pm$0.00 & 0.12$\pm$0.01 & \cellcolor{red!60}14.00 & 0.38$\pm$0.02 & 0.06$\pm$0.01 & 11.00 & \cellcolor{red!20}11.00 \\
HyTREL & 0.73$\pm$0.01 & 0.00$\pm$0.00 & 0.00$\pm$0.00 & 0.24$\pm$0.00 & 7.50 & 0.09$\pm$0.01 & 0.41$\pm$0.01 & 0.05$\pm$0.01 & 0.18$\pm$0.01 & 7.75 & 0.55$\pm$0.01 & 0.40$\pm$0.03 & 4.50 & 6.58 \\
Armadillo & 0.85$\pm$0.01 & 0.21$\pm$0.01 & 0.23$\pm$0.01 & 0.43$\pm$0.01 & \cellcolor{green!40}2.50 & 0.11$\pm$0.01 & 0.42$\pm$0.01 & 0.06$\pm$0.01 & 0.20$\pm$0.01 & 5.75 & 0.79$\pm$0.02 & 0.69$\pm$0.02 & \cellcolor{green!40}2.00 & \cellcolor{green!20}3.42 \\
TabPFN (mean) & 0.50$\pm$0.01 & 0.00$\pm$0.00 & 0.00$\pm$0.00 & 0.17$\pm$0.00 & 8.50 & 0.07$\pm$0.01 & 0.31$\pm$0.02 & 0.03$\pm$0.01 & 0.14$\pm$0.01 & \cellcolor{red!20}11.50 & 0.38$\pm$0.02 & 0.18$\pm$0.02 & 6.00 & 8.67 \\
TabPFN (max) & 0.60$\pm$0.01 & 0.00$\pm$0.00 & 0.00$\pm$0.00 & 0.20$\pm$0.00 & 8.00 & 0.08$\pm$0.00 & 0.32$\pm$0.01 & 0.02$\pm$0.00 & 0.14$\pm$0.00 & 11.00 & 0.34$\pm$0.02 & 0.16$\pm$0.03 & 8.50 & 9.17 \\
\textsc{ConTextTab} (mean) & 0.55$\pm$0.01 & 0.00$\pm$0.00 & 0.00$\pm$0.00 & 0.18$\pm$0.00 & 9.75 & 0.07$\pm$0.00 & 0.35$\pm$0.01 & 0.04$\pm$0.00 & 0.15$\pm$0.00 & 9.50 & 0.27$\pm$0.01 & 0.15$\pm$0.02 & 10.00 & 9.75 \\
\textsc{ConTextTab} (max) & 0.23$\pm$0.01 & 0.00$\pm$0.00 & 0.00$\pm$0.00 & 0.08$\pm$0.00 & \cellcolor{red!40}13.00 & 0.07$\pm$0.00 & 0.31$\pm$0.01 & 0.03$\pm$0.00 & 0.14$\pm$0.01 & \cellcolor{red!40}12.50 & 0.25$\pm$0.01 & 0.13$\pm$0.01 & \cellcolor{red!40}11.50 & \cellcolor{red!60}12.33 \\
\bottomrule
\end{tabular}
}
\end{table*}

\paragraph{Semantic labeling highlights global grouping strength of serialization-based text embedding models.}
Under Semantic labeling (Table~\ref{tab:results_synthetic_meta_2}), HashingSchema achieves the strongest triplet performance, implying that schema tokens correlate with the generative factors used to define the semantic labels.
However, LLMText provides the best clustering alignment by a substantial margin, indicating that semantic text encoders better preserve the global grouping structure induced by the generators, beyond local ranking.
Retrieval scores are generally high in this regime, with HashingText reaching near-perfect retrieval. Its linear probe, however, reaches only $0.63$ once folds are stratified by source table, while HashingSchema and SchemaContent stay near $0.87$--$0.88$, so part of the content-fingerprint advantage on these labels tracks source identity rather than semantic decodability (Appendix~\ref{app:lp_protocols}).

\begin{table*}[!ht]
\centering
\small
\caption{D2 Metrics for Synthetic data - Semantic Labels (mean $\pm$ 95\% CI over 10 seeds)}
\label{tab:results_synthetic_meta_2}
\resizebox{\textwidth}{!}{
\begin{tabular}{lccccc|ccccc|ccc|c}
\toprule
\multirow{2}{*}{Embedder} & \multicolumn{5}{c}{Triplet (Semantic)} & \multicolumn{5}{c}{Clustering (Semantic)} & \multicolumn{3}{c}{Retrieval/Probe (Semantic)} & \multicolumn{1}{c}{Overall} \\
\cmidrule(lr){2-6} \cmidrule(lr){7-11} \cmidrule(lr){12-14} \cmidrule(lr){15-15}
 & TR-R & TR-H & TR-CH & TR-Avg & Rank & Purity & NMI & ARI & CL-Avg & Rank & R@5 & LP & Rank & Avg Rank \\
\midrule
HashingSchema & \textbf{0.96$\pm$0.01} & \textbf{0.49$\pm$0.03} & \textbf{0.50$\pm$0.02} & \textbf{0.65$\pm$0.02} & \cellcolor{green!60}1.25 & \textbf{0.79$\pm$0.02} & 0.85$\pm$0.02 & 0.70$\pm$0.04 & 0.78$\pm$0.03 & \cellcolor{green!40}2.00 & 0.99$\pm$0.00 & 0.88$\pm$0.02 & \cellcolor{green!40}2.50 & \cellcolor{green!40}1.92 \\
SchemaContent & 0.02$\pm$0.00 & 0.00$\pm$0.00 & 0.00$\pm$0.00 & 0.01$\pm$0.00 & \cellcolor{red!60}13.50 & 0.41$\pm$0.02 & 0.45$\pm$0.02 & 0.21$\pm$0.02 & 0.36$\pm$0.02 & 9.50 & 0.98$\pm$0.00 & 0.87$\pm$0.02 & \cellcolor{green!20}3.50 & 8.83 \\
TableStatistics & 0.40$\pm$0.02 & 0.00$\pm$0.00 & 0.00$\pm$0.00 & 0.13$\pm$0.01 & \cellcolor{red!20}12.50 & 0.53$\pm$0.02 & 0.61$\pm$0.03 & 0.42$\pm$0.03 & 0.52$\pm$0.02 & 5.25 & 0.92$\pm$0.01 & 0.58$\pm$0.03 & 9.50 & 9.08 \\
StatisticalSummary & 0.83$\pm$0.01 & 0.02$\pm$0.01 & 0.03$\pm$0.02 & 0.30$\pm$0.01 & 5.25 & 0.59$\pm$0.03 & 0.66$\pm$0.02 & 0.45$\pm$0.02 & 0.57$\pm$0.02 & 3.75 & 0.97$\pm$0.00 & 0.68$\pm$0.03 & 5.50 & 4.83 \\
MatrixFactorization & 0.71$\pm$0.03 & 0.00$\pm$0.00 & 0.00$\pm$0.00 & 0.24$\pm$0.01 & 8.25 & 0.56$\pm$0.03 & 0.56$\pm$0.02 & 0.29$\pm$0.04 & 0.47$\pm$0.03 & 5.75 & 0.89$\pm$0.03 & 0.51$\pm$0.02 & \cellcolor{red!20}12.00 & 8.67 \\
HashingText & 0.78$\pm$0.01 & 0.11$\pm$0.01 & 0.16$\pm$0.01 & 0.35$\pm$0.01 & \cellcolor{green!20}3.25 & 0.59$\pm$0.03 & 0.69$\pm$0.03 & 0.50$\pm$0.05 & 0.59$\pm$0.03 & \cellcolor{green!20}3.25 & \textbf{1.00$\pm$0.00} & 0.67$\pm$0.09 & 4.50 & \cellcolor{green!20}3.67 \\
LLMText & \textbf{0.96$\pm$0.01} & 0.46$\pm$0.03 & 0.48$\pm$0.04 & 0.64$\pm$0.03 & \cellcolor{green!40}1.75 & \textbf{0.79$\pm$0.04} & \textbf{0.88$\pm$0.02} & \textbf{0.72$\pm$0.06} & \textbf{0.80$\pm$0.04} & \cellcolor{green!60}1.00 & \textbf{0.99$\pm$0.00} & \textbf{0.91$\pm$0.02} & \cellcolor{green!60}1.50 & \cellcolor{green!60}1.42 \\
TableVectorizer & 0.51$\pm$0.02 & 0.04$\pm$0.01 & 0.04$\pm$0.01 & 0.20$\pm$0.01 & 7.50 & 0.38$\pm$0.03 & 0.29$\pm$0.02 & 0.14$\pm$0.01 & 0.27$\pm$0.02 & \cellcolor{red!60}13.75 & 0.90$\pm$0.01 & 0.35$\pm$0.03 & \cellcolor{red!40}12.00 & \cellcolor{red!40}11.08 \\
HyTREL & 0.74$\pm$0.01 & 0.03$\pm$0.01 & 0.04$\pm$0.01 & 0.27$\pm$0.01 & 5.50 & 0.50$\pm$0.02 & 0.50$\pm$0.02 & 0.28$\pm$0.03 & 0.43$\pm$0.02 & 7.00 & 0.97$\pm$0.00 & 0.83$\pm$0.03 & 5.00 & 5.83 \\
Armadillo & 0.69$\pm$0.01 & 0.05$\pm$0.00 & 0.05$\pm$0.01 & 0.26$\pm$0.01 & 5.25 & 0.44$\pm$0.03 & 0.39$\pm$0.02 & 0.26$\pm$0.03 & 0.36$\pm$0.02 & 9.00 & 0.92$\pm$0.01 & 0.50$\pm$0.11 & 10.00 & 8.08 \\
TabPFN (mean) & 0.51$\pm$0.01 & 0.00$\pm$0.00 & 0.00$\pm$0.00 & 0.17$\pm$0.00 & 9.75 & 0.38$\pm$0.03 & 0.38$\pm$0.03 & 0.17$\pm$0.03 & 0.31$\pm$0.03 & \cellcolor{red!40}13.00 & 0.89$\pm$0.01 & 0.61$\pm$0.02 & 10.00 & \cellcolor{red!20}10.92 \\
TabPFN (max) & 0.61$\pm$0.01 & 0.00$\pm$0.00 & 0.00$\pm$0.00 & 0.21$\pm$0.00 & 8.00 & 0.39$\pm$0.02 & 0.39$\pm$0.01 & 0.18$\pm$0.02 & 0.32$\pm$0.01 & 11.50 & 0.88$\pm$0.01 & 0.58$\pm$0.01 & \cellcolor{red!60}12.00 & 10.50 \\
\textsc{ConTextTab} (mean) & 0.55$\pm$0.01 & 0.00$\pm$0.00 & 0.00$\pm$0.00 & 0.18$\pm$0.00 & 10.25 & 0.44$\pm$0.02 & 0.41$\pm$0.01 & 0.25$\pm$0.01 & 0.37$\pm$0.01 & 8.50 & 0.92$\pm$0.01 & 0.75$\pm$0.02 & 7.00 & 8.58 \\
\textsc{ConTextTab} (max) & 0.23$\pm$0.01 & 0.00$\pm$0.00 & 0.00$\pm$0.00 & 0.08$\pm$0.00 & \cellcolor{red!40}13.00 & 0.40$\pm$0.03 & 0.35$\pm$0.03 & 0.21$\pm$0.03 & 0.32$\pm$0.02 & \cellcolor{red!20}11.75 & 0.88$\pm$0.01 & 0.67$\pm$0.02 & 10.00 & \cellcolor{red!60}11.58 \\
\bottomrule
\end{tabular}
}
\end{table*}

\paragraph{Semantic+Difficulty labeling reveals schema-only limitations under value-complexity shifts.}
Under Semantic+Difficulty labeling (Table~\ref{tab:results_synthetic_meta_3}), we refine semantic groups by adding a difficulty factor, so tables with the same schema can receive different labels depending on content complexity.
Triplet scores drop broadly, indicating that many methods fail to distinguish partitions based on value-level complexity.
The degradation is most pronounced for schema-centric hashing, with HashingSchema largely collapsing on TR-H and TR-CH, suggesting that difficulty is not recoverable from schema tokens alone and is instead expressed through value distributions and nonlinear relationships.
In this harder setting, HashingText provides the strongest triplet ranking (including hard and cluster-hard negatives), whereas LLMText again achieves the best clustering alignment. 
This reinforces a recurring trade-off between content fingerprints which excel at fine-grained neighbor ranking, and serialization-based methods that stabilize global grouping.
The Semantic+Difficulty labeling scheme supports one of the central narratives of this paper, namely that incorporating difficulty granularity into semantic labeling renders evaluation sensitive to whether embeddings capture numerical scale and complexity, as opposed to merely shallow schema-derived features.

\begin{table*}[!ht]
\centering
\small
\caption{D2 Metrics for Synthetic data - Semantic+Difficulty Labels (mean $\pm$ 95\% CI over 10 seeds)}
\label{tab:results_synthetic_meta_3}
\resizebox{\textwidth}{!}{
\begin{tabular}{lccccc|ccccc|ccc|c}
\toprule
\multirow{2}{*}{Embedder} & \multicolumn{5}{c}{Triplet (Semantic+Difficulty)} & \multicolumn{5}{c}{Clustering (Semantic+Difficulty)} & \multicolumn{3}{c}{Retrieval/Probe (Semantic+Difficulty)} & \multicolumn{1}{c}{Overall} \\
\cmidrule(lr){2-6} \cmidrule(lr){7-11} \cmidrule(lr){12-14} \cmidrule(lr){15-15}
 & TR-R & TR-H & TR-CH & TR-Avg & Rank & Purity & NMI & ARI & CL-Avg & Rank & R@5 & LP & Rank & Avg Rank \\
\midrule
HashingSchema & 0.91$\pm$0.00 & 0.01$\pm$0.01 & 0.01$\pm$0.01 & 0.31$\pm$0.00 & 4.25 & \textbf{0.40$\pm$0.02} & 0.70$\pm$0.02 & 0.35$\pm$0.02 & 0.48$\pm$0.02 & \cellcolor{green!40}2.00 & 0.62$\pm$0.03 & 0.44$\pm$0.03 & 9.00 & 5.08 \\
SchemaContent & 0.02$\pm$0.00 & 0.00$\pm$0.00 & 0.00$\pm$0.00 & 0.01$\pm$0.00 & \cellcolor{red!60}13.38 & 0.21$\pm$0.01 & 0.38$\pm$0.02 & 0.12$\pm$0.01 & 0.24$\pm$0.02 & 9.50 & 0.79$\pm$0.02 & 0.43$\pm$0.03 & 6.00 & 9.62 \\
TableStatistics & 0.38$\pm$0.02 & 0.00$\pm$0.00 & 0.00$\pm$0.00 & 0.13$\pm$0.01 & \cellcolor{red!20}12.38 & 0.27$\pm$0.01 & 0.52$\pm$0.02 & 0.21$\pm$0.02 & 0.33$\pm$0.02 & 5.25 & 0.64$\pm$0.01 & 0.31$\pm$0.02 & \cellcolor{red!40}12.00 & 9.88 \\
StatisticalSummary & 0.81$\pm$0.01 & 0.01$\pm$0.01 & 0.01$\pm$0.01 & 0.28$\pm$0.01 & 4.25 & 0.31$\pm$0.01 & 0.59$\pm$0.01 & 0.24$\pm$0.01 & 0.38$\pm$0.01 & 4.00 & 0.91$\pm$0.01 & 0.40$\pm$0.03 & 4.50 & \cellcolor{green!20}4.25 \\
MatrixFactorization & 0.70$\pm$0.03 & 0.00$\pm$0.00 & 0.00$\pm$0.00 & 0.23$\pm$0.01 & 8.25 & 0.30$\pm$0.02 & 0.50$\pm$0.01 & 0.15$\pm$0.02 & 0.32$\pm$0.02 & 5.75 & 0.78$\pm$0.02 & 0.39$\pm$0.03 & 7.50 & 7.17 \\
HashingText & 0.83$\pm$0.02 & \textbf{0.23$\pm$0.01} & \textbf{0.25$\pm$0.01} & \textbf{0.44$\pm$0.01} & \cellcolor{green!60}1.50 & 0.34$\pm$0.01 & 0.63$\pm$0.02 & 0.27$\pm$0.02 & 0.41$\pm$0.02 & \cellcolor{green!20}3.00 & \textbf{1.00$\pm$0.00} & \textbf{0.99$\pm$0.00} & \cellcolor{green!60}1.00 & \cellcolor{green!60}1.83 \\
LLMText & \textbf{0.93$\pm$0.01} & 0.01$\pm$0.01 & 0.01$\pm$0.01 & 0.32$\pm$0.01 & \cellcolor{green!40}2.75 & \textbf{0.40$\pm$0.02} & \textbf{0.73$\pm$0.02} & \textbf{0.37$\pm$0.03} & \textbf{0.50$\pm$0.02} & \cellcolor{green!60}1.00 & 0.90$\pm$0.01 & 0.72$\pm$0.04 & \cellcolor{green!40}2.50 & \cellcolor{green!40}2.08 \\
TableVectorizer & 0.50$\pm$0.02 & 0.01$\pm$0.01 & 0.01$\pm$0.01 & 0.17$\pm$0.01 & 7.25 & 0.20$\pm$0.01 & 0.26$\pm$0.02 & 0.07$\pm$0.01 & 0.18$\pm$0.01 & \cellcolor{red!60}13.75 & 0.71$\pm$0.02 & 0.18$\pm$0.02 & \cellcolor{red!20}11.00 & \cellcolor{red!40}10.67 \\
HyTREL & 0.72$\pm$0.01 & 0.00$\pm$0.00 & 0.00$\pm$0.00 & 0.24$\pm$0.00 & 7.50 & 0.26$\pm$0.01 & 0.43$\pm$0.01 & 0.15$\pm$0.01 & 0.28$\pm$0.01 & 7.00 & 0.84$\pm$0.01 & 0.67$\pm$0.03 & 4.00 & 6.17 \\
Armadillo & 0.71$\pm$0.02 & 0.07$\pm$0.01 & 0.08$\pm$0.01 & 0.29$\pm$0.01 & \cellcolor{green!20}3.50 & 0.23$\pm$0.02 & 0.38$\pm$0.02 & 0.15$\pm$0.02 & 0.25$\pm$0.02 & 8.25 & 0.85$\pm$0.01 & 0.65$\pm$0.02 & \cellcolor{green!20}4.00 & 5.25 \\
TabPFN (mean) & 0.51$\pm$0.01 & 0.00$\pm$0.00 & 0.00$\pm$0.00 & 0.17$\pm$0.00 & 9.00 & 0.20$\pm$0.02 & 0.33$\pm$0.02 & 0.08$\pm$0.02 & 0.20$\pm$0.02 & \cellcolor{red!40}13.00 & 0.67$\pm$0.02 & 0.35$\pm$0.03 & 9.50 & \cellcolor{red!20}10.50 \\
TabPFN (max) & 0.61$\pm$0.01 & 0.00$\pm$0.00 & 0.00$\pm$0.00 & 0.20$\pm$0.00 & 7.75 & 0.20$\pm$0.01 & 0.34$\pm$0.01 & 0.09$\pm$0.01 & 0.21$\pm$0.01 & \cellcolor{red!20}11.75 & 0.65$\pm$0.01 & 0.33$\pm$0.03 & 11.00 & 10.17 \\
\textsc{ConTextTab} (mean) & 0.54$\pm$0.01 & 0.00$\pm$0.00 & 0.00$\pm$0.00 & 0.18$\pm$0.00 & 10.38 & 0.23$\pm$0.02 & 0.36$\pm$0.01 & 0.14$\pm$0.01 & 0.24$\pm$0.01 & 9.25 & 0.64$\pm$0.01 & 0.37$\pm$0.02 & 10.50 & 10.04 \\
\textsc{ConTextTab} (max) & 0.23$\pm$0.01 & 0.00$\pm$0.00 & 0.00$\pm$0.00 & 0.08$\pm$0.00 & \cellcolor{red!40}12.88 & 0.21$\pm$0.01 & 0.31$\pm$0.02 & 0.11$\pm$0.01 & 0.21$\pm$0.01 & 11.50 & 0.60$\pm$0.01 & 0.34$\pm$0.02 & \cellcolor{red!60}12.50 & \cellcolor{red!60}12.29 \\
\bottomrule
\end{tabular}
}
\end{table*}

\paragraph{Stat labeling shows that retrieval and linear probing for majority-type labels are near-perfect, leaving triplet and clustering to separate methods.}
Under the Stat labeling scheme (Table~\ref{tab:results_synthetic_stat}), retrieval and linear probe are near-saturated for many methods, indicating that coarse statistical type signatures are easy to identify under the current partition budget.
Therefore, triplet ranking and clustering become the primary discriminators in this regime.
HashingSchema dominates overall, consistent with type- and schema-token features being sufficient for coarse statistical grouping.
LLMText remains competitive, especially in triplet ranking, while content- and factorization-based embedding models vary more widely, reflecting that some methods preserve coarse type separability without forming clean global cluster structure.

\begin{table*}[!ht]
\centering
\small
\caption{D2 Metrics for Synthetic data - Stat Labels (mean $\pm$ 95\% CI over 10 seeds)}
\label{tab:results_synthetic_stat}
\resizebox{\textwidth}{!}{
\begin{tabular}{lccccc|ccccc|ccc|c}
\toprule
\multirow{2}{*}{Embedder} & \multicolumn{5}{c}{Triplet (Stat)} & \multicolumn{5}{c}{Clustering (Stat)} & \multicolumn{3}{c}{Retrieval/Probe (Stat)} & \multicolumn{1}{c}{Overall} \\
\cmidrule(lr){2-6} \cmidrule(lr){7-11} \cmidrule(lr){12-14} \cmidrule(lr){15-15}
 & TR-R & TR-H & TR-CH & TR-Avg & Rank & Purity & NMI & ARI & CL-Avg & Rank & R@5 & LP & Rank & Avg Rank \\
\midrule
HashingSchema & \textbf{0.96$\pm$0.01} & \textbf{0.80$\pm$0.02} & \textbf{0.80$\pm$0.02} & \textbf{0.86$\pm$0.02} & \cellcolor{green!60}1.00 & \textbf{0.96$\pm$0.01} & \textbf{0.55$\pm$0.03} & \textbf{0.42$\pm$0.04} & \textbf{0.64$\pm$0.02} & \cellcolor{green!60}1.25 & \textbf{1.00$\pm$0.00} & \textbf{1.00$\pm$0.00} & \cellcolor{green!60}1.00 & \cellcolor{green!60}1.08 \\
SchemaContent & 0.02$\pm$0.00 & 0.00$\pm$0.00 & 0.00$\pm$0.00 & 0.01$\pm$0.00 & \cellcolor{red!60}14.00 & 0.78$\pm$0.02 & 0.07$\pm$0.05 & 0.05$\pm$0.03 & 0.30$\pm$0.03 & \cellcolor{red!40}12.50 & \textbf{1.00$\pm$0.00} & \textbf{0.99$\pm$0.00} & \cellcolor{green!40}2.50 & 9.67 \\
TableStatistics & 0.24$\pm$0.03 & 0.07$\pm$0.02 & 0.07$\pm$0.02 & 0.13$\pm$0.02 & 9.75 & 0.77$\pm$0.03 & 0.18$\pm$0.02 & -0.06$\pm$0.04 & 0.30$\pm$0.02 & \cellcolor{red!20}11.75 & \textbf{0.99$\pm$0.00} & \textbf{0.99$\pm$0.00} & 5.00 & 8.83 \\
StatisticalSummary & 0.86$\pm$0.03 & 0.45$\pm$0.03 & 0.47$\pm$0.04 & 0.59$\pm$0.03 & \cellcolor{green!20}3.00 & 0.93$\pm$0.01 & 0.44$\pm$0.03 & 0.38$\pm$0.05 & 0.59$\pm$0.02 & 3.25 & 0.99$\pm$0.00 & 0.99$\pm$0.00 & 6.00 & \cellcolor{green!20}4.08 \\
MatrixFactorization & 0.44$\pm$0.03 & 0.12$\pm$0.02 & 0.12$\pm$0.02 & 0.23$\pm$0.03 & 6.25 & 0.79$\pm$0.02 & 0.14$\pm$0.02 & -0.02$\pm$0.02 & 0.30$\pm$0.01 & 10.50 & 0.97$\pm$0.01 & 0.92$\pm$0.01 & \cellcolor{red!20}11.50 & 9.42 \\
HashingText & 0.55$\pm$0.02 & 0.13$\pm$0.02 & 0.24$\pm$0.05 & 0.31$\pm$0.02 & 4.50 & 0.94$\pm$0.03 & 0.47$\pm$0.07 & 0.37$\pm$0.07 & 0.59$\pm$0.06 & \cellcolor{green!40}2.75 & \textbf{0.99$\pm$0.00} & 0.94$\pm$0.01 & 6.50 & 4.58 \\
LLMText & 0.94$\pm$0.01 & 0.71$\pm$0.03 & 0.71$\pm$0.03 & 0.79$\pm$0.02 & \cellcolor{green!40}2.00 & \textbf{0.96$\pm$0.01} & 0.49$\pm$0.05 & 0.31$\pm$0.07 & 0.59$\pm$0.04 & \cellcolor{green!20}2.75 & \textbf{1.00$\pm$0.00} & \textbf{1.00$\pm$0.00} & \cellcolor{green!20}2.50 & \cellcolor{green!40}2.42 \\
TableVectorizer & 0.39$\pm$0.02 & 0.04$\pm$0.01 & 0.06$\pm$0.01 & 0.16$\pm$0.01 & 9.75 & 0.78$\pm$0.02 & 0.09$\pm$0.03 & -0.00$\pm$0.03 & 0.29$\pm$0.01 & \cellcolor{red!60}12.75 & 0.97$\pm$0.00 & 0.92$\pm$0.01 & 10.50 & \cellcolor{red!60}11.00 \\
HyTREL & 0.57$\pm$0.03 & 0.04$\pm$0.01 & 0.04$\pm$0.01 & 0.22$\pm$0.02 & 8.25 & 0.84$\pm$0.03 & 0.23$\pm$0.03 & 0.22$\pm$0.03 & 0.43$\pm$0.03 & 5.25 & \textbf{0.99$\pm$0.00} & 0.97$\pm$0.01 & 5.50 & 6.33 \\
Armadillo & 0.62$\pm$0.02 & 0.10$\pm$0.01 & 0.11$\pm$0.01 & 0.28$\pm$0.01 & 5.25 & 0.84$\pm$0.03 & 0.24$\pm$0.04 & 0.17$\pm$0.04 & 0.41$\pm$0.03 & 6.00 & 0.98$\pm$0.00 & 0.85$\pm$0.02 & 10.50 & 7.25 \\
TabPFN (mean) & 0.43$\pm$0.03 & 0.03$\pm$0.01 & 0.03$\pm$0.01 & 0.16$\pm$0.01 & \cellcolor{red!20}11.00 & 0.80$\pm$0.02 & 0.12$\pm$0.03 & 0.17$\pm$0.07 & 0.36$\pm$0.04 & 8.25 & 0.97$\pm$0.00 & 0.85$\pm$0.02 & \cellcolor{red!60}13.50 & \cellcolor{red!40}10.92 \\
TabPFN (max) & 0.52$\pm$0.01 & 0.06$\pm$0.02 & 0.08$\pm$0.04 & 0.22$\pm$0.02 & 7.50 & 0.79$\pm$0.02 & 0.12$\pm$0.02 & 0.06$\pm$0.03 & 0.32$\pm$0.02 & 9.75 & 0.97$\pm$0.01 & 0.86$\pm$0.02 & \cellcolor{red!40}13.00 & \cellcolor{red!20}10.08 \\
\textsc{ConTextTab} (mean) & 0.46$\pm$0.01 & 0.02$\pm$0.00 & 0.05$\pm$0.02 & 0.18$\pm$0.01 & 9.75 & 0.81$\pm$0.03 & 0.10$\pm$0.07 & 0.04$\pm$0.06 & 0.32$\pm$0.05 & 10.25 & 0.98$\pm$0.00 & 0.96$\pm$0.01 & 8.00 & 9.33 \\
\textsc{ConTextTab} (max) & 0.22$\pm$0.01 & 0.00$\pm$0.00 & 0.03$\pm$0.03 & 0.08$\pm$0.01 & \cellcolor{red!40}13.00 & 0.83$\pm$0.03 & 0.14$\pm$0.07 & 0.08$\pm$0.06 & 0.35$\pm$0.05 & 8.00 & 0.98$\pm$0.00 & 0.95$\pm$0.01 & 9.00 & 10.00 \\
\bottomrule
\end{tabular}
}
\end{table*}

\paragraph{Averaging across labels hides specialization, but preserves the big picture.}
Averaging over label types (Table~\ref{tab:results_synthetic_appendix_avg}) clarifies why no single method consistently performs best across D2 proxy tasks.
LLMText achieves the best overall D2 ranking due to consistently strong clustering under the Semantic labeling scheme, while HashingText is the best on retrieval, linear probe and remains most reliable for Direct-style identity ranking under hard negatives.
In contrast, HashingSchema ranks best on average triplet performance but is comparatively weaker on retrieval and linear probe, highlighting that schema fingerprints can match semantic groupings without yielding robust instance-level neighborhoods.
This decomposition explains the main-table D2 average scores, where different labeling schemes reflect different notions of table similarity, and Semantic+Difficulty specifically reveals where schema features stop being sufficient.

\begin{table*}[!ht]
\centering
\small
\caption{Detailed D2 Metrics for Synthetic data - Averaged over Label Types (mean $\pm$ 95\% CI over 10 seeds)}
\label{tab:results_synthetic_appendix_avg}
\resizebox{\textwidth}{!}{
\begin{tabular}{lccccc|ccccc|ccc|c}
\toprule
\multirow{2}{*}{Embedder} & \multicolumn{5}{c}{Triplet (Avg over Labels)} & \multicolumn{5}{c}{Clustering (Avg over Labels)} & \multicolumn{3}{c}{Retrieval/Probe (Avg over Labels)} & \multicolumn{1}{c}{Overall} \\
\cmidrule(lr){2-6} \cmidrule(lr){7-11} \cmidrule(lr){12-14} \cmidrule(lr){15-15}
 & TR-R & TR-H & TR-CH & TR-Avg & Rank & Purity & NMI & ARI & CL-Avg & Rank & R@5 & LP & Rank & Avg Rank \\
\midrule
HashingSchema & \textbf{0.93$\pm$0.01} & \textbf{0.33$\pm$0.01} & 0.33$\pm$0.01 & \textbf{0.53$\pm$0.01} & \cellcolor{green!60}1.50 & \textbf{0.56$\pm$0.02} & \textbf{0.67$\pm$0.02} & \textbf{0.39$\pm$0.03} & \textbf{0.54$\pm$0.02} & \cellcolor{green!60}1.50 & 0.71$\pm$0.01 & 0.61$\pm$0.02 & 8.50 & \cellcolor{green!20}3.83 \\
SchemaContent & 0.02$\pm$0.00 & 0.00$\pm$0.00 & 0.00$\pm$0.00 & 0.01$\pm$0.00 & \cellcolor{red!60}14.00 & 0.37$\pm$0.02 & 0.31$\pm$0.03 & 0.10$\pm$0.02 & 0.26$\pm$0.02 & 10.50 & 0.78$\pm$0.01 & 0.60$\pm$0.02 & 6.50 & 10.33 \\
TableStatistics & 0.35$\pm$0.02 & 0.02$\pm$0.00 & 0.02$\pm$0.00 & 0.13$\pm$0.01 & 10.75 & 0.42$\pm$0.02 & 0.45$\pm$0.02 & 0.16$\pm$0.02 & 0.34$\pm$0.02 & 5.75 & 0.71$\pm$0.01 & 0.50$\pm$0.02 & 10.00 & 8.83 \\
StatisticalSummary & 0.83$\pm$0.02 & 0.12$\pm$0.01 & 0.13$\pm$0.02 & 0.36$\pm$0.01 & 3.75 & 0.48$\pm$0.02 & 0.55$\pm$0.02 & 0.29$\pm$0.02 & 0.44$\pm$0.01 & 4.00 & 0.91$\pm$0.01 & 0.56$\pm$0.02 & 4.50 & 4.08 \\
MatrixFactorization & 0.64$\pm$0.03 & 0.03$\pm$0.01 & 0.03$\pm$0.01 & 0.23$\pm$0.01 & 6.50 & 0.43$\pm$0.02 & 0.41$\pm$0.02 & 0.12$\pm$0.02 & 0.32$\pm$0.02 & 6.75 & 0.79$\pm$0.02 & 0.49$\pm$0.02 & 9.00 & 7.42 \\
HashingText & 0.79$\pm$0.01 & 0.29$\pm$0.01 & \textbf{0.34$\pm$0.02} & 0.47$\pm$0.01 & \cellcolor{green!20}2.75 & 0.50$\pm$0.02 & 0.59$\pm$0.03 & 0.31$\pm$0.04 & 0.46$\pm$0.03 & \cellcolor{green!20}3.00 & \textbf{1.00$\pm$0.00} & \textbf{0.90$\pm$0.03} & \cellcolor{green!60}1.00 & \cellcolor{green!40}2.25 \\
LLMText & \textbf{0.94$\pm$0.01} & 0.30$\pm$0.02 & 0.30$\pm$0.02 & 0.51$\pm$0.01 & \cellcolor{green!40}2.00 & \textbf{0.57$\pm$0.02} & \textbf{0.67$\pm$0.03} & 0.37$\pm$0.05 & \textbf{0.54$\pm$0.03} & \cellcolor{green!40}1.50 & 0.88$\pm$0.01 & 0.77$\pm$0.02 & \cellcolor{green!40}3.00 & \cellcolor{green!60}2.17 \\
TableVectorizer & 0.47$\pm$0.02 & 0.02$\pm$0.01 & 0.03$\pm$0.01 & 0.17$\pm$0.01 & 8.75 & 0.36$\pm$0.02 & 0.23$\pm$0.02 & 0.06$\pm$0.01 & 0.22$\pm$0.01 & \cellcolor{red!60}14.00 & 0.74$\pm$0.01 & 0.38$\pm$0.02 & \cellcolor{red!20}11.00 & \cellcolor{red!40}11.25 \\
HyTREL & 0.69$\pm$0.01 & 0.02$\pm$0.01 & 0.02$\pm$0.01 & 0.24$\pm$0.01 & 7.25 & 0.42$\pm$0.02 & 0.39$\pm$0.02 & 0.18$\pm$0.02 & 0.33$\pm$0.02 & 6.00 & 0.84$\pm$0.01 & 0.72$\pm$0.02 & 4.00 & 5.75 \\
Armadillo & 0.72$\pm$0.01 & 0.11$\pm$0.01 & 0.12$\pm$0.01 & 0.31$\pm$0.01 & 5.00 & 0.41$\pm$0.02 & 0.36$\pm$0.02 & 0.16$\pm$0.02 & 0.31$\pm$0.02 & 7.75 & 0.89$\pm$0.01 & 0.67$\pm$0.04 & \cellcolor{green!20}3.50 & 5.42 \\
TabPFN (mean) & 0.49$\pm$0.01 & 0.01$\pm$0.00 & 0.01$\pm$0.00 & 0.17$\pm$0.00 & \cellcolor{red!20}11.00 & 0.36$\pm$0.02 & 0.28$\pm$0.02 & 0.11$\pm$0.03 & 0.25$\pm$0.02 & \cellcolor{red!20}11.75 & 0.73$\pm$0.01 & 0.50$\pm$0.02 & 10.00 & \cellcolor{red!20}10.92 \\
TabPFN (max) & 0.59$\pm$0.01 & 0.02$\pm$0.01 & 0.02$\pm$0.01 & 0.21$\pm$0.01 & 8.50 & 0.36$\pm$0.01 & 0.29$\pm$0.01 & 0.09$\pm$0.01 & 0.25$\pm$0.01 & \cellcolor{red!40}12.25 & 0.71$\pm$0.01 & 0.48$\pm$0.02 & \cellcolor{red!60}12.00 & 10.92 \\
\textsc{ConTextTab} (mean) & 0.53$\pm$0.01 & 0.00$\pm$0.00 & 0.01$\pm$0.01 & 0.18$\pm$0.00 & 10.25 & 0.39$\pm$0.02 & 0.31$\pm$0.03 & 0.12$\pm$0.02 & 0.27$\pm$0.02 & 9.00 & 0.70$\pm$0.01 & 0.55$\pm$0.02 & 10.50 & 9.92 \\
\textsc{ConTextTab} (max) & 0.23$\pm$0.01 & 0.00$\pm$0.00 & 0.01$\pm$0.01 & 0.08$\pm$0.01 & \cellcolor{red!40}13.00 & 0.38$\pm$0.02 & 0.28$\pm$0.03 & 0.11$\pm$0.03 & 0.26$\pm$0.02 & 11.25 & 0.68$\pm$0.01 & 0.52$\pm$0.01 & \cellcolor{red!40}11.50 & \cellcolor{red!60}11.92 \\
\bottomrule
\end{tabular}
}
\end{table*}

\subsubsection{Real data results}

Across open-source corpora, D2 scores remain strongly label-dependent and separate methods by \emph{what notion of similarity they preserve}, namely instance identity (Direct), benchmark/task semantics (Semantic), or coarse type signatures (Stat). We therefore report label-wise D2 tables below and the averaged summary (Table~\ref{tab:results_real_appendix_avg}) to make the specialization pattern explicit rather than relying on a single aggregated number.

\paragraph{Direct labeling shows partial collapse for hard and cluster-hard negatives rather than the complete collapse.}
Under Direct labeling scheme (Table~\ref{tab:results_real_direct}), several methods attain non-trivial performance on TR-H and TR-CH, indicating that the identity task is challenging but not degenerate for open-source tables. 
Overall, LLMText attains the strongest TR-Avg and clustering scores, while HashingText remains among the best retrieval methods and gives the strongest linear-probe performance, consistent with content-aware signals being reliable for separating confusable tables.
Schema-driven fingerprints (e.g., HashingSchema) remain competitive and substantially outperform purely statistical baselines under hard negatives, but still trail the content/serialization methods, suggesting that schema features alone cannot fully resolve similar tables that share column types and structural templates.
Model-pooled embeddings (TabPFN, ConTextTab) remain weak on hard negatives and clustering, indicating that pooling yields coarse neighborhoods that are insufficient for strict identity ranking even when some separability exists.

\begin{table*}[!ht]
\centering
\small
\caption{D2 Metrics for Real data - Direct Labels (mean $\pm$ 95\% CI over 10 seeds)}
\label{tab:results_real_direct}
\resizebox{\textwidth}{!}{
\begin{tabular}{lccccc|ccccc|ccc|c}
\toprule
\multirow{2}{*}{Embedder} & \multicolumn{5}{c}{Triplet (Direct)} & \multicolumn{5}{c}{Clustering (Direct)} & \multicolumn{3}{c}{Retrieval/Probe (Direct)} & \multicolumn{1}{c}{Overall} \\
\cmidrule(lr){2-6} \cmidrule(lr){7-11} \cmidrule(lr){12-14} \cmidrule(lr){15-15}
 & TR-R & TR-H & TR-CH & TR-Avg & Rank & Purity & NMI & ARI & CL-Avg & Rank & R@5 & LP & Rank & Avg Rank \\
\midrule
HashingSchema & 0.95$\pm$0.01 & 0.26$\pm$0.03 & 0.27$\pm$0.03 & 0.49$\pm$0.02 & \cellcolor{green!20}3.00 & 0.12$\pm$0.00 & 0.49$\pm$0.01 & 0.05$\pm$0.00 & 0.22$\pm$0.01 & \cellcolor{green!40}3.50 & 0.92$\pm$0.02 & 0.73$\pm$0.03 & \cellcolor{green!40}2.50 & \cellcolor{green!20}3.00 \\
SchemaContent & 0.35$\pm$0.02 & 0.00$\pm$0.00 & 0.00$\pm$0.00 & 0.12$\pm$0.01 & \cellcolor{red!40}13.12 & 0.08$\pm$0.00 & 0.32$\pm$0.02 & 0.02$\pm$0.00 & 0.14$\pm$0.01 & 11.50 & 0.64$\pm$0.01 & 0.71$\pm$0.03 & 5.00 & 9.88 \\
TableStatistics & 0.29$\pm$0.02 & 0.00$\pm$0.00 & 0.00$\pm$0.00 & 0.10$\pm$0.01 & \cellcolor{red!60}13.88 & 0.11$\pm$0.01 & 0.45$\pm$0.01 & 0.07$\pm$0.00 & 0.21$\pm$0.01 & 4.75 & 0.59$\pm$0.02 & 0.31$\pm$0.03 & 8.50 & 9.04 \\
StatisticalSummary & 0.82$\pm$0.01 & 0.05$\pm$0.01 & 0.06$\pm$0.01 & 0.31$\pm$0.01 & 5.00 & 0.11$\pm$0.01 & 0.45$\pm$0.02 & 0.07$\pm$0.01 & 0.21$\pm$0.01 & \cellcolor{green!20}3.50 & 0.74$\pm$0.02 & 0.29$\pm$0.02 & 8.00 & 5.50 \\
MatrixFactorization & 0.71$\pm$0.01 & 0.00$\pm$0.00 & 0.00$\pm$0.00 & 0.24$\pm$0.00 & 9.50 & 0.10$\pm$0.01 & 0.38$\pm$0.01 & 0.04$\pm$0.00 & 0.18$\pm$0.01 & 9.25 & 0.45$\pm$0.02 & 0.09$\pm$0.01 & \cellcolor{red!20}12.00 & 10.25 \\
HashingText & 0.96$\pm$0.00 & \textbf{0.36$\pm$0.03} & \textbf{0.39$\pm$0.03} & \textbf{0.57$\pm$0.02} & \cellcolor{green!60}1.25 & 0.11$\pm$0.00 & 0.48$\pm$0.02 & 0.05$\pm$0.01 & 0.22$\pm$0.01 & 4.00 & \textbf{0.95$\pm$0.01} & \textbf{0.78$\pm$0.02} & \cellcolor{green!60}1.00 & \cellcolor{green!40}2.08 \\
LLMText & \textbf{0.98$\pm$0.00} & 0.35$\pm$0.02 & 0.36$\pm$0.02 & \textbf{0.57$\pm$0.01} & \cellcolor{green!40}1.75 & \textbf{0.12$\pm$0.01} & \textbf{0.60$\pm$0.01} & \textbf{0.11$\pm$0.01} & \textbf{0.28$\pm$0.01} & \cellcolor{green!60}1.00 & \textbf{0.94$\pm$0.01} & 0.71$\pm$0.04 & \cellcolor{green!20}3.00 & \cellcolor{green!60}1.92 \\
TableVectorizer & 0.53$\pm$0.01 & 0.03$\pm$0.00 & 0.03$\pm$0.01 & 0.20$\pm$0.01 & 7.50 & 0.07$\pm$0.01 & 0.29$\pm$0.01 & 0.02$\pm$0.00 & 0.13$\pm$0.01 & \cellcolor{red!60}13.25 & 0.34$\pm$0.01 & 0.07$\pm$0.01 & \cellcolor{red!60}14.00 & \cellcolor{red!40}11.58 \\
HyTREL & 0.79$\pm$0.02 & 0.02$\pm$0.00 & 0.03$\pm$0.00 & 0.28$\pm$0.01 & 6.50 & 0.10$\pm$0.01 & 0.41$\pm$0.02 & 0.05$\pm$0.01 & 0.19$\pm$0.01 & 7.00 & 0.74$\pm$0.02 & 0.57$\pm$0.03 & 5.00 & 6.17 \\
Armadillo & 0.83$\pm$0.01 & 0.17$\pm$0.02 & 0.19$\pm$0.02 & 0.40$\pm$0.02 & 4.00 & 0.11$\pm$0.01 & 0.40$\pm$0.02 & 0.06$\pm$0.01 & 0.19$\pm$0.01 & 6.00 & 0.76$\pm$0.03 & 0.52$\pm$0.03 & 5.00 & 5.00 \\
TabPFN (mean) & 0.38$\pm$0.02 & 0.01$\pm$0.00 & 0.01$\pm$0.00 & 0.13$\pm$0.01 & 10.75 & 0.08$\pm$0.01 & 0.28$\pm$0.01 & 0.02$\pm$0.00 & 0.13$\pm$0.00 & \cellcolor{red!40}13.00 & 0.37$\pm$0.02 & 0.20$\pm$0.01 & \cellcolor{red!40}12.00 & \cellcolor{red!60}11.92 \\
TabPFN (max) & 0.48$\pm$0.02 & 0.01$\pm$0.00 & 0.01$\pm$0.00 & 0.17$\pm$0.01 & 9.75 & 0.08$\pm$0.01 & 0.29$\pm$0.01 & 0.02$\pm$0.00 & 0.13$\pm$0.01 & \cellcolor{red!20}12.25 & 0.36$\pm$0.01 & 0.20$\pm$0.01 & 12.00 & \cellcolor{red!20}11.33 \\
\textsc{ConTextTab} (mean) & 0.74$\pm$0.02 & 0.02$\pm$0.01 & 0.03$\pm$0.01 & 0.26$\pm$0.01 & 7.50 & 0.10$\pm$0.01 & 0.42$\pm$0.02 & 0.05$\pm$0.01 & 0.19$\pm$0.01 & 6.25 & 0.59$\pm$0.02 & 0.46$\pm$0.03 & 8.00 & 7.25 \\
\textsc{ConTextTab} (max) & 0.45$\pm$0.03 & 0.00$\pm$0.00 & 0.00$\pm$0.00 & 0.15$\pm$0.01 & \cellcolor{red!20}11.50 & 0.09$\pm$0.01 & 0.38$\pm$0.02 & 0.05$\pm$0.01 & 0.17$\pm$0.01 & 9.75 & 0.53$\pm$0.02 & 0.40$\pm$0.03 & 9.00 & 10.08 \\
\bottomrule
\end{tabular}
}
\end{table*}

\paragraph{Semantic labeling highlights the global grouping strength of serialization-based text embeddings for benchmark and task structure.}
For Semantic labeling (Table~\ref{tab:results_real_meta_2}), LLMText is the clear winner on both triplet ranking and clustering alignment, indicating that semantic encoders best preserve the global grouping induced by benchmark and task semantics.
HashingText remains highly competitive and achieves near-top retrieval and linear probe, while schema-oriented methods (HashingSchema) retain strong triplet performance but weaker clustering, consistent with capturing coarse correlates of benchmark structure without producing globally coherent clusters.

\begin{table*}[!ht]
\centering
\small
\caption{D2 Metrics for Real data - Semantic Labels (mean $\pm$ 95\% CI over 10 seeds)}
\label{tab:results_real_meta_2}
\resizebox{\textwidth}{!}{
\begin{tabular}{lccccc|ccccc|ccc|c}
\toprule
\multirow{2}{*}{Embedder} & \multicolumn{5}{c}{Triplet (Semantic)} & \multicolumn{5}{c}{Clustering (Semantic)} & \multicolumn{3}{c}{Retrieval/Probe (Semantic)} & \multicolumn{1}{c}{Overall} \\
\cmidrule(lr){2-6} \cmidrule(lr){7-11} \cmidrule(lr){12-14} \cmidrule(lr){15-15}
 & TR-R & TR-H & TR-CH & TR-Avg & Rank & Purity & NMI & ARI & CL-Avg & Rank & R@5 & LP & Rank & Avg Rank \\
\midrule
HashingSchema & 0.95$\pm$0.01 & 0.45$\pm$0.03 & 0.46$\pm$0.03 & 0.62$\pm$0.02 & \cellcolor{green!20}3.00 & 0.21$\pm$0.01 & 0.52$\pm$0.01 & 0.09$\pm$0.01 & 0.27$\pm$0.01 & 4.00 & 0.97$\pm$0.01 & 0.97$\pm$0.01 & \cellcolor{green!20}2.50 & \cellcolor{green!20}3.17 \\
SchemaContent & 0.36$\pm$0.02 & 0.00$\pm$0.00 & 0.00$\pm$0.00 & 0.12$\pm$0.01 & \cellcolor{red!40}13.12 & 0.13$\pm$0.01 & 0.32$\pm$0.02 & 0.04$\pm$0.01 & 0.16$\pm$0.01 & 11.00 & 0.76$\pm$0.02 & 0.94$\pm$0.02 & 5.50 & 9.88 \\
TableStatistics & 0.28$\pm$0.02 & 0.00$\pm$0.00 & 0.00$\pm$0.00 & 0.09$\pm$0.01 & \cellcolor{red!60}13.88 & 0.18$\pm$0.01 & 0.47$\pm$0.01 & 0.11$\pm$0.01 & 0.25$\pm$0.01 & 5.25 & 0.71$\pm$0.02 & 0.47$\pm$0.03 & 8.50 & 9.21 \\
StatisticalSummary & 0.83$\pm$0.01 & 0.12$\pm$0.01 & 0.12$\pm$0.01 & 0.36$\pm$0.01 & 5.00 & 0.20$\pm$0.01 & 0.48$\pm$0.02 & 0.12$\pm$0.02 & 0.27$\pm$0.01 & \cellcolor{green!40}3.50 & 0.83$\pm$0.01 & 0.42$\pm$0.02 & 8.00 & 5.50 \\
MatrixFactorization & 0.71$\pm$0.01 & 0.00$\pm$0.00 & 0.00$\pm$0.00 & 0.24$\pm$0.00 & 9.50 & 0.17$\pm$0.01 & 0.40$\pm$0.01 & 0.07$\pm$0.01 & 0.21$\pm$0.01 & 9.25 & 0.55$\pm$0.02 & 0.12$\pm$0.02 & \cellcolor{red!20}12.50 & 10.42 \\
HashingText & 0.96$\pm$0.01 & 0.62$\pm$0.02 & 0.66$\pm$0.02 & 0.75$\pm$0.01 & \cellcolor{green!40}2.00 & 0.22$\pm$0.01 & 0.51$\pm$0.03 & 0.09$\pm$0.02 & 0.27$\pm$0.02 & \cellcolor{green!20}3.50 & \textbf{0.99$\pm$0.00} & \textbf{0.99$\pm$0.01} & \cellcolor{green!60}1.50 & \cellcolor{green!40}2.33 \\
LLMText & \textbf{0.99$\pm$0.00} & \textbf{0.70$\pm$0.03} & \textbf{0.71$\pm$0.03} & \textbf{0.80$\pm$0.02} & \cellcolor{green!60}1.00 & \textbf{0.26$\pm$0.02} & \textbf{0.64$\pm$0.01} & \textbf{0.20$\pm$0.02} & \textbf{0.37$\pm$0.02} & \cellcolor{green!60}1.00 & \textbf{0.99$\pm$0.00} & 0.96$\pm$0.01 & \cellcolor{green!40}2.00 & \cellcolor{green!60}1.33 \\
TableVectorizer & 0.53$\pm$0.01 & 0.04$\pm$0.01 & 0.04$\pm$0.01 & 0.20$\pm$0.01 & 8.50 & 0.12$\pm$0.01 & 0.28$\pm$0.01 & 0.04$\pm$0.01 & 0.15$\pm$0.01 & \cellcolor{red!40}13.00 & 0.45$\pm$0.02 & 0.12$\pm$0.03 & \cellcolor{red!60}13.50 & \cellcolor{red!40}11.67 \\
HyTREL & 0.80$\pm$0.02 & 0.05$\pm$0.01 & 0.05$\pm$0.01 & 0.30$\pm$0.01 & 6.50 & 0.19$\pm$0.02 & 0.42$\pm$0.02 & 0.09$\pm$0.01 & 0.23$\pm$0.02 & 6.25 & 0.83$\pm$0.02 & 0.81$\pm$0.02 & 5.00 & 5.92 \\
Armadillo & 0.83$\pm$0.01 & 0.28$\pm$0.03 & 0.30$\pm$0.03 & 0.47$\pm$0.02 & 4.00 & 0.19$\pm$0.01 & 0.41$\pm$0.02 & 0.10$\pm$0.01 & 0.23$\pm$0.01 & 6.25 & 0.84$\pm$0.02 & 0.72$\pm$0.04 & 5.00 & 5.08 \\
TabPFN (mean) & 0.38$\pm$0.02 & 0.01$\pm$0.00 & 0.01$\pm$0.00 & 0.13$\pm$0.01 & 10.75 & 0.12$\pm$0.01 & 0.29$\pm$0.01 & 0.03$\pm$0.01 & 0.15$\pm$0.01 & \cellcolor{red!60}13.50 & 0.48$\pm$0.02 & 0.30$\pm$0.03 & 11.50 & \cellcolor{red!60}11.92 \\
TabPFN (max) & 0.48$\pm$0.02 & 0.01$\pm$0.00 & 0.01$\pm$0.00 & 0.17$\pm$0.01 & 9.75 & 0.12$\pm$0.01 & 0.29$\pm$0.01 & 0.03$\pm$0.01 & 0.15$\pm$0.01 & \cellcolor{red!20}12.50 & 0.47$\pm$0.02 & 0.29$\pm$0.03 & \cellcolor{red!40}12.50 & \cellcolor{red!20}11.58 \\
\textsc{ConTextTab} (mean) & 0.75$\pm$0.02 & 0.05$\pm$0.02 & 0.05$\pm$0.02 & 0.28$\pm$0.02 & 6.50 & 0.19$\pm$0.02 & 0.43$\pm$0.02 & 0.09$\pm$0.01 & 0.24$\pm$0.02 & 6.25 & 0.71$\pm$0.02 & 0.70$\pm$0.03 & 8.00 & 6.92 \\
\textsc{ConTextTab} (max) & 0.46$\pm$0.03 & 0.00$\pm$0.00 & 0.00$\pm$0.00 & 0.15$\pm$0.01 & \cellcolor{red!20}11.50 & 0.16$\pm$0.02 & 0.39$\pm$0.02 & 0.08$\pm$0.01 & 0.21$\pm$0.02 & 9.75 & 0.64$\pm$0.02 & 0.60$\pm$0.02 & 9.00 & 10.08 \\
\bottomrule
\end{tabular}
}
\end{table*}

\paragraph{Stat labeling shows that retrieval and linear probing for majority-type labels are near-perfect, with
schema features dominating the remaining signal.}
For the Stat labeling scheme (Table~\ref{tab:results_real_stat}), retrieval and linear probe accuracy are near-saturated for many methods, so the more discriminative signal comes from triplet and clustering. In this regime, HashingSchema dominates overall, consistent with majority-type grouping being largely determined by column-type and schema tokens, while LLMText remains the strongest non-schema alternative.

\begin{table*}[!ht]
\centering
\small
\caption{D2 Metrics for Real data - Stat Labels (mean $\pm$ 95\% CI over 10 seeds)}
\label{tab:results_real_stat}
\resizebox{\textwidth}{!}{
\begin{tabular}{lccccc|ccccc|ccc|c}
\toprule
\multirow{2}{*}{Embedder} & \multicolumn{5}{c}{Triplet (Stat)} & \multicolumn{5}{c}{Clustering (Stat)} & \multicolumn{3}{c}{Retrieval/Probe (Stat)} & \multicolumn{1}{c}{Overall} \\
\cmidrule(lr){2-6} \cmidrule(lr){7-11} \cmidrule(lr){12-14} \cmidrule(lr){15-15}
 & TR-R & TR-H & TR-CH & TR-Avg & Rank & Purity & NMI & ARI & CL-Avg & Rank & R@5 & LP & Rank & Avg Rank \\
\midrule
HashingSchema & \textbf{0.78$\pm$0.04} & \textbf{0.57$\pm$0.05} & \textbf{0.60$\pm$0.05} & \textbf{0.65$\pm$0.04} & \cellcolor{green!60}1.00 & \textbf{0.91$\pm$0.03} & \textbf{0.58$\pm$0.09} & \textbf{0.67$\pm$0.10} & \textbf{0.72$\pm$0.07} & \cellcolor{green!60}1.00 & \textbf{1.00$\pm$0.00} & 0.84$\pm$0.02 & \cellcolor{green!20}4.50 & \cellcolor{green!60}2.17 \\
SchemaContent & 0.24$\pm$0.02 & 0.00$\pm$0.01 & 0.01$\pm$0.01 & 0.09$\pm$0.01 & \cellcolor{red!20}11.50 & 0.62$\pm$0.03 & 0.04$\pm$0.03 & -0.02$\pm$0.01 & 0.21$\pm$0.01 & \cellcolor{red!40}12.75 & \textbf{1.00$\pm$0.00} & 0.96$\pm$0.01 & \cellcolor{green!60}2.00 & 8.75 \\
TableStatistics & 0.17$\pm$0.02 & 0.00$\pm$0.00 & 0.00$\pm$0.00 & 0.06$\pm$0.01 & \cellcolor{red!60}14.00 & 0.62$\pm$0.02 & 0.08$\pm$0.03 & 0.03$\pm$0.02 & 0.25$\pm$0.02 & 9.00 & 0.98$\pm$0.00 & \textbf{0.99$\pm$0.00} & 5.00 & 9.33 \\
StatisticalSummary & 0.67$\pm$0.03 & 0.06$\pm$0.02 & 0.06$\pm$0.02 & 0.26$\pm$0.02 & \cellcolor{green!20}3.25 & 0.78$\pm$0.02 & 0.23$\pm$0.06 & 0.31$\pm$0.06 & 0.44$\pm$0.05 & \cellcolor{green!20}3.00 & 0.99$\pm$0.00 & 0.90$\pm$0.02 & 5.50 & \cellcolor{green!20}3.92 \\
MatrixFactorization & 0.47$\pm$0.01 & 0.00$\pm$0.00 & 0.00$\pm$0.00 & 0.16$\pm$0.00 & 9.00 & 0.64$\pm$0.02 & 0.01$\pm$0.01 & 0.01$\pm$0.01 & 0.22$\pm$0.01 & 11.00 & 0.96$\pm$0.00 & 0.65$\pm$0.02 & \cellcolor{red!60}13.50 & \cellcolor{red!60}11.17 \\
HashingText & 0.54$\pm$0.01 & 0.06$\pm$0.01 & 0.06$\pm$0.01 & 0.22$\pm$0.01 & 3.75 & 0.70$\pm$0.04 & 0.18$\pm$0.05 & 0.17$\pm$0.07 & 0.35$\pm$0.05 & 4.25 & 0.99$\pm$0.00 & 0.83$\pm$0.02 & 7.00 & 5.00 \\
LLMText & 0.68$\pm$0.02 & 0.10$\pm$0.02 & 0.11$\pm$0.02 & 0.30$\pm$0.02 & \cellcolor{green!40}2.00 & 0.82$\pm$0.07 & 0.35$\pm$0.10 & 0.45$\pm$0.12 & 0.54$\pm$0.09 & \cellcolor{green!40}2.00 & \textbf{0.99$\pm$0.00} & 0.90$\pm$0.01 & 5.00 & \cellcolor{green!40}3.00 \\
TableVectorizer & 0.43$\pm$0.01 & 0.01$\pm$0.00 & 0.02$\pm$0.00 & 0.15$\pm$0.00 & 8.00 & 0.64$\pm$0.01 & 0.07$\pm$0.02 & 0.04$\pm$0.04 & 0.25$\pm$0.02 & 8.75 & 0.95$\pm$0.01 & 0.68$\pm$0.04 & \cellcolor{red!40}13.50 & \cellcolor{red!40}10.08 \\
HyTREL & 0.45$\pm$0.01 & 0.01$\pm$0.00 & 0.01$\pm$0.00 & 0.16$\pm$0.00 & 8.25 & 0.61$\pm$0.03 & 0.04$\pm$0.02 & -0.01$\pm$0.01 & 0.21$\pm$0.01 & \cellcolor{red!60}12.75 & 0.98$\pm$0.00 & 0.90$\pm$0.02 & 7.50 & 9.50 \\
Armadillo & 0.51$\pm$0.01 & 0.02$\pm$0.00 & 0.02$\pm$0.00 & 0.18$\pm$0.00 & 6.00 & 0.62$\pm$0.04 & 0.02$\pm$0.01 & 0.01$\pm$0.01 & 0.22$\pm$0.02 & \cellcolor{red!20}12.00 & 0.97$\pm$0.00 & 0.71$\pm$0.03 & \cellcolor{red!20}11.00 & \cellcolor{red!20}9.67 \\
TabPFN (mean) & 0.36$\pm$0.03 & 0.00$\pm$0.00 & 0.00$\pm$0.00 & 0.12$\pm$0.01 & 11.50 & 0.70$\pm$0.03 & 0.11$\pm$0.02 & 0.15$\pm$0.04 & 0.32$\pm$0.03 & 5.00 & 0.96$\pm$0.00 & 0.76$\pm$0.01 & 11.00 & 9.17 \\
TabPFN (max) & 0.43$\pm$0.03 & 0.01$\pm$0.00 & 0.01$\pm$0.01 & 0.15$\pm$0.01 & 9.25 & 0.67$\pm$0.02 & 0.06$\pm$0.02 & 0.10$\pm$0.03 & 0.27$\pm$0.02 & 7.75 & 0.96$\pm$0.00 & 0.76$\pm$0.02 & 11.00 & 9.33 \\
\textsc{ConTextTab} (mean) & 0.53$\pm$0.02 & 0.02$\pm$0.01 & 0.03$\pm$0.01 & 0.20$\pm$0.01 & 5.00 & 0.64$\pm$0.02 & 0.07$\pm$0.03 & -0.03$\pm$0.01 & 0.23$\pm$0.01 & 10.00 & \textbf{0.99$\pm$0.00} & 0.94$\pm$0.01 & \cellcolor{green!40}3.00 & 6.00 \\
\textsc{ConTextTab} (max) & 0.35$\pm$0.03 & 0.00$\pm$0.00 & 0.00$\pm$0.00 & 0.12$\pm$0.01 & \cellcolor{red!40}12.50 & 0.68$\pm$0.02 & 0.12$\pm$0.04 & 0.12$\pm$0.04 & 0.31$\pm$0.03 & 5.75 & 0.98$\pm$0.00 & 0.92$\pm$0.01 & 5.50 & 7.92 \\
\bottomrule
\end{tabular}
}
\end{table*}

\paragraph{Averaging over label types hides specialization, but still provides useful intuition.}
When averaged over label types (Table~\ref{tab:results_real_appendix_avg}), HashingSchema ranks best on average triplet ranking, LLMText ranks best via clustering alignment, and HashingText ranks best on retrieval and probe. 
This decomposition matches the interpretation in the main paper, as different labeling schemes lead to different similarity notions, so a single averaged D2 score can hide meaningful specialization.

\begin{table*}[!ht]
\centering
\small
\caption{Detailed D2 Metrics for Real data - Averaged over Label Types (mean $\pm$ 95\% CI over 10 seeds)}
\label{tab:results_real_appendix_avg}
\resizebox{\textwidth}{!}{
\begin{tabular}{lccccc|ccccc|ccc|c}
\toprule
\multirow{2}{*}{Embedder} & \multicolumn{5}{c}{Triplet (Avg over Labels)} & \multicolumn{5}{c}{Clustering (Avg over Labels)} & \multicolumn{3}{c}{Retrieval/Probe (Avg over Labels)} & \multicolumn{1}{c}{Overall} \\
\cmidrule(lr){2-6} \cmidrule(lr){7-11} \cmidrule(lr){12-14} \cmidrule(lr){15-15}
 & TR-R & TR-H & TR-CH & TR-Avg & Rank & Purity & NMI & ARI & CL-Avg & Rank & R@5 & LP & Rank & Avg Rank \\
\midrule
HashingSchema & \textbf{0.89$\pm$0.02} & \textbf{0.43$\pm$0.04} & \textbf{0.44$\pm$0.03} & \textbf{0.59$\pm$0.03} & \cellcolor{green!60}1.00 & \textbf{0.41$\pm$0.02} & \textbf{0.53$\pm$0.04} & \textbf{0.27$\pm$0.04} & \textbf{0.40$\pm$0.03} & \cellcolor{green!60}1.00 & 0.96$\pm$0.01 & 0.84$\pm$0.02 & \cellcolor{green!20}3.50 & \cellcolor{green!60}1.83 \\
SchemaContent & 0.32$\pm$0.02 & 0.00$\pm$0.00 & 0.00$\pm$0.00 & 0.11$\pm$0.01 & \cellcolor{red!40}12.50 & 0.28$\pm$0.01 & 0.23$\pm$0.02 & 0.01$\pm$0.01 & 0.17$\pm$0.01 & \cellcolor{red!60}13.50 & 0.80$\pm$0.01 & \textbf{0.87$\pm$0.02} & 4.00 & 10.00 \\
TableStatistics & 0.25$\pm$0.02 & 0.00$\pm$0.00 & 0.00$\pm$0.00 & 0.08$\pm$0.01 & \cellcolor{red!60}14.00 & 0.30$\pm$0.01 & 0.34$\pm$0.02 & 0.07$\pm$0.01 & 0.24$\pm$0.01 & 6.00 & 0.76$\pm$0.01 & 0.59$\pm$0.02 & 9.00 & 9.67 \\
StatisticalSummary & 0.77$\pm$0.02 & 0.08$\pm$0.01 & 0.08$\pm$0.01 & 0.31$\pm$0.01 & 4.75 & 0.37$\pm$0.01 & 0.39$\pm$0.03 & 0.16$\pm$0.03 & 0.31$\pm$0.02 & \cellcolor{green!20}3.25 & 0.85$\pm$0.01 & 0.54$\pm$0.02 & 8.00 & 5.33 \\
MatrixFactorization & 0.63$\pm$0.01 & 0.00$\pm$0.00 & 0.00$\pm$0.00 & 0.21$\pm$0.00 & 9.50 & 0.30$\pm$0.01 & 0.26$\pm$0.01 & 0.04$\pm$0.01 & 0.20$\pm$0.01 & 10.00 & 0.65$\pm$0.01 & 0.28$\pm$0.02 & \cellcolor{red!20}12.50 & 10.67 \\
HashingText & 0.82$\pm$0.01 & 0.35$\pm$0.02 & 0.37$\pm$0.02 & 0.51$\pm$0.02 & \cellcolor{green!20}3.00 & 0.35$\pm$0.02 & 0.39$\pm$0.03 & 0.10$\pm$0.03 & 0.28$\pm$0.03 & 3.75 & \textbf{0.98$\pm$0.00} & \textbf{0.87$\pm$0.02} & \cellcolor{green!60}1.50 & \cellcolor{green!20}2.75 \\
LLMText & \textbf{0.89$\pm$0.01} & 0.38$\pm$0.02 & 0.39$\pm$0.03 & 0.55$\pm$0.02 & \cellcolor{green!40}2.00 & 0.40$\pm$0.03 & \textbf{0.53$\pm$0.04} & 0.25$\pm$0.05 & 0.40$\pm$0.04 & \cellcolor{green!40}2.00 & \textbf{0.97$\pm$0.00} & 0.86$\pm$0.02 & \cellcolor{green!40}2.50 & \cellcolor{green!40}2.17 \\
TableVectorizer & 0.50$\pm$0.01 & 0.03$\pm$0.01 & 0.03$\pm$0.01 & 0.18$\pm$0.01 & 8.25 & 0.28$\pm$0.01 & 0.21$\pm$0.02 & 0.03$\pm$0.02 & 0.17$\pm$0.01 & \cellcolor{red!40}13.25 & 0.58$\pm$0.01 & 0.29$\pm$0.03 & \cellcolor{red!60}13.50 & \cellcolor{red!60}11.67 \\
HyTREL & 0.68$\pm$0.01 & 0.03$\pm$0.01 & 0.03$\pm$0.01 & 0.25$\pm$0.01 & 7.00 & 0.30$\pm$0.02 & 0.29$\pm$0.02 & 0.05$\pm$0.01 & 0.21$\pm$0.01 & 9.50 & 0.85$\pm$0.01 & 0.76$\pm$0.02 & 5.00 & 7.17 \\
Armadillo & 0.72$\pm$0.01 & 0.16$\pm$0.02 & 0.17$\pm$0.02 & 0.35$\pm$0.01 & 4.25 & 0.31$\pm$0.02 & 0.28$\pm$0.02 & 0.05$\pm$0.01 & 0.21$\pm$0.01 & 8.00 & 0.86$\pm$0.02 & 0.65$\pm$0.03 & 5.50 & 5.92 \\
TabPFN (mean) & 0.38$\pm$0.02 & 0.01$\pm$0.00 & 0.01$\pm$0.00 & 0.13$\pm$0.01 & 11.00 & 0.30$\pm$0.01 & 0.23$\pm$0.01 & 0.07$\pm$0.02 & 0.20$\pm$0.01 & 9.75 & 0.60$\pm$0.02 & 0.42$\pm$0.02 & 11.50 & \cellcolor{red!20}10.75 \\
TabPFN (max) & 0.47$\pm$0.02 & 0.01$\pm$0.00 & 0.01$\pm$0.00 & 0.16$\pm$0.01 & 9.50 & 0.29$\pm$0.01 & 0.21$\pm$0.02 & 0.05$\pm$0.01 & 0.18$\pm$0.01 & \cellcolor{red!20}11.50 & 0.60$\pm$0.01 & 0.42$\pm$0.02 & \cellcolor{red!40}12.50 & \cellcolor{red!40}11.17 \\
\textsc{ConTextTab} (mean) & 0.68$\pm$0.02 & 0.03$\pm$0.01 & 0.04$\pm$0.01 & 0.25$\pm$0.01 & 6.25 & 0.31$\pm$0.02 & 0.31$\pm$0.02 & 0.04$\pm$0.01 & 0.22$\pm$0.01 & 7.50 & 0.76$\pm$0.01 & 0.70$\pm$0.02 & 7.00 & 6.92 \\
\textsc{ConTextTab} (max) & 0.42$\pm$0.03 & 0.00$\pm$0.00 & 0.00$\pm$0.00 & 0.14$\pm$0.01 & \cellcolor{red!20}12.00 & 0.31$\pm$0.01 & 0.30$\pm$0.02 & 0.08$\pm$0.02 & 0.23$\pm$0.02 & 6.00 & 0.72$\pm$0.02 & 0.64$\pm$0.02 & 9.00 & 9.00 \\
\bottomrule
\end{tabular}
}
\end{table*}

\subsection{Training and validation details for learned representation heads}
\label{sec:app_head_training_details}

The learned representation heads reported in the paper follow the default real-data training configuration and are selected with a lightweight validation pass before full benchmark evaluation. Table~\ref{tab:head_train_eval_protocol} summarizes the differences between training, checkpoint selection, and final reporting.

 The training objective is aligned with the desiderata, but it is not identical to the benchmark metrics.
The head is trained with an IoU-based SOFTCE listwise objective over same-table partitions (the SCDM objective of Section~\ref{sec:frozen_ctt}), while D1 is reported as the Spearman correlation between IoU and cosine similarity over within-table pairs. Likewise, the training loss uses Direct same-table identity (which can be obtained for free) only to define local support and, through local/global normalization, induces an implicit contrastive signal against other tables in the batch. It does not train directly for the full D2 suite used in evaluation. The metrics of triplet ranking, clustering alignment, retrieval, and linear probing are all measured only at evaluation time. The learned-composition experiment therefore tests how much a learned compositional rule can improve the table-level embeddings of frozen \textsc{ConTextTab} states under an IoU-based self-supervised objective that is aligned with, but not identical to, the benchmark.

\begin{table*}[!ht]
\centering
\caption{Training, validation, and final benchmark protocols for learned representation heads.}
\label{tab:head_train_eval_protocol}
\small
\renewcommand{\arraystretch}{1.2}
\setlength{\tabcolsep}{5pt}
\resizebox{\textwidth}{!}{
\begin{tabular}{p{2.1cm} p{2.4cm} p{3.2cm} p{4.4cm} p{7.0cm}}
\toprule
\textbf{Phase} & \textbf{Data source} & \textbf{Tables / partitions} & \textbf{Partition policy} & \textbf{Notes} \\
\midrule

Training
& \makecell[l]{Real\\(training)}
& \makecell[l]{16 tables per batch\\8 partitions per table}
& \makecell[l]{Row/column fractions: $0.5/0.5$\\Minimum partition: $8\times4$\\Overlap ratio: $0.7$}
& \makecell[l]{Train the head on top of a fixed encoder\\IoU-based SOFTCE listwise objective\\Same-table support defined by Direct identity}

\\
\midrule

Validation
& \makecell[l]{Real\\(validation)}
& \makecell[l]{20 tables\\5 partitions per table}
& \makecell[l]{Row/column fractions \\ sampled from $[0.2,0.5]$\\Minimum partition: $10\times5$\\Overlap ratio: $0.5$}
& \makecell[l]{Evaluator-based \texttt{val\_score}\\Mean of D1 Spearman and averaged D2 sections\\Computed on the validation pass \\ Used to select checkpoints and seeds}

\\
\midrule

Benchmark
& \makecell[l]{Real + synthetic\\(evaluation)}
& \makecell[l]{100 tables\\10 partitions per table}
& \makecell[l]{Row/column fractions \\ sampled from $[0.2,0.5]$\\Minimum partition: $10\times5$\\Overlap ratio: $0.5$}
& \makecell[l]{Full D1--D3 benchmark reporting\\Evaluation-time partitions\\Synthetic results reflect transfer from heads \\ trained on real tables}

\\
\bottomrule
\end{tabular}
}
\end{table*}

Each head configuration is trained with three training seeds (1, 42, 96). All three runs are scored on the validation split with the \texttt{val\_score} above, and the checkpoint with the highest score is kept. The selected head then enters the final benchmark as a single fixed model and is evaluated with the same sampler seeds as all other methods. The validation split is disjoint from the tables used in the final benchmark.

The current setup uses separate dataset roots and separate samplers for optimization, checkpoint selection, and final reporting. In particular, training uses a fixed partition policy ($0.5/0.5$ row/column fractions, minimum partition $8\times4$, overlap ratio $0.7$), whereas validation and final benchmark reporting use a different evaluation-time policy with row/column fractions sampled from $[0.2,0.5]$, minimum partition $10\times5$, and overlap ratio $0.5$. The final benchmark also includes synthetic evaluation although the heads are trained on real data only. We therefore read the learned-head study as a controlled experiment under a shifted evaluation protocol, where the head is trained on real data under one partitioning regime, then evaluated under different sampler settings and, in the final benchmark, also on synthetic data. The improvements therefore extend beyond the exact training configuration. How such compositional rules generalize across unseen source tables, other partition regimes, and new data families remains open.

\subsection{Full ablation for learned representation heads}
\label{sec:app_head_ablation}

For completeness, we report the full $\alpha$-sweep for all representation-head families and both data regimes. Each entry is the validation-selected checkpoint of that head family and $\alpha$, as described in Appendix~\ref{sec:app_head_training_details}.
The representative configurations shown in the main paper
(Table~\ref{tab:head_ablation_compact_combined})
are, within each head family and data regime, the configurations with the lowest D1--D3 average rank; they serve as a compact summary, and the conclusions are stated over the full sweep.
This section makes the selection transparent and shows that the main claim does not depend on a single hyperparameter setting.

The full sweep reveals three consistent patterns.
First, the learned heads improve over the pooled \textsc{ConTextTab} baseline on D1 and D2 across the sweep; the only exception is Attention with $\alpha=1$ on real data, which does not improve D1 or triplet ranking.
Second, the best trade-off rarely occurs at the same $\alpha$ for all metrics, as
D1 often prefers a stronger local component, while D2 benefits from a mixture of local and global supervision.
Third, the architectural differences are systematic rather than incidental, as
Attention tends to peak on D1 and often provides the best overall quality ranking, while GatedAttention and GatedMLP remain competitive on retrieval and robustness.
\begin{table*}[ht]
\centering
\footnotesize
\setlength{\tabcolsep}{4pt}

\resizebox{\textwidth}{!}{
\begin{tabular}{llc|ccc|ccc}
\toprule
\multirow{2}{*}{Head} & \multirow{2}{*}{$\alpha$} & \multicolumn{1}{c|}{D1 Consistency} & \multicolumn{3}{c|}{D2 Label-Based - Avg} & \multicolumn{3}{c}{D3 Robustness} \\
\cmidrule(lr){3-3} \cmidrule(lr){4-6} \cmidrule(lr){7-9}
 & & Spearman & TR-Avg & CL-Avg & R@5 & Perm & Noise & Mask \\
\midrule
\textsc{ConTextTab} (pooled) & -- & 0.04$\pm$0.03 & 0.18$\pm$0.00 & 0.27$\pm$0.02 & 0.70$\pm$0.01 & 0.97$\pm$0.00 & 0.97$\pm$0.00 & 0.97$\pm$0.00 \\
\midrule
\multirow{5}{*}{Attention} & 0.00 & 0.44$\pm$0.02 & 0.35$\pm$0.01 & 0.46$\pm$0.03 & 0.87$\pm$0.01 & \textbf{1.00$\pm$0.00} & \textbf{0.99$\pm$0.00} & 0.98$\pm$0.00 \\
 & 0.25 & 0.46$\pm$0.02 & 0.35$\pm$0.01 & 0.44$\pm$0.01 & 0.87$\pm$0.01 & \textbf{1.00$\pm$0.00} & \textbf{0.99$\pm$0.00} & \textbf{0.99$\pm$0.00} \\
 & 0.50 & 0.45$\pm$0.02 & 0.34$\pm$0.01 & \textbf{0.47$\pm$0.02} & 0.88$\pm$0.01 & \textbf{1.00$\pm$0.00} & \textbf{1.00$\pm$0.00} & \textbf{0.99$\pm$0.00} \\
 & 0.75 & \textbf{0.50$\pm$0.02} & 0.34$\pm$0.01 & 0.46$\pm$0.02 & 0.88$\pm$0.01 & \textbf{1.00$\pm$0.00} & \textbf{1.00$\pm$0.00} & \textbf{0.99$\pm$0.00} \\
 & 1.00 & 0.47$\pm$0.02 & 0.28$\pm$0.01 & 0.46$\pm$0.02 & 0.87$\pm$0.01 & \textbf{1.00$\pm$0.00} & \textbf{1.00$\pm$0.00} & \textbf{0.99$\pm$0.00} \\
\midrule
\multirow{5}{*}{GatedAttention} & 0.00 & 0.47$\pm$0.02 & 0.34$\pm$0.01 & 0.44$\pm$0.03 & 0.88$\pm$0.01 & \textbf{1.00$\pm$0.00} & \textbf{0.99$\pm$0.00} & \textbf{0.99$\pm$0.00} \\
 & 0.25 & 0.45$\pm$0.02 & 0.35$\pm$0.01 & 0.43$\pm$0.04 & 0.88$\pm$0.01 & \textbf{1.00$\pm$0.00} & \textbf{0.99$\pm$0.00} & \textbf{0.98$\pm$0.00} \\
 & 0.50 & 0.48$\pm$0.02 & \textbf{0.37$\pm$0.01} & 0.46$\pm$0.02 & 0.89$\pm$0.01 & \textbf{1.00$\pm$0.00} & \textbf{0.99$\pm$0.00} & \textbf{0.99$\pm$0.00} \\
 & 0.75 & 0.43$\pm$0.02 & 0.34$\pm$0.01 & 0.45$\pm$0.03 & 0.88$\pm$0.01 & \textbf{1.00$\pm$0.00} & \textbf{0.99$\pm$0.00} & \textbf{0.99$\pm$0.00} \\
 & 1.00 & 0.48$\pm$0.02 & 0.31$\pm$0.01 & 0.46$\pm$0.03 & \textbf{0.89$\pm$0.01} & \textbf{1.00$\pm$0.00} & \textbf{1.00$\pm$0.00} & \textbf{0.99$\pm$0.00} \\
\midrule
\multirow{5}{*}{GatedMLP} & 0.00 & 0.46$\pm$0.02 & 0.33$\pm$0.01 & \textbf{0.47$\pm$0.02} & \textbf{0.89$\pm$0.01} & \textbf{1.00$\pm$0.00} & \textbf{0.99$\pm$0.00} & \textbf{0.99$\pm$0.00} \\
 & 0.25 & 0.46$\pm$0.02 & 0.33$\pm$0.01 & 0.44$\pm$0.04 & \textbf{0.89$\pm$0.01} & \textbf{1.00$\pm$0.00} & \textbf{0.99$\pm$0.00} & \textbf{0.99$\pm$0.00} \\
 & 0.50 & 0.47$\pm$0.02 & 0.34$\pm$0.01 & 0.44$\pm$0.05 & 0.88$\pm$0.01 & \textbf{1.00$\pm$0.00} & \textbf{1.00$\pm$0.00} & \textbf{0.99$\pm$0.00} \\
 & 0.75 & 0.48$\pm$0.02 & 0.30$\pm$0.01 & 0.43$\pm$0.04 & 0.88$\pm$0.01 & \textbf{1.00$\pm$0.00} & \textbf{1.00$\pm$0.00} & \textbf{0.99$\pm$0.00} \\
 & 1.00 & 0.46$\pm$0.02 & 0.31$\pm$0.01 & 0.45$\pm$0.04 & \textbf{0.89$\pm$0.01} & \textbf{1.00$\pm$0.00} & \textbf{1.00$\pm$0.00} & \textbf{0.99$\pm$0.00} \\
\bottomrule
\end{tabular}
}

\vspace{0.8em}

\includegraphics[width=\textwidth]{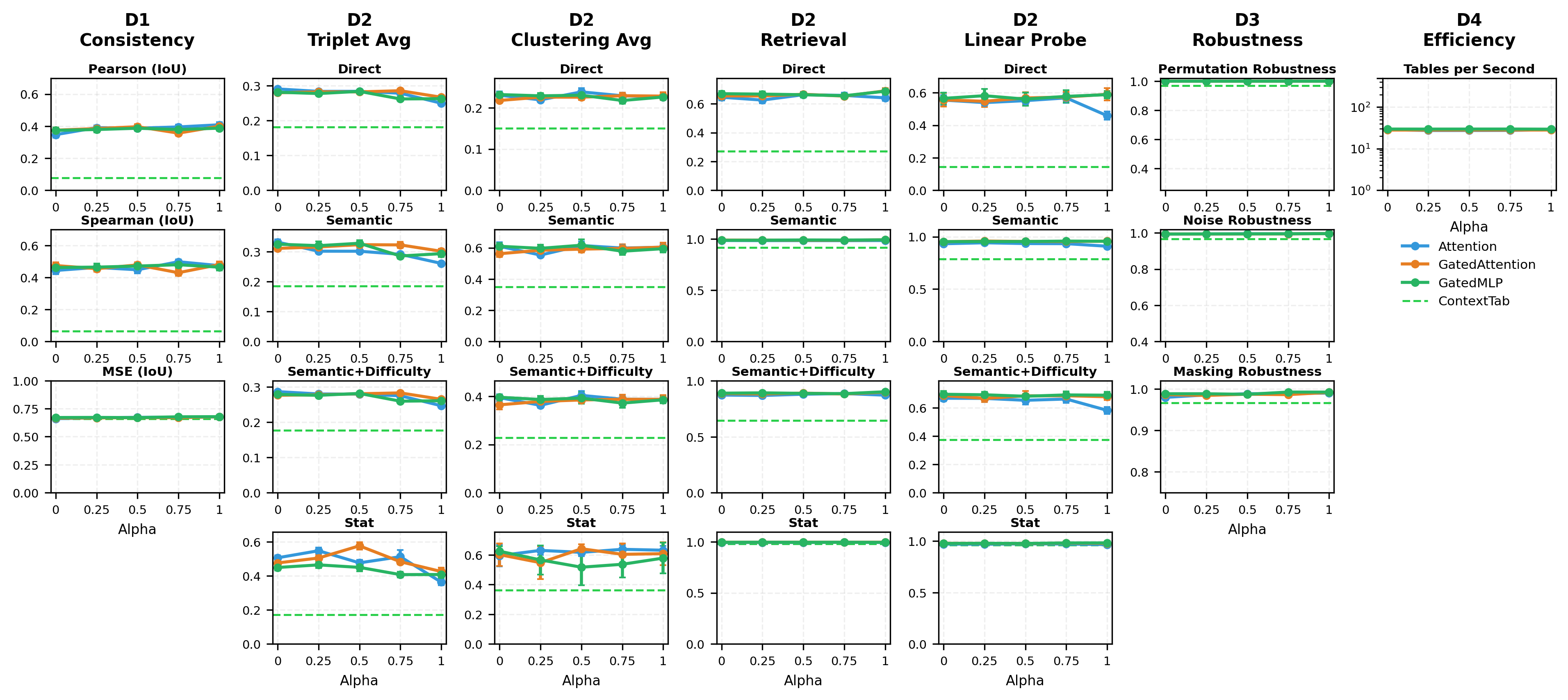}

\caption{Full Synthetic-data ablation for learned representation heads on top of frozen \textsc{ConTextTab}. The table reports the complete $\alpha$-sweep. The plot below shows each metric as a function of $\alpha$. The dashed green line corresponds to the pooled \textsc{ConTextTab} baseline.}
\label{tabfig:head_ablation_synthetic_full}
\end{table*}

\begin{table*}[ht]
\centering
\footnotesize
\setlength{\tabcolsep}{4pt}

\resizebox{\textwidth}{!}{
\begin{tabular}{llc|ccc|ccc}
\toprule
\multirow{2}{*}{Head} & \multirow{2}{*}{$\alpha$} & \multicolumn{1}{c|}{D1 Consistency} & \multicolumn{3}{c|}{D2 Label-Based - Avg} & \multicolumn{3}{c}{D3 Robustness} \\
\cmidrule(lr){3-3} \cmidrule(lr){4-6} \cmidrule(lr){7-9}
 & & Spearman & TR-Avg & CL-Avg & R@5 & Perm & Noise & Mask \\
\midrule
\textsc{ConTextTab} (pooled) & -- & 0.19$\pm$0.03 & 0.25$\pm$0.01 & 0.22$\pm$0.01 & 0.76$\pm$0.01 & 0.97$\pm$0.00 & 0.97$\pm$0.00 & 0.97$\pm$0.00 \\
\midrule
\multirow{5}{*}{Attention} & 0.00 & 0.26$\pm$0.03 & 0.40$\pm$0.02 & 0.29$\pm$0.03 & \textbf{0.96$\pm$0.01} & \textbf{1.00$\pm$0.00} & 0.99$\pm$0.01 & 0.98$\pm$0.00 \\
 & 0.25 & \textbf{0.27$\pm$0.03} & 0.41$\pm$0.02 & 0.30$\pm$0.03 & \textbf{0.96$\pm$0.01} & \textbf{1.00$\pm$0.00} & 0.99$\pm$0.00 & 0.98$\pm$0.00 \\
 & 0.50 & 0.25$\pm$0.03 & \textbf{0.42$\pm$0.02} & 0.31$\pm$0.04 & \textbf{0.96$\pm$0.00} & \textbf{1.00$\pm$0.00} & \textbf{0.99$\pm$0.00} & 0.98$\pm$0.00 \\
 & 0.75 & 0.23$\pm$0.03 & 0.37$\pm$0.02 & 0.27$\pm$0.02 & 0.95$\pm$0.01 & \textbf{1.00$\pm$0.00} & \textbf{0.99$\pm$0.00} & \textbf{0.99$\pm$0.00} \\
 & 1.00 & 0.17$\pm$0.03 & 0.24$\pm$0.01 & 0.28$\pm$0.04 & 0.91$\pm$0.01 & \textbf{1.00$\pm$0.00} & \textbf{1.00$\pm$0.00} & \textbf{1.00$\pm$0.00} \\
\midrule
\multirow{5}{*}{GatedAttention} & 0.00 & 0.26$\pm$0.03 & 0.41$\pm$0.02 & 0.30$\pm$0.03 & \textbf{0.95$\pm$0.01} & \textbf{1.00$\pm$0.00} & \textbf{0.99$\pm$0.00} & 0.98$\pm$0.00 \\
 & 0.25 & 0.23$\pm$0.03 & \textbf{0.42$\pm$0.02} & \textbf{0.34$\pm$0.05} & \textbf{0.95$\pm$0.01} & \textbf{1.00$\pm$0.00} & 0.99$\pm$0.00 & 0.98$\pm$0.00 \\
 & 0.50 & 0.26$\pm$0.03 & 0.38$\pm$0.02 & 0.27$\pm$0.02 & 0.95$\pm$0.01 & \textbf{1.00$\pm$0.00} & \textbf{0.99$\pm$0.00} & 0.99$\pm$0.00 \\
 & 0.75 & 0.27$\pm$0.04 & 0.39$\pm$0.02 & 0.30$\pm$0.04 & \textbf{0.95$\pm$0.01} & \textbf{1.00$\pm$0.00} & \textbf{0.99$\pm$0.00} & 0.98$\pm$0.00 \\
 & 1.00 & 0.21$\pm$0.03 & 0.32$\pm$0.02 & 0.29$\pm$0.04 & 0.95$\pm$0.01 & \textbf{1.00$\pm$0.00} & \textbf{1.00$\pm$0.00} & \textbf{0.99$\pm$0.00} \\
\midrule
\multirow{5}{*}{GatedMLP} & 0.00 & 0.24$\pm$0.04 & 0.41$\pm$0.02 & 0.28$\pm$0.02 & \textbf{0.96$\pm$0.01} & \textbf{1.00$\pm$0.00} & \textbf{0.99$\pm$0.00} & 0.99$\pm$0.00 \\
 & 0.25 & 0.21$\pm$0.04 & 0.41$\pm$0.02 & 0.30$\pm$0.04 & \textbf{0.95$\pm$0.01} & \textbf{1.00$\pm$0.00} & \textbf{0.99$\pm$0.00} & 0.99$\pm$0.00 \\
 & 0.50 & 0.23$\pm$0.03 & 0.40$\pm$0.02 & 0.27$\pm$0.02 & \textbf{0.96$\pm$0.01} & \textbf{1.00$\pm$0.00} & \textbf{0.99$\pm$0.00} & 0.99$\pm$0.00 \\
 & 0.75 & 0.23$\pm$0.03 & 0.37$\pm$0.02 & 0.30$\pm$0.03 & 0.94$\pm$0.01 & \textbf{1.00$\pm$0.00} & \textbf{0.99$\pm$0.00} & \textbf{0.99$\pm$0.00} \\
 & 1.00 & 0.24$\pm$0.03 & 0.31$\pm$0.02 & 0.27$\pm$0.03 & 0.95$\pm$0.01 & \textbf{1.00$\pm$0.00} & \textbf{1.00$\pm$0.00} & \textbf{0.99$\pm$0.00} \\
\bottomrule
\end{tabular}
}

\vspace{0.8em}

\includegraphics[width=\textwidth]{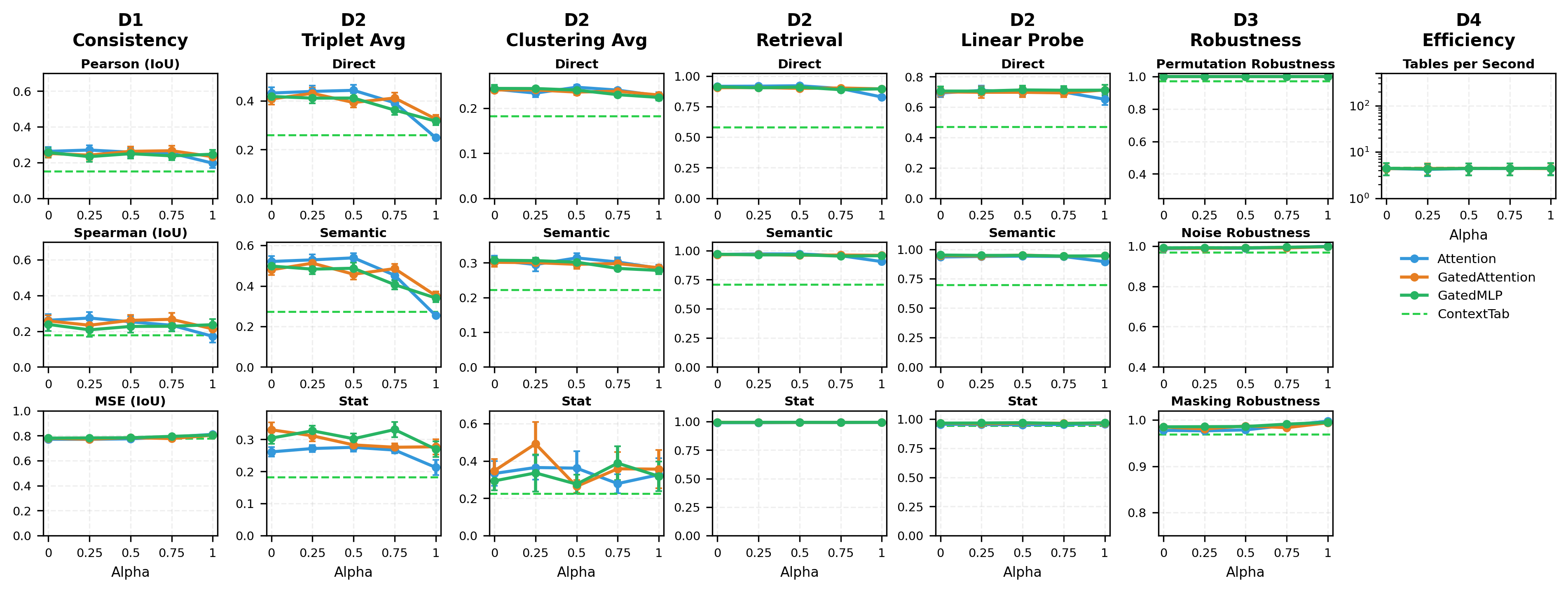}

\caption{Full Real-data ablation for learned representation heads on top of frozen \textsc{ConTextTab}. The table reports the complete $\alpha$-sweep. The plot below shows each metric as a function of $\alpha$. The dashed green line corresponds to the pooled \textsc{ConTextTab} baseline.}
\label{tabfig:head_ablation_real_full}
\end{table*}

\subsection{Learned \textsc{ConTextTab} heads versus benchmark baselines}
\label{sec:app_augmented_leaderboard}

To summarize how much learned composition closes the gap between pooled tabular foundation-model embeddings and the strongest lightweight baselines, we insert the strongest representative \textsc{ConTextTab} head from each family into the original leaderboard and recompute all ranks over the augmented method set.
For synthetic data, we use the representative variants from the main paper, namely
\textsc{ConTextTab}-R Attention ($\alpha=0.75$),
\textsc{ConTextTab}-R GatedAttention ($\alpha=1.0$),
and \textsc{ConTextTab}-R GatedMLP ($\alpha=0.75$).
For real data, we use
\textsc{ConTextTab}-R Attention ($\alpha=0.25$),
\textsc{ConTextTab}-R GatedAttention ($\alpha=0.5$),
and \textsc{ConTextTab}-R GatedMLP ($\alpha=1.0$).
Ranks are recomputed over the displayed values.
D2 rank is the average rank over triplet accuracy, clustering, and retrieval, the D2 components used for all comparisons of compositional rules (Section~\ref{sec:pooling_main});
D3 rank is the average rank over permutation, noise, and masking robustness;
and the overall average rank is the average of D1, D2, and D3 ranks. Baseline values are taken from Tables~\ref{tab:results_synthetic} and~\ref{tab:results_real}.

On synthetic data, learned composition moves \textsc{ConTextTab} from 14th place to the top of the augmented ranking, where \textsc{ConTextTab}-R Attention ranks first, GatedAttention third, and GatedMLP fifth, interleaved with HashingSchema and HashingText. On real data, the gains are smaller, and \textsc{ConTextTab}-R Attention ranks second behind HashingText, ahead of HashingSchema, while pooled \textsc{ConTextTab} ranks 12th. The heads' overall ranks rely partly on D3, where they match the most robust methods, whereas on D1 they remain below the best fingerprints.

\begin{table*}[!ht]
\centering
\small
\caption{Augmented synthetic-data leaderboard after adding the representative \textsc{ConTextTab}-R head from each family. Baseline values are those of Table~\ref{tab:results_synthetic} (TabPFN and \textsc{ConTextTab} under mean pooling) and head values those of Table~\ref{tab:head_ablation_compact_combined}; all values are means over 10 seeds. D2 rank averages the ranks of TR-Avg, CL-Avg, and R@5; D3 rank averages Perm, Noise, and Mask; Avg Rank averages D1, D2, and D3 ranks. Ranks are computed over the displayed values, with tied values sharing their average rank. Lower ranks are better.}
\label{tab:results_synthetic_ctt_augmented}
\resizebox{\textwidth}{!}{
\begin{tabular}{lcc|cccc|cccc|c}
\toprule
\multirow{2}{*}{Embedder} & \multicolumn{2}{c}{D1 Consistency} & \multicolumn{4}{c}{D2 Label-Based - Avg} & \multicolumn{4}{c}{D3 Robustness} & \multicolumn{1}{c}{Overall} \\
\cmidrule(lr){2-3} \cmidrule(lr){4-7} \cmidrule(lr){8-11} \cmidrule(lr){12-12}
 & Spearman & Rank & TR-Avg & CL-Avg & R@5 & Rank & Perm & Noise & Mask & Rank & Avg Rank \\
\midrule
\textbf{\textsc{ConTextTab}-R Attention ($\alpha=0.75$)} & 0.50 & \cellcolor{green!20}3.00 & 0.34 & 0.46 & 0.88 & 5.00 & 1.00 & 1.00 & 0.99 & \cellcolor{green!20}5.67 & \cellcolor{green!60}4.56 \\
HashingSchema & 0.59 & \cellcolor{green!60}1.00 & 0.53 & 0.54 & 0.71 & 5.33 & 1.00 & 0.98 & 0.99 & 7.67 & \cellcolor{green!40}4.67 \\
\textbf{\textsc{ConTextTab}-R GatedAttention ($\alpha=1.0$)} & 0.48 & 4.50 & 0.31 & 0.46 & 0.89 & 4.67 & 1.00 & 1.00 & 0.99 & \cellcolor{green!20}5.67 & \cellcolor{green!20}4.94 \\
HashingText & 0.54 & \cellcolor{green!40}2.00 & 0.47 & 0.46 & 1.00 & \cellcolor{green!60}2.67 & 1.00 & 0.65 & 0.97 & 10.33 & 5.00 \\
\textbf{\textsc{ConTextTab}-R GatedMLP ($\alpha=0.75$)} & 0.48 & 4.50 & 0.30 & 0.43 & 0.88 & 7.00 & 1.00 & 1.00 & 0.99 & \cellcolor{green!20}5.67 & 5.72 \\
StatisticalSummary & 0.38 & 8.00 & 0.36 & 0.44 & 0.91 & \cellcolor{green!20}4.00 & 1.00 & 0.99 & 1.00 & \cellcolor{green!40}5.33 & 5.78 \\
MatrixFactorization & 0.42 & 6.00 & 0.23 & 0.32 & 0.79 & 9.67 & 1.00 & 1.00 & 1.00 & \cellcolor{green!60}4.00 & 6.56 \\
TableStatistics & 0.34 & 9.00 & 0.13 & 0.34 & 0.71 & \cellcolor{red!20}11.83 & 1.00 & 1.00 & 1.00 & \cellcolor{green!60}4.00 & 8.28 \\
LLMText & 0.28 & 12.00 & 0.51 & 0.54 & 0.88 & \cellcolor{green!40}3.17 & 0.97 & 0.98 & 0.99 & 10.00 & 8.39 \\
Armadillo & 0.39 & 7.00 & 0.31 & 0.31 & 0.89 & 7.00 & 1.00 & 0.48 & 0.87 & \cellcolor{red!40}11.50 & 8.50 \\
SchemaContent & 0.29 & 10.50 & 0.01 & 0.26 & 0.78 & \cellcolor{red!40}12.67 & 1.00 & 1.00 & 1.00 & \cellcolor{green!60}4.00 & 9.06 \\
HyTREL & 0.16 & \cellcolor{red!20}13.00 & 0.24 & 0.33 & 0.84 & 8.67 & 0.99 & 0.98 & 0.89 & \cellcolor{red!20}11.33 & 11.00 \\
TableVectorizer & 0.29 & 10.50 & 0.17 & 0.22 & 0.74 & \cellcolor{red!60}12.83 & 0.29 & 1.00 & 0.80 & \cellcolor{red!20}11.33 & \cellcolor{red!20}11.56 \\
\textbf{\textsc{ConTextTab} (pooled)} & 0.04 & \cellcolor{red!40}14.00 & 0.18 & 0.27 & 0.70 & \cellcolor{red!40}12.67 & 0.97 & 0.97 & 0.97 & \cellcolor{red!60}12.00 & \cellcolor{red!40}12.89 \\
TabPFN & -0.08 & \cellcolor{red!60}15.00 & 0.17 & 0.25 & 0.73 & \cellcolor{red!60}12.83 & 0.75 & 0.93 & 0.99 & \cellcolor{red!40}11.50 & \cellcolor{red!60}13.11 \\
\bottomrule
\end{tabular}
}
\end{table*}

\begin{table*}[!ht]
\centering
\small
\caption{Augmented real-data leaderboard after adding the representative \textsc{ConTextTab}-R head from each family. Baseline values are those of Table~\ref{tab:results_real} (TabPFN and \textsc{ConTextTab} under mean pooling) and head values those of Table~\ref{tab:head_ablation_compact_combined}; all values are means over 10 seeds. D2 rank averages the ranks of TR-Avg, CL-Avg, and R@5; D3 rank averages Perm, Noise, and Mask; Avg Rank averages D1, D2, and D3 ranks. Ranks are computed over the displayed values, with tied values sharing their average rank. Lower ranks are better.}
\label{tab:results_real_ctt_augmented}
\resizebox{\textwidth}{!}{
\begin{tabular}{lcc|cccc|cccc|c}
\toprule
\multirow{2}{*}{Embedder} & \multicolumn{2}{c}{D1 Consistency} & \multicolumn{4}{c}{D2 Label-Based - Avg} & \multicolumn{4}{c}{D3 Robustness} & \multicolumn{1}{c}{Overall} \\
\cmidrule(lr){2-3} \cmidrule(lr){4-7} \cmidrule(lr){8-11} \cmidrule(lr){12-12}
 & Spearman & Rank & TR-Avg & CL-Avg & R@5 & Rank & Perm & Noise & Mask & Rank & Avg Rank \\
\midrule
HashingText & 0.41 & \cellcolor{green!60}1.00 & 0.51 & 0.28 & 0.98 & \cellcolor{green!20}3.00 & 1.00 & 0.73 & 0.98 & 9.33 & \cellcolor{green!60}4.44 \\
\textbf{\textsc{ConTextTab}-R Attention ($\alpha=0.25$)} & 0.27 & \cellcolor{green!20}3.00 & 0.41 & 0.30 & 0.96 & 3.83 & 1.00 & 0.99 & 0.98 & 6.83 & \cellcolor{green!40}4.56 \\
HashingSchema & 0.31 & \cellcolor{green!40}2.00 & 0.59 & 0.40 & 0.96 & \cellcolor{green!40}2.00 & 1.00 & 0.94 & 0.95 & 10.17 & \cellcolor{green!20}4.72 \\
\textbf{\textsc{ConTextTab}-R GatedAttention ($\alpha=0.5$)} & 0.26 & 4.50 & 0.38 & 0.27 & 0.95 & 5.67 & 1.00 & 0.99 & 0.99 & \cellcolor{green!20}6.00 & 5.39 \\
\textbf{\textsc{ConTextTab}-R GatedMLP ($\alpha=1.0$)} & 0.24 & 7.00 & 0.31 & 0.27 & 0.95 & 6.50 & 1.00 & 1.00 & 0.99 & \cellcolor{green!40}4.83 & 6.11 \\
StatisticalSummary & 0.19 & 9.50 & 0.31 & 0.31 & 0.85 & 6.33 & 1.00 & 1.00 & 1.00 & \cellcolor{green!60}3.67 & 6.50 \\
SchemaContent & 0.26 & 4.50 & 0.11 & 0.17 & 0.80 & \cellcolor{red!20}12.83 & 1.00 & 1.00 & 1.00 & \cellcolor{green!60}3.67 & 7.00 \\
TableStatistics & 0.21 & 8.00 & 0.08 & 0.24 & 0.76 & 11.50 & 1.00 & 1.00 & 1.00 & \cellcolor{green!60}3.67 & 7.72 \\
LLMText & 0.13 & \cellcolor{red!20}13.00 & 0.55 & 0.40 & 0.97 & \cellcolor{green!60}1.83 & 0.91 & 0.95 & 0.99 & 10.00 & 8.28 \\
Armadillo & 0.25 & 6.00 & 0.35 & 0.21 & 0.86 & 7.83 & 1.00 & 0.60 & 0.78 & \cellcolor{red!20}11.83 & 8.56 \\
MatrixFactorization & 0.16 & 11.00 & 0.21 & 0.20 & 0.65 & 12.17 & 1.00 & 1.00 & 1.00 & \cellcolor{green!60}3.67 & 8.94 \\
\textbf{\textsc{ConTextTab} (pooled)} & 0.19 & 9.50 & 0.25 & 0.22 & 0.76 & 10.00 & 0.97 & 0.97 & 0.97 & 10.83 & 10.11 \\
HyTREL & 0.15 & 12.00 & 0.25 & 0.21 & 0.85 & 9.50 & 0.99 & 0.89 & 0.96 & \cellcolor{red!40}12.00 & \cellcolor{red!20}11.17 \\
TabPFN & 0.06 & \cellcolor{red!40}14.00 & 0.13 & 0.20 & 0.60 & \cellcolor{red!40}13.17 & 0.79 & 0.98 & 0.97 & 11.00 & \cellcolor{red!40}12.72 \\
TableVectorizer & 0.03 & \cellcolor{red!60}15.00 & 0.18 & 0.17 & 0.58 & \cellcolor{red!60}13.83 & 0.33 & 0.98 & 0.85 & \cellcolor{red!60}12.50 & \cellcolor{red!60}13.78 \\
\bottomrule
\end{tabular}
}
\end{table*}

\subsection{Controlled D4-efficiency experiment}
\label{sec:app_efficiency}

We report D4 separately from the D1-D3 quality tables because throughput is best measured in a controlled encode-only setting rather than mixed into quality rankings.

The controlled study evaluates six fixed scenarios obtained over three input modalities (numeric, text, mixed) with two table sizes.
Small size (S) denotes a fixed $64\times16$ table and large size (L) denotes a fixed $2048\times64$ table.
The L condition is an intentional scaling stress test for encode-time behavior and is therefore substantially larger than the partition budget used in the main D1--D3 benchmark.

For each embedder and scenario, the evaluator generates 50 synthetic tables once, runs one untimed warm-up call to \texttt{encode\_tables}, and then measures throughput over 5 timed encode-only repeats on the same 50 tables.
We report throughput in tables per second (Tbl/s) together with $95\%$ confidence intervals over repeats.
As the main aggregate summary, we use the geometric mean (GeoMean) across the six scenario means.
The Rank column in Table~\ref{tab:controlled_efficiency_combined} is computed from GeoMean only.

Numeric tables contain normally distributed \texttt{float32} values, text tables contain short random lowercase ASCII strings of length $4$--$12$, and mixed tables split the columns between numeric and text values.
In particular, \textsc{ConTextTab} and the learned \textsc{ConTextTab}-R heads are evaluated with their own tokenizer and backbone, while LLMText is the separate sentence-transformer baseline and uses at most the first 32 rows.
For serialization-based methods such as HashingText, numeric tables need not be faster than text tables because numeric \texttt{float32} values are serialized as decimal CSV strings, whereas the controlled text cells are short tokens.

Table~\ref{tab:controlled_efficiency_combined} shows a clear order-of-magnitude spread in encode throughput.
HashingSchema is the fastest method overall, followed by SchemaContent and TableStatistics.
Prediction-oriented and graph-based methods are substantially slower under this controlled setup.
For the learned heads on frozen \textsc{ConTextTab}, we report all three representative variants used in the paper, namely Attention, GatedAttention, and Gated MLP.
Their throughput remains very close to pooled \textsc{ConTextTab}, showing that learned pooling improves representation quality while leaving the overall D4 efficiency largely unchanged.

\begin{table*}[!ht]
\centering
\small
\caption{Controlled encode-only efficiency results for baseline embedders and learned \textsc{ConTextTab}-R heads. Throughput is reported in tables per second (Tbl/s). S denotes $64\times16$ tables and L denotes $2048\times64$ tables. GeoMean is computed over the six scenario means, and Rank is computed from GeoMean only for the main baseline models. Baseline embedders are listed above the separator; learned heads are listed below it without ranks.}
\label{tab:controlled_efficiency_combined}
\resizebox{\textwidth}{!}{
\begin{tabular}{lcccccccc}
\toprule
Embedder & Numeric-S & Numeric-L & Text-S & Text-L & Mixed-S & Mixed-L & GeoMean Tbl/s & Rank \\
\midrule
HashingSchema & 9793.76$\pm$112.47 & 2618.63$\pm$28.77 & 11675.20$\pm$77.04 & 3213.25$\pm$6.29 & 11694.96$\pm$911.74 & 3206.73$\pm$56.35 & 5748.43$\pm$104.02 & \cellcolor{green!60}1.00 \\
SchemaContent & 2293.53$\pm$17.80 & 928.61$\pm$2.98 & 1653.73$\pm$7.43 & 161.07$\pm$1.08 & 1765.45$\pm$31.58 & 270.29$\pm$4.26 & 804.30$\pm$3.58 & \cellcolor{green!40}2.00 \\
TableStatistics & 1471.32$\pm$15.68 & 201.71$\pm$4.73 & 1227.06$\pm$79.31 & 53.83$\pm$1.73 & 1201.78$\pm$10.75 & 87.43$\pm$3.62 & 356.70$\pm$8.17 & \cellcolor{green!20}3.00 \\
StatisticalSummary & 529.07$\pm$1.23 & 163.42$\pm$5.61 & 116.95$\pm$0.31 & 13.66$\pm$0.17 & 161.30$\pm$2.37 & 24.64$\pm$0.20 & 90.48$\pm$0.73 & 6.00 \\
MatrixFactorization & 368.22$\pm$11.62 & 72.43$\pm$0.26 & 360.33$\pm$9.88 & 73.04$\pm$0.91 & 355.93$\pm$1.53 & 74.27$\pm$0.50 & 162.71$\pm$0.96 & 5.00 \\
HashingText & 715.00$\pm$1.51 & 6.41$\pm$0.05 & 1265.68$\pm$1.79 & 12.26$\pm$0.04 & 900.99$\pm$3.22 & 8.34$\pm$0.13 & 90.10$\pm$0.34 & 7.00 \\
LLMText & 391.27$\pm$10.30 & 144.85$\pm$1.53 & 490.64$\pm$2.04 & 198.64$\pm$2.69 & 424.95$\pm$7.25 & 165.63$\pm$1.10 & 270.16$\pm$0.82 & 4.00 \\
TableVectorizer & 39.34$\pm$0.78 & 9.75$\pm$0.25 & 4.38$\pm$0.01 & 0.08$\pm$0.00 & 8.36$\pm$0.09 & 0.16$\pm$0.00 & 2.38$\pm$0.02 & \cellcolor{red!20}10.00 \\
HyTREL & 3.54$\pm$0.05 & 2.80$\pm$0.02 & 3.47$\pm$0.03 & 2.76$\pm$0.02 & 3.55$\pm$0.02 & 2.79$\pm$0.02 & 3.13$\pm$0.02 & 9.00 \\
Armadillo & 17.74$\pm$0.53 & 0.09$\pm$0.00 & 19.03$\pm$0.45 & 0.09$\pm$0.00 & 18.60$\pm$0.28 & 0.09$\pm$0.00 & 1.31$\pm$0.01 & \cellcolor{red!60}12.00 \\
TabPFN & 6.65$\pm$0.12 & 2.87$\pm$0.07 & 6.93$\pm$0.09 & 2.38$\pm$0.01 & 7.16$\pm$0.08 & 2.63$\pm$0.03 & 4.25$\pm$0.02 & 8.00 \\
\textsc{ConTextTab} & 12.16$\pm$0.10 & 0.27$\pm$0.00 & 23.54$\pm$0.07 & 0.12$\pm$0.00 & 15.99$\pm$0.12 & 0.17$\pm$0.00 & 1.72$\pm$0.00 & \cellcolor{red!40}11.00 \\
\midrule
\multicolumn{9}{l}{\emph{Learned heads on frozen \textsc{ConTextTab}}} \\
\midrule
\textsc{ConTextTab}-R Attention & 11.76$\pm$0.06 & 0.26$\pm$0.00 & 21.55$\pm$0.06 & 0.12$\pm$0.00 & 15.27$\pm$0.05 & 0.17$\pm$0.00 & 1.65$\pm$0.00 & - \\
\textsc{ConTextTab}-R GatedAttention & 11.73$\pm$0.04 & 0.26$\pm$0.00 & 21.40$\pm$0.07 & 0.12$\pm$0.00 & 15.25$\pm$0.07 & 0.17$\pm$0.00 & 1.65$\pm$0.00 & - \\
\textsc{ConTextTab}-R Gated MLP & 12.07$\pm$0.04 & 0.28$\pm$0.00 & 22.48$\pm$0.07 & 0.12$\pm$0.00 & 15.61$\pm$0.04 & 0.17$\pm$0.00 & 1.71$\pm$0.00 & - \\
\bottomrule
\end{tabular}
}
\end{table*}

\subsection{Code and implementation details}
\label{app:code_implementation}

The implementation is organized as an extensible benchmark harness. A shared sampler materializes partial-table views with coordinate sets and multi-granularity labels, and each embedding method implements the same \texttt{encode\_tables} interface. The main \texttt{Evaluator} computes D1--D3 metrics from shared sampled partitions, the \texttt{EfficiencyEvaluator} isolates D4 through encode-only throughput measurements, and the \texttt{DownstreamEvaluator} implements the closed-corpus parent-table retrieval proof of concept. For the learned \textsc{ConTextTab}-R heads, the code uses separate training and evaluation dataset roots, batched fixed-shape partitions, vectorized pairwise similarity/overlap computations, Hydra/OmegaConf configurations, checkpoint evaluation, and ablation rendering. 

We provide an anonymized ZIP package with the benchmark code, configurations, reproduction scripts, and table/figure rendering utilities as supplementary material; a clean public repository will be released upon acceptance.

\subsection{Large language model (LLM) usage disclosure}
We used large language models (ChatGPT from OpenAI and Claude from Anthropic) to improve grammar and clarity, to assist in restructuring and drafting revised text for this resubmission, and to suggest code refactorings during implementation.
All benchmark design decisions, experiments, results, and conclusions are the authors' own; all LLM outputs were reviewed and validated, and the authors take full responsibility for the content.

\subsection{T-SNE plots}
We provide t-SNE projections of partition embeddings as a qualitative sanity check for the label-wise D2 decomposition (Appendix~\ref{sec:app_results}).
Each row corresponds to an embedding model, and columns show the same embeddings colored by different labeling schemes.
We emphasize that t-SNE is not metric-faithful; global distances are distorted and apparent cluster sizes are not comparable across methods.
Nevertheless, label-homogeneous neighborhoods and visually coherent group structures often mirror the separability trends measured by triplet and clustering scores in the per-label D2 tables.

\begin{figure}[!t]
    \centering
    \vspace{-1cm}
    \includegraphics[height=\textheight]{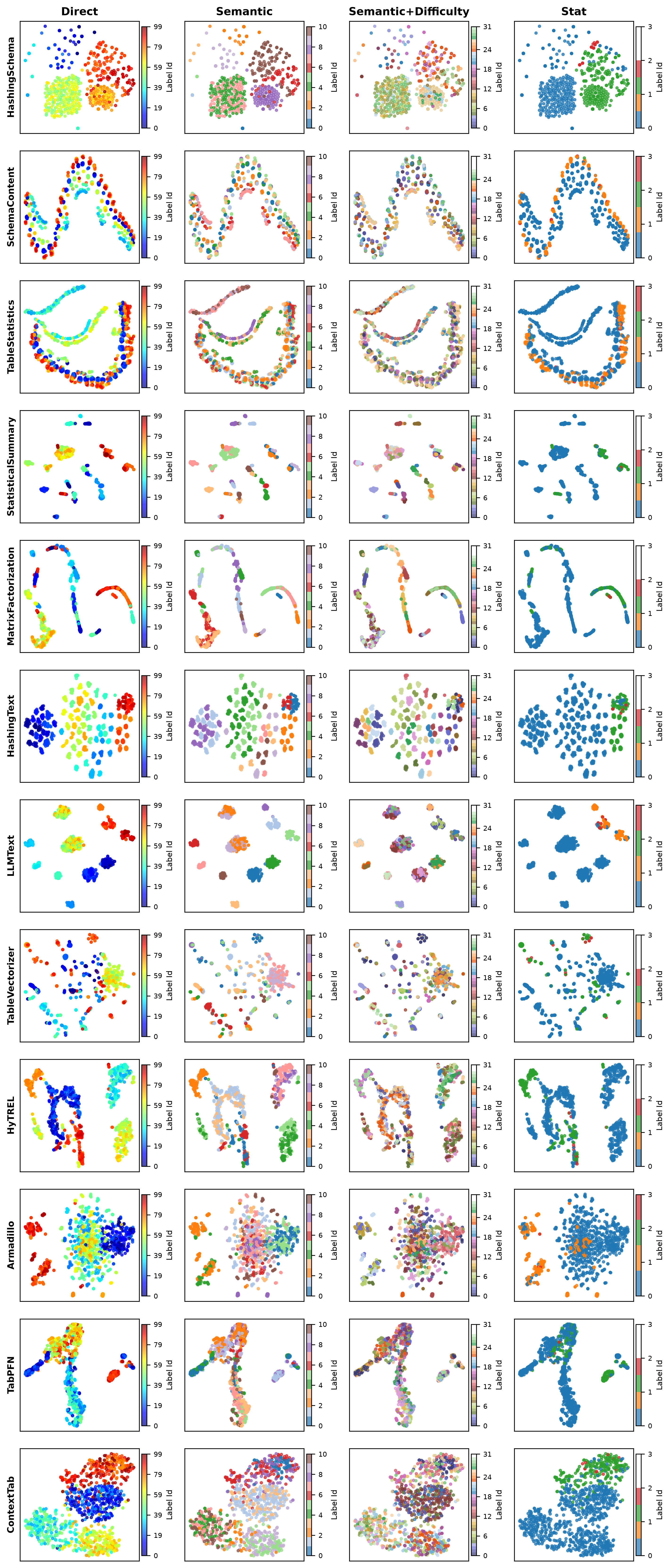}
    \caption{t-SNE projections of partition embeddings for synthetic data, colored by different labeling schemes (Direct, Semantic, Semantic+Difficulty, Stat).}
    \label{fig:tsne-sync}
\end{figure}

\paragraph{Synthetic.}
For synthetic corpora (Figure~\ref{fig:tsne-sync}), the t-SNE projections qualitatively mirror the label-type dependence observed in the D2 decomposition (Appendix~\ref{sec:app_results}).
Under Direct labeling, methods that sustain hard-negative identity ranking (notably HashingText) exhibit more locally label-consistent neighborhoods, whereas pooled TFM representations (e.g., TabPFN, ConTextTab) form mixed-color manifolds with limited fine-grained separation.
Under Semantic labeling, schema- and text-driven methods yield clearer group structure, aligning with their higher clustering scores.
Crucially, moving from Semantic to Semantic+Difficulty visibly reduces separability for schema-centric fingerprints (e.g., HashingSchema), consistent with the Semantic+Difficulty triplet degradation in Table~\ref{tab:results_synthetic_meta_3}; this supports the interpretation that the added difficulty factor is weakly expressed in schema tokens and is instead reflected in value distributions and interaction patterns.
We therefore treat t-SNE strictly as a qualitative check that supports the quantitative D2 conclusions reported in Tables~\ref{tab:results_synthetic_direct}--\ref{tab:results_synthetic_meta_3}.

\begin{figure}[!t]
    \centering
     \vspace{-1cm}
    \includegraphics[height=\textheight]{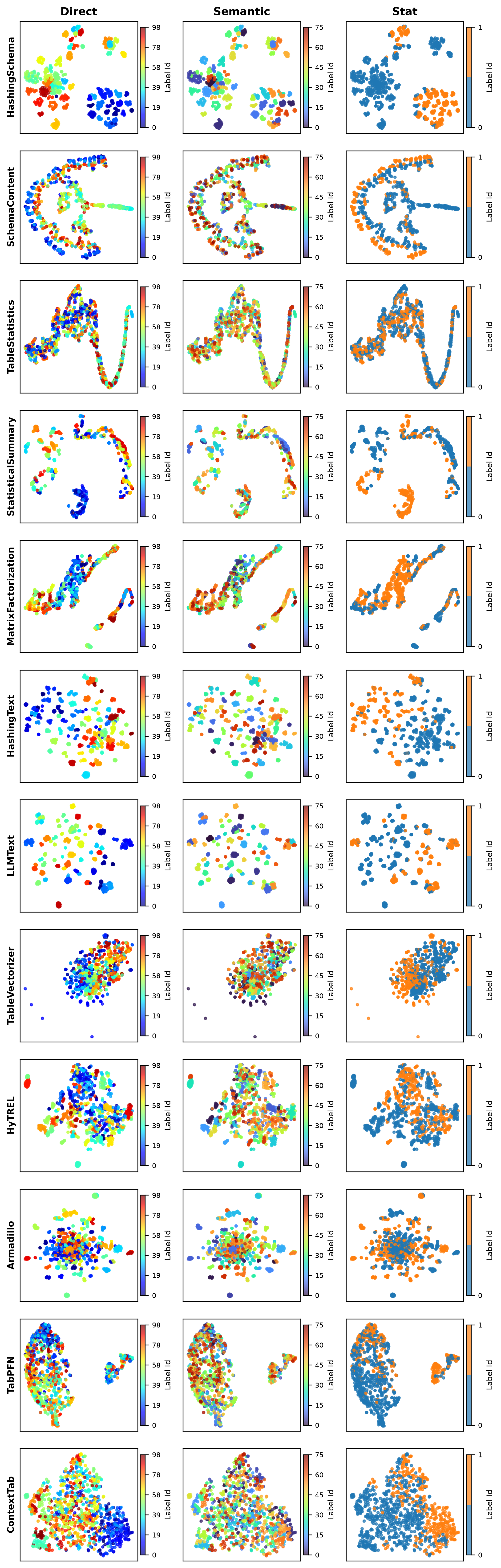}
    \caption{t-SNE projections of partition embeddings for real open-source data, colored by different labeling schemes (Direct, Semantic, Stat).}
    \label{fig:tsne-real}
\end{figure}

\paragraph{Real open-source.}
For open-source tables (Figure~\ref{fig:tsne-real}), the qualitative structure mirrors the D2 label dependence in Tables~\ref{tab:results_real_direct}--\ref{tab:results_real_stat}. 
Under Stat labeling, many methods exhibit visually separated regions, consistent with the relative ease of coarse type grouping.
Under Direct labels, most embedding models show substantial color mixing, consistent with stricter identity ranking remaining challenging. Triplet ranking under hard and cluster-hard negatives exposes near-collisions even when retrieval and linear probe can be high.
Semantic labels typically sit between these extremes, with semantic- and content-driven methods showing clearer group structure than purely schema- and statistical baselines.

\end{document}